\documentclass{article}

\title{Higher-Order-LaSDI}

\usepackage{amsmath}
\usepackage{amssymb}
\usepackage{amsthm}
\usepackage{graphicx}
\usepackage{subcaption}
\usepackage[numbers]{natbib}
\usepackage[margin=1in]{geometry}

\usepackage[utf8]{inputenc}         
\usepackage[T1]{fontenc}            
\usepackage{hyperref}               
\usepackage{url}                    
\usepackage{booktabs}               
\usepackage{amsfonts}               
\usepackage{nicefrac}               
\usepackage{microtype}              
\usepackage{xcolor}                 
\usepackage{parskip}                
\usepackage[bb=boondox]{mathalfa}   

\usepackage{tabulary}
\usepackage{booktabs}
\usepackage[table]{xcolor}

\usepackage{makecell} 
\usepackage{tabularx}

\newcommand{\vc}{\mathbf}

\title{Higher-Order LaSDI: Reduced Order Modeling with Multiple Time Derivatives.}
\date{}

\author{
    Robert Stephany \\
    Center for Applied Scientific Computing \\
    Lawrence Livermore National Laboratory \\
    Livermore, Ca 94550 \\
    \texttt{stephany1@llnl.gov}
    \and 
    William Anderson \\
    Center for Applied Scientific Computing \\
    Lawrence Livermore National Laboratory \\
    Livermore, Ca 94550 \\
    \texttt{anderson316@llnl.gov} \\  
    \and
    Youngsoo Choi \\
    Center for Applied Scientific Computing \\
    Lawrence Livermore National Laboratory \\
    Livermore, Ca 94550 \\
    \texttt{choi15@llnl.gov} \\
}

\begin{document}

\maketitle

\begin{abstract}
Solving complex partial differential equations (PDEs) is essential across scientific disciplines but often requires numerical models that can be prohibitively expensive in time sensitive applications. 
Reduced-order models (ROMs) address this challenge by exploiting low-dimensional structure to create fast approximations. 
The Latent Space Dynamics Identification (LaSDI) framework has shown success in learning ROMs for parameterized PDE families, but is limited to first-order systems. 
In this paper, we propose Higher-Order LaSDI (HLaSDI), which extends the LaSDI framework to PDEs with arbitrary time derivative orders. 
This generalization significantly expands the applicability of LaSDI-based methods to systems previously beyond their scope, including hyperbolic PDEs. 
We demonstrate HLaSDI's accuracy and efficiency on several linear and nonlinear benchmark problems.
\end{abstract}

\section{Introduction}
\label{Intro}

Since the late 2000s, computational power, algorithmic advances \cite{glorot2010initialization, hinton2006networks, vaswani2017attention}, and open-source automatic differentiation libraries \cite{paszke2019pytorch, abadi2016tensorflow} have made \emph{Deep Neural Networks (DNNs)} viable function approximators. 
DNNs offer key advantages: nonlinearity, customizable architectures with problem-specific inductive biases, and training dominated by highly-parallelizable matrix-matrix products. These attributes enable DNNs to approximate extremely complex functions for tasks like image classification \cite{he2016residual, krizhevsky2017alexnet}, language translation \cite{he2016residual, krizhevsky2017alexnet}, and language modeling \cite{brown2020gpt3}.

The success of DNNs spawned \emph{Scientific Machine Learning (SciML)}, which applies machine learning to science and engineering challenges. 
SciML has produced novel PDE solvers \cite{raissi2019pinn}, system identification and PDE discovery algorithms \cite{brunton2016sindy, rudy2017pdefind, rackauckas2020ude, stephany2022pde_read, stephany2024pde_learn, stephany2025dde_find}, and \emph{Reduced-Order Models (ROMs)} that use dimensionality reduction to accelerate simulations \cite{he2023glasdi, bonneville2024lasdi_review}.. 
This paper introduces a novel ROM framework. 
This section highlights key milestones in modern DNN-based ROM development.

Numerically solving \emph{Partial Differential Equations (PDEs)} is central to physical sciences \cite{thijssen2007comp_physics} and engineering \cite{jasak2007openfoam, anderson2021mfem}. 
However, high-fidelity solvers are too expensive \cite{andrej2024mfem_exascale} for applications requiring rapid solutions. 
Of particular interest are parameterized PDE families (with varying coefficients, initial conditions, or boundary conditions). 
For instance, online control algorithms repeatedly solve governing PDEs with different control inputs \cite{kaiser2018sindy_mpc} to determine appropriate control policies. 
ROMs address this by providing fast surrogate models that yield approximate PDE solutions \cite{fries2022lasdi}.

We refer to the governing PDE with its \emph{initial (IC)} and \emph{boundary conditions (BCs)} as the \emph{full-order model (FOM)}.

A typical ROM consists of (1) an encoder that maps \emph{FOM snapshots} (solutions at specific times) to a low-dimensional \emph{latent space}, (2) \emph{latent dynamics} (typically ODEs) describing the time evolution of encoded snapshots, and (3) a decoder that maps latent states back to FOM snapshots \cite{berkooz1993pod, schmid2010dmd, fries2022lasdi, champion2019sindy_autoencoder}. 
The decoder approximates the inverse of the encoder when restricted to FOM snapshots.

Once constructed, a ROM approximates FOM solutions through the following workflow:
\begin{enumerate}
    \item Encode the initial FOM state to its latent representation.
    \item Solve the latent dynamics using a numerical ODE solver to obtain the future latent state.
    \item Decode the final latent state back to an approximate final FOM state.

\end{enumerate}
Time-stepping in latent space --- often involving simple, linear dynamics --- can be orders of magnitude faster than solving the FOM \cite{bonneville2024gplasdi}.

Many early ROM techniques, such as \emph{Proper Orthogonal Decomposition} \cite{berkooz1993pod} and \emph{Dynamic Mode Decomposition} \cite{schmid2010dmd}, used linear projection methods. 
These approaches begin with a subspace $S$ of the FOM space $U$, a basis $v_1, \ldots, v_N$ of $S$, and a linear projection operator $P \in \mathcal{L}(U, S)$ onto $S$. 
Given $x \in U$ with $Px = c_1 v_1 + \cdots + c_N v_N$, the encoder $E : U \to \mathbb{R}^N$ maps $x$ to $c_1 e_1 + \cdots + c_N e_N \in \mathbb{R}^N$, where $e_k$ denotes the $k$th standard basis vector. 
The latent dynamics is a system of ODEs in $\mathbb{R}^N$, and the decoder $D : \mathbb{R}^N \to U$ maps $d_1 e_1 + \cdots + d_N e_N$ to $d_1 v_1 + \cdots + d_N v_N \in U$. 
Linear ROMs have succeeded in many applications, including solving Navier-Stokes equations \cite{iliescu2014NavierStokesPOD, stabile2018NavierStokesPOD}, Advection-Diffusion equations \cite{kim2021AdvectionDiffusionROMs, mclaughlin2016AdvectionDiffusionROM}, and design optimization \cite{choi2019DesignROM, mcbane2021DesignROM}. 
Despite this success, linear ROMs struggle in complex systems like advection-dominated fluid simulations \cite{fries2022lasdi}.

Several DNN-based ROMs have been proposed using \emph{autoencoders} --- DNN pairs serving as encoder and decoder. 
\emph{SINDy-Autoencoder} \cite{champion2019sindy_autoencoder} assumes linear latent dynamics and learns them by embedding the SINDy \cite{brunton2016sindy} residual into the autoencoder loss function. 
Rubanova et al. \cite{rubanova2019latent_NODE} introduced Latent Neural ODEs, uses Neural ODEs \cite{chen2018NODE} to enhance \emph{Recurrent Neural Networks (RNNs)}.
Their approach uses the hidden state output by one RNN cell as the initial condition to a Neural ODE, whose future solution acts as the hidden input to the next RNN cell. 
This approach allows RNNs, which usually are limited to uniformly spaced time series, to model non-uniform time series.
The paper employs both a classical RNN and an encoder-decoder model to predicting future states from past trajectories.

These approaches have been extended to parameterized families of differential equations.
Parametrized Neural ODE \cite{lee2021parameterized_NODE} extends Neural ODE to solve parameterized\footnote{Parameters may alter the initial condition and/or governing equation.} ODE families.
Similarly, \emph{Latent Space Dynamics Identification (LaSDI)} \cite{fries2022lasdi} extended SINDy-Autoencoder to parameterized PDE families.
We explore the SINDy-Autoencoder, LaSDI frameworks, and related extensions in Section \ref{Background}.

\textbf{Contributions:} This paper builds upon Rollout-LaSDI and GPLaSDI to build a non-intrusive ROM framework for parameterized systems of PDEs with multiple time derivatives.
We generalize to FOMs with $N_t \in \mathbb{N}$ time derivatives, where the FOM solution comprises the PDE solution and its first $N_t - 1$ time derivatives.
Our ROM utilizes $N_t$ autoencoders, where the $k$th operates on the $k$th time derivative of the FOM solution.
The latent dynamics consists of an $N_t$th-order system of linear ODEs for a joint latent variable, $z$.
The latent space of the $k$'th autoencoder approximates the $k$'th derivative of $z$.
A higher-order ODE solver solves the latent dynamics to predict future ROM solutions, which we map through the decoders to predict the future FOM solution.
We exploit problem structure to couple network training—for instance, training the $k$th encoder to predict the time derivative of the $(k-1)$th encoder.
This structure yields multiple novel loss functions that stabilize training.
We also introduce a novel extension of the Rollout loss \cite{stephany2025rollout}, further enhancing test-time performance.
Finally, we extend the finite difference schemes from \cite{stephany2025rollout} (which provide higher-order accuracy on non-uniform time series) to multiple time derivatives.
At test time, we use GPLaSDI's approach to identify latent dynamics for new parameter instances.
These techniques yield a flexible ROM framework for parameterized families of higher-order PDEs, which we call \emph{Higher-Order LaSDI} or \emph{HLaSDI} for short.
We demonstrate its efficacy on several FOMs, including hyperbolic linear systems and nonlinear oscillatory problems from mechanics.

\textbf{Outline:} In section \ref{Problem}, we formally state the problem we aim to solve.
Next, in section \ref{Background}, we outline algorithms that Higher-Order LaSDI builds upon, including LaSDI \cite{fries2022lasdi}, GPLaSDI \cite{bonneville2024gplasdi, bonneville2023gplasdi_neuralips}, and Rollout-LaSDI \cite{stephany2025rollout}.
In section \ref{Method}, we describe the HLaSDI algorithm in detail.
Then, in section \ref{Experiments}, we demonstrate HLaSDI's performance on several example problems.
Section \ref{Discussion} discusses some of these results and dives deeper into various aspects of our algorithm. 
We finish with concluding remarks in section \ref{Conclusion}.

\section{Problem Statement}
\label{Problem}

The \emph{Full Order Model (FOM)} is a initial value problem for a \emph{Partial Differential Equation (PDE)}:
\begin{align}
    \frac{d^{K} u_{\theta}}{dt^{K}} \left(t, X \right) &= \left(F \left(u_{\theta}, \theta \right) \right)\left(t, X\right) \qquad &&t \in (0, T],\ X \in \Omega \label{eq:PDE} \\
    D_t^k u_{\theta}\left( 0, X \right) &= u_0^{(k)} \left(X, \theta \right) \qquad && X \in \Omega,\ k \in \left\{ 0, 1, \ldots, K - 1 \right\} \label{eq:IC}
\end{align}
Here, $K \in \mathbb{N}$ denotes the highest-order time derivative in the PDE.
$F$ is a nonlinear partial derivative operator, $T > 0$ is the final solution time, and $\Omega$ is the spatial portion of the FOM domain (in this paper, $\Omega$ is a subset of Euclidean space).
We refer to $(0, T] \times \Omega$ as the FOM's \emph{problem domain}.
The PDE solution $u_{\theta} : (0, T] \times \Omega \to \mathbb{R}^{N_v}$ is \emph{vector-valued}, meaning it takes values in $\mathbb{R}^{N_v}$ for some $N_v \in \mathbb{N}$; thus \eqref{eq:PDE} represents a system of nonlinear PDEs.
We assume $F\left( \cdot, \theta \right)$ acts on the PDE solution and that time derivatives in $F\left( u_{\theta}, \theta \right)$ have order strictly below $K$.
Finally, $u_0^{(k)}$ denotes the initial condition for the $k$th time derivative of the FOM solution.
For brevity, we write $u_0 \equiv u_0^{(0)}$.

Here, $\theta \in \Theta$ denotes the \emph{parameter} value, where $\Theta$ is a subset of some finite dimensional inner product space.
Both $F$ and $u_0^{(k)}$ depend on $\theta$; changing $\theta$ alters the governing equation and initial condition, thereby changing the FOM solution.
We denote this dependence by writing $u_{\theta}$ for the solution when the parameter value is $\theta$.

We do not assume knowledge of the full FOM (equation \eqref{eq:PDE}); our approach is \emph{non-intrusive} \cite{bonneville2024gplasdi}.
However, we assume \eqref{eq:PDE} can be solved by a high-fidelity solver for all $\theta \in \Theta$ and has been solved for \emph{training parameters} $\{\theta_i\}_{i=1}^{N_{\theta}}$.
For FOMs with $K$ time derivatives, we assume the numerical solution consists of time series for $\vec{u}$ and its first $K - 1$ time derivatives.
Let $\{ D_t^k \vec{u}_{\theta}(t_j^{\theta})\}_{j = 0}^{N_t(\theta)} \subseteq \mathbb{R}^{N_u}$ denote the numerical solution for the $k$th time derivative at times $0 = t_0^{\theta}\le\cdots\le t_{N_t(\theta)}^{\theta} \le T$ when using $N_u$ spatial nodes to discretize the FOM's spatial domain.
We call $\mathbb{R}^{N_u}$ the \emph{discretized spatial domain} and write $\vec{u}_{\theta}(t_j^{\theta}) \equiv D_t^0 \vec{u}_{\theta} (t_j^{\theta})$.
Thus, the numerical FOM solution for parameter $\theta$ consists of $K$ time series with frames in $\mathbb{R}^{N_u}$. 
Note that $N_u$ is constant across parameters; we assume the spatial node locations (and correspondingly, the FOM domain) are fixed for all $\theta$.
The spatial nodes can be arbitrary points in $\Omega$, representing, e.g., grid points in a finite-difference scheme or vertices in a finite element mesh.

\textbf{Goal:} Higher-Order LaSDI aims to use numerical FOM solutions for training parameters $\{\theta_i\}_{i=1}^{N_{\theta}}$ to train a model that rapidly approximates $D_t^k u_{\theta}(t, X)$ for $k \in \{ 0, 1, \ldots, K - 1 \}$ and arbitrary $\theta \in \Theta$.
Specifically, we seek a model that predicts numerical FOM solutions for arbitrary parameter values; interpolating these solutions approximates $u_{\theta}$ at arbitrary points in the FOM problem domain.

Table \ref{Table:Problem:Notation} lists the notation we introduced in this section.

\begin{table}[hbt]
    \centering 
    \rowcolors{2}{cyan!10}{white}
    
    \begin{tabulary}{1.0\linewidth}{p{1.5cm}L}
        \toprule[0.3ex]
        \textbf{Notation} & \textbf{Meaning} \\
        \midrule[0.1ex]
        $K$ & The highest time derivative in the FOM. See equation \eqref{eq:PDE}. \\
        \addlinespace[0.4em]
        $\Omega$ & The spatial portion of the problem domain for the FOM. We assume this is some subset of a euclidean domain. \\
        \addlinespace[0.4em]
        $T$ & A positive real number representing the final time of the FOM solution. See equation \eqref{eq:PDE}. \\
        \addlinespace[0.4em]
        $N_V$ & A natural number denoting the dimensionality of the FOM solution. $u_{\theta}$ takes values in $\mathbb{R}^{N_v}$. \\
        \addlinespace[0.4em]
        $\theta$ & A parameter value. This determines both the FOM's initial condition and the governing equation. See equation \eqref{eq:PDE}. \\
        \addlinespace[0.4em]
        $N_{\theta}$ & The number of parameter values in the training set. \\
        \addlinespace[0.4em]
        $u_{\theta}$ & The FOM solution when $\theta$ is the parameter value (used to define the governing PDE and initial condition, see equations \eqref{eq:PDE} and \eqref{eq:IC}). We assume the PDE solution is vector valued, meaning that $u_{\theta}$ takes values in $\mathbb{R}^{N_v}$. \\
        \addlinespace[0.4em]
        $\Theta$ & The set of all possible parameter ($\theta$) values. \\
        \addlinespace[0.4em]
        $F$ & Given some $\theta \in \Theta$, $F\left( \cdot, \theta \right)$ is a non-linear partial derivative operator that represents the right-hand side of the FOM's PDE. See equation \eqref{eq:PDE}. \\
        \addlinespace[0.4em] 
        $u_0^{(k)}$ & A real valued function on $\Omega \times \Theta$ that defines the initial condition for the $k$'th time derivative of the FOM. We write $u_0$ in place of $u_0^{(0)}$. See equation \eqref{eq:IC}. \\
        \addlinespace[0.4em] 
        $N_u$ & The number of spatial nodes in a numerical solution to the PDE. We assume this is the same for all $\theta$ values. \\
        \addlinespace[0.4em] 
        $D_t^k \vec{u}_{\theta}(t_j^{\theta})$ & An element of $\mathbb{R}^{N_u}$ representing the $k$'th time derivative of the numerical solution to the FOM at time $t_j^{\theta} \in [0, T]$ when the FOM uses parameter value $\theta$. We write $\vec{u}_{\theta}(t_j^{\theta})$ in place of $D_t^0 \vec{u}_{\theta}(t_j^{\theta})$. \\
        \bottomrule[0.3ex]
    \end{tabulary}
    
    \caption{The notation and terminology of section (\ref{Problem}).} 
    \label{Table:Problem:Notation}.
\end{table}
\section{Background}
\label{Background}

Higher-Order LaSDI builds upon GPLaSDI \cite{bonneville2024gplasdi, bonneville2023gplasdi_neuralips} and Rollout-LaSDI \cite{stephany2025rollout}, both of which use autoencoder-based Reduced Order Modeling to address the problem outlined in section \ref{Problem}.
In this section, we introduce the autoencoder-based Reduced Order Modeling, then discuss GPLaSDI and Rollout-LaSDI.
In this section, we specialize to the case $K = 1$, as most existing ROM frameworks make this assumption.
In the next section, we will extend the ideas presented here to arbitrary $K$.

\subsection{Reduced Order Modeling}
\label{Background:ROMs}

\begin{figure}
    \centering
    \includegraphics[width=\linewidth, trim=100 150 100 150]{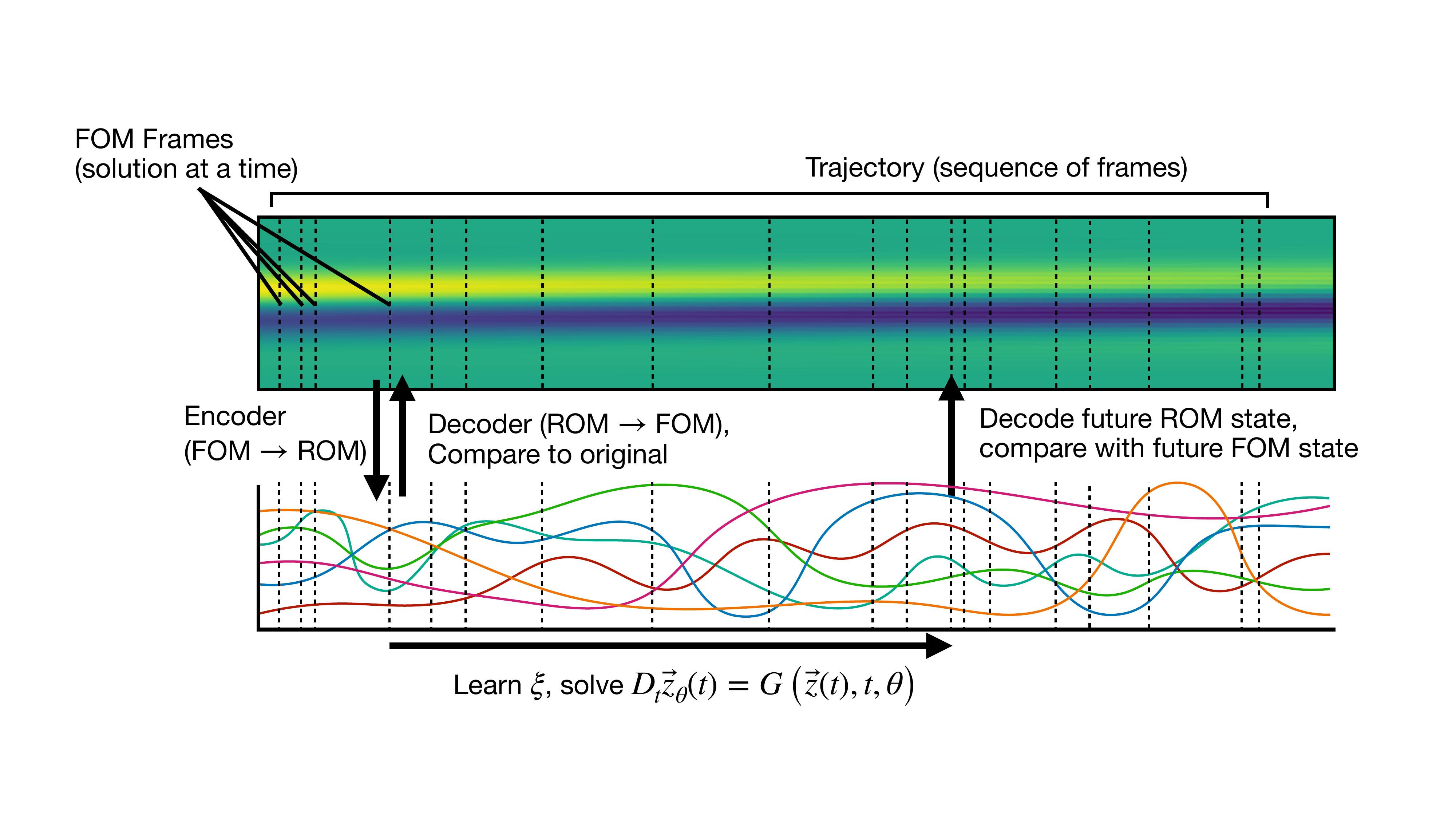}
    \caption{A schematic for a parameterized ROM. A ROM can be used to predict the future FOM state by a) encoding the initial FOM state, b) using the latent dynamics (denoted here as $G$) to integrate the initial latent state, and c) decoding the final latent state to predict the future FOM state.}
    \label{fig:ROM}
\end{figure}

\emph{Reduced Order Modeling} addresses the problem in section \ref{Problem} by training a surrogate model that approximates the FOM's evolution.
A \emph{Reduced Order Model (ROM)} consists of three components:
\begin{enumerate}
    \item An encoder $\varphi_e : \mathbb{R}^{N_u} \to \mathbb{R}^L$ that maps a \emph{frame} $\vec{u}_{\theta}(t) \in \mathbb{R}^{N_u}$ from the numerical FOM solution to a low-dimensional latent representation $\varphi_e\left( \vec{u}_{\theta}(t) \right) \in \mathbb{R}^L$. 
    \item Latent dynamics that map parameter values $\theta$ to an associated system of ODEs describing the evolution of latent encodings for that parameter.
    \item A decoder $\varphi_d : \mathbb{R}^L \to \mathbb{R}^{N_u}$ that maps latent space elements to the discretized spatial domain $\mathbb{R}^{N_u}$.
\end{enumerate}
We call $\mathbb{R}^L$ the \emph{Latent Space} and $\varphi_e \left( \hat{u}_{\theta}(t) \right)$ the \emph{latent encoding} of frame $\hat{u}_{\theta}(t)$.
We assume $L \ll N_u$, so the latent space has much smaller dimension than the discretized spatial domain.
In practice, the encoder, decoder, and latent dynamics are differentiable parameterized functions trained via supervised learning on available training data (see section \ref{Background:Autoencoder}).
Figure \ref{fig:ROM} gives a schematic for a basic, parameterized ROM.

$$\nabla_{\theta} v_{\pi(\cdot|\cdot, \theta)}(s_0) = \sum_{s \in S} \sum_{t \in \mathbb{N}\cup 0} \sum_{a \in A(s)} p(S_t = s) q_{\pi(\cdot|\cdot, \theta}(s, a) \gamma^t \nabla_{\theta} \pi(a|s, \theta)$$

The Reduced Order Modeling approach relies on the \emph{Manifold Hypothesis} \cite{gorban2018manifold_hypothesis}, which asserts that high-dimensional real-world data lies on or near a low-dimensional manifold $M$ (empirically validated across countless applications).
We assume each $\vec{u}_{\theta}\left(t^{\theta}_j \right)$ lies close to $M$, making the high-dimensional data effectively low-dimensional.
If the encoder and decoder approximate the manifold's chart and its inverse, they can absorb much of the dataset's complexity, leaving relatively simple latent dynamics to learn.
To achieve this, we train the encoder-decoder pair to satisfy $\varphi_d \circ \varphi_e|_{M} \approx \mathrm{Id}_{M}$ by minimizing the normalized mean absolute error on the training set, 
\begin{equation}
    \mathcal{L}_{\text{Recon}}\left( \varphi_e, \varphi_d \right) = \frac{1}{N_t} \sum_{i = 1}^{N_{\theta}} \frac{1}{\sigma_{\theta_i}} \sum_{j = 0}^{N_t(\theta_i)} \left\| \vec{u}_{\theta_i}\left(t_{j}^{\theta_i}\right) - \left( \varphi_d \circ \varphi_e \right)\left( \vec{u}_{\theta_i}\left(t_{j}^{\theta_i}\right) \right) \right\|_1, \quad N_t = \sum_{i = 1}^{N_{\theta}} N_t(\theta_i),
\label{eq:loss:recon}
\end{equation}
where 
\begin{equation*}
    \sigma_{\theta_i} = \text{STD}\left( \vec{u}_{\theta_i}\left(t_j^{\theta_i}\right) \cdot e_r\ \Big|\ r \in \left\{ 1, 2, \ldots, N_u \right\},\ j \in \left\{ 1, 2, \ldots, N_t(\theta_i) \right\} \right) 
\end{equation*} 
is the standard deviation of the numerical FOM solution components for $\theta_i$.
$\sigma_{\theta_i}$ normalizes the loss across parameter values, preventing the model from ignoring FOM solutions with smaller variance.
The encoder maps each element of a numerical FOM solution to a latent time series,
\begin{equation}
    \varphi_e \left( \vec{u}_{\theta}\left(t^{\theta}_1\right) \right) \ldots, \varphi_e \left( \vec{u}_{\theta}\left(t^{\theta}_{N_t(\theta)} \right) \right).
\end{equation}

\subsection{Autoencoder-based Reduced Order Modeling}
\label{Background:Autoencoder}

Many modern Reduced Order Modeling frameworks \cite{lee2021parameterized_NODE, rubanova2019latent_NODE, bonneville2024gplasdi, bonneville2023gplasdi_neuralips, he2023glasdi, tran2024wlasdi, park2024tlasdi, bonneville2024lasdi_review, chung2025LaSDI_it, anderson2025mlasdi, stephany2025rollout, he2025wgLaSDI} use an autoencoder as the encoder-decoder pair, where both components are deep neural networks.
Here, we explore two pioneering approaches: SINDy Autoencoder and the Latent Space Identification of Dynamics.

\textbf{SINDy Autoencoder:} Champion et al. \cite{champion2019sindy_autoencoder} proposed Autoencoder-SINDy, which uses an autoencoder and linear latent dynamics to approximate PDE solution evolution.
This approach operates in the non-parameterized regime: $\Theta$ contains a single element $\theta$ with corresponding numerical solution $\left\{ \vec{u}_{\theta}\left( t_j \right) \right\}_{j = 0}^{N_t(\theta)}$.
The latent dynamics associate $\theta$ with the \emph{latent initial value problem (Latent IVP)}:
\begin{equation}
\begin{aligned}
    \dot{\vec{z}}(t) &= A \vec{z}(t) + b,\\
    \vec{z}(0) &= z_0,
\end{aligned}
\label{eq:Latent:IVP:Linear}
\end{equation}
where $\vec{z}(t) \in \mathbb{R}^L$ represents the latent state at time $t \in [0, T]$, $A \in \mathbb{R}^{L \times L}$ is a matrix, and $b \in \mathbb{R}^L$ is a vector.
To learn $A$ and $b$, SINDy Autoencoder encodes $\left\{ \vec{u}_{\theta}\left( t_j \right) \right\}_{j = 0}^{N_t(\theta)}$ to obtain $\left\{ \varphi_e \left( \vec{u}_{\theta}\left( t_j \right) \right) \right\}_{j = 0}^{N_t(\theta)}$, which is assumed to be frames of a solution to \eqref{eq:Latent:IVP:Linear} for some $A, b$. 
The algorithm uses this encoded sequence, finite differences, the FOM time series, and encoder/decoder gradients to approximate both sides of \eqref{eq:Latent:IVP:Linear}.
This yields SINDy-based loss functions \cite{brunton2016sindy} whose minimization determines $A$ and $b$.
See \cite{champion2019sindy_autoencoder} for details.

After training, the learned latent dynamics predict the future FOM state:
\begin{enumerate}
    \item Start with the numerical FOM solution $\vec{u}_{\theta} \left( t \right)$ at time $t$.
    \item Encode $\vec{u}_{\theta} \left( t \right)$ to obtain $\vec{z}(t) = \varphi_e \left( \vec{u}_{\theta} \left( t \right) \right)$ and use it as the initial condition for the latent IVP \eqref{eq:Latent:IVP:Linear}. 
    \item Integrate \eqref{eq:Latent:IVP:Linear} to obtain a prediction for the future latent state $\hat{z}(t + \tau)$. 
    \item Decode $\hat{z}(t + \tau)$ to obtain $\varphi_d \left( \hat{z}(t + \tau) \right)$, which approximates $\vec{u}_{\theta}(t + \tau)$. 
\end{enumerate}

\textbf{LaSDI:} The \emph{Latent Space Identification of Dynamics (LaSDI)} framework extends SINDy Autoencoder to the parameterized regime.
Like SINDy Autoencoder, LaSDI uses an autoencoder for the encoder/decoder.
LaSDI uses the same latent dynamics model as SINDy Autoencoder, but allows $A$ and $b$ to vary with $\theta$, assuming a learnable map $\theta \to A_\theta, b_\theta$. 
$A_{\theta}$, $b_{\theta}$ are the \emph{latent coefficients} for $\theta$.
HLaSDI uses the same latent dynamics model.
For each training parameter, $\theta_i$, LaSDI finds $A_{\theta_i}$, $b_{\theta_i}$ by encoding the FOM sequence
$$
\left\{ \vec{u}_{\theta_i} \left( t_j^{\theta_i} \right) \right\}_{j = 0}^{N_t(\theta_i)}
$$
to obtain
$$
\left\{ \varphi_e \left( \vec{u}_{\theta_i} \left( t_j^{\theta_i} \right) \right) \right\}_{j = 0}^{N_t(\theta_i)} = \left\{ \vec{z}_{\theta_i}\left( t_j^{\theta_i} \right) \right\}_{j = 0}^{N_t(\theta_i)}.
$$
LaSDI then uses finite differences to approximate $\left\{ \dot{\vec{z}}_{\theta_i}\left( t_j^{\theta_i} \right) \right\}_{j = 0}^{N_t(\theta_i)}$.

To learn $A_{\theta_i}$ and $b_{\theta_i}$, LaSDI solves the least squares problem
\begin{equation}
    \text{argmin}_{(A_{\theta_i}, b_{\theta_i})} \left\| A_{\theta_i} Z + b_{\theta_i} \mathbb{1} - D \right\|_{2}^2,
    \label{eq:Latent:LSQ}
\end{equation}
where $\mathbb{1} \in \mathbb{R}^{1 \times N_t(\theta_i)}$ is all ones, $D \in \mathbb{R}^{L \times N_u}$ has $j$-th column $\dot{\vec{z}}_{\theta_i}\left( t_j^{\theta_i} \right)$, and $Z \in \mathbb{R}^{L \times N_t(\theta_i)}$ has $j$-th column $\vec{z}_{\theta_i}\left( t_j^{\theta_i} \right)$.
LaSDI uses a differentiable least-squares solver to find $A_{\theta_i}, b_{\theta_i}$ and the corresponding residual, producing the collection $\left\{ A_{\theta_i}, b_{\theta_i} \right\}_{i = 1}^{N_{\theta}}$.
Since the vectors building $Z$ are FOM encodings, the residual is differentiable with respect to the encoder's parameters, so minimizing it optimizes the encoder to produce encodings that best satisfy the latent dynamics.
LaSDI minimizes the average least squares residual across training parameters (the \emph{Latent Dynamics loss}) alongside the reconstruction loss, equation \eqref{eq:loss:recon}.

After training, LaSDI uses the trained encoder-decoder pair to encode the training set and compute final latent coefficients $\left\{ A_{\theta_i}, b_{\theta_i} \right\}_{i = 1}^{N_\theta}$.
During inference, LaSDI defines the latent dynamics map $\theta \to A_\theta, b_\theta$ by interpolating these coefficients using various methods across different LaSDI variants.
In subsection \ref{Background:GPLaSDI}, we describe the Gaussian Process variant that HLaSDI uses.
To approximate the FOM solution for arbitrary $\theta \in \Theta$, LaSDI:
\begin{enumerate}
    \item Evaluates the initial condition function $u_0$ at $\theta$ to obtain the FOM initial condition. 
    \item Encodes the initial condition to obtain the latent initial condition.
    \item Interpolates the learned latent coefficients at $\theta$ to obtain $A_{\theta}, b_{\theta}$. 
    \item Numerically integrates \eqref{eq:Latent:IVP:Linear} with $A_{\theta}, b_{\theta}$ and the latent initial condition, yielding predicted future latent states.
    \item Decodes the latent time series to predict the numerical FOM solution for $\theta$. 
\end{enumerate}
This approach enables LaSDI to approximate solutions across a parameterized family of PDEs.

Several LaSDI extensions exist.
Some endow the latent dynamics with specific properties \cite{park2024tlasdi, chung2025LaSDI_it} or harden it to noise \cite{tran2024wlasdi, he2025wgLaSDI}. 
One recent extension replaces the decoder with a sequence of decoders that progressively predict residuals \cite{anderson2025mlasdi}.
Other extensions introduce new interpolation methods for the latent dynamics.
Two notable examples are GLaSDI \cite{he2023glasdi} and \emph{GPLaSDI} \cite{bonneville2024gplasdi, bonneville2023gplasdi_neuralips}.
GLaSDI is \emph{intrusive} (assumes knowledge of the FOM) and leverages it to estimate ROM performance across testing parameters, greedily adding the worst-performing parameters to the training set before re-training.
GPLaSDI is \emph{non-intrusive} (assumes no FOM knowledge) and uses \emph{Gaussian processes (GPs)} \cite{williams2006gaussian} to add parameters and interpolate dynamics. 
We discuss GPLaSDI in the next section.

\subsection{GPLaSDI}
\label{Background:GPLaSDI}

GPLaSDI modifies LaSDI by using Gaussian Processes \cite{williams2006gaussian} to learn the latent dynamics through a sequence of training episodes.
It begins with an initial training set
\begin{equation*}
    \left\{ \left\{ \vec{u}_{\theta_i}\left( t_{j}^{\theta_i} \right) \right\}_{j = 0}^{N_t(\theta_i)} \right\}_{i = 1}^{N_{\theta}},
\end{equation*}
and testing parameters
\begin{equation*}
 \left\{ \theta_i \right\}_{i = N_{\theta} + 1}^{N_{\theta} + N_{\theta}(\text{test})}.   
\end{equation*}
GPLaSDI finds the latent coefficients using the procedure in section \ref{Background:Autoencoder}.

After initial training, GPLaSDI trains $L(L + 1)$ Gaussian Processes (one per component: $L^2$ for $A$ and $L$ for $b$) that map parameter values to latent coefficient components.
GPLaSDI trains these Gaussian Processes using the final latent coefficients $\left\{ A_{\theta_i}, b_{\theta_i} \right\}_{i = 1}^{N_{\theta}}$.

After training these Gaussian Processes, GPLaSDI uses a greedy procedure to identify the worst-performing testing parameter through \emph{Greedy Sampling}:
\begin{enumerate}
    \item Select a testing parameter $\theta_i$ and evaluate the FOM initial condition at $\theta_i$ using $u_0$. 
    \item Encode the initial condition for $\theta_i$ to the latent space. 
    \item Draw $N_{s} = 20$ samples from each Gaussian Process posterior conditioned on $\theta_i$. 
    \item For each sample, form the corresponding latent coefficients and numerically integrate the latent dynamics with the latent initial condition, generating $N_s$ latent time series for $\theta_i$. 
    \item Decode each latent time series to obtain $N_s$ predictions for the numerical FOM solution at $\theta_i$.
    \item Compute the variance across the $N_s$ samples for each component of each frame and store the maximum variance.
    \item Repeat for all testing parameters and select the parameter with largest variance. 
\end{enumerate}
Greedy sampling assumes that prediction variance is correlated with model uncertainty.
Thus, if the predicted solution for a particular testing parameter has a large variance, the ROM has a lot to learn from the corresponding numerical FOM solution.

\begin{figure}
    \centering
    \includegraphics[width=\linewidth, trim=0 200 0 200]{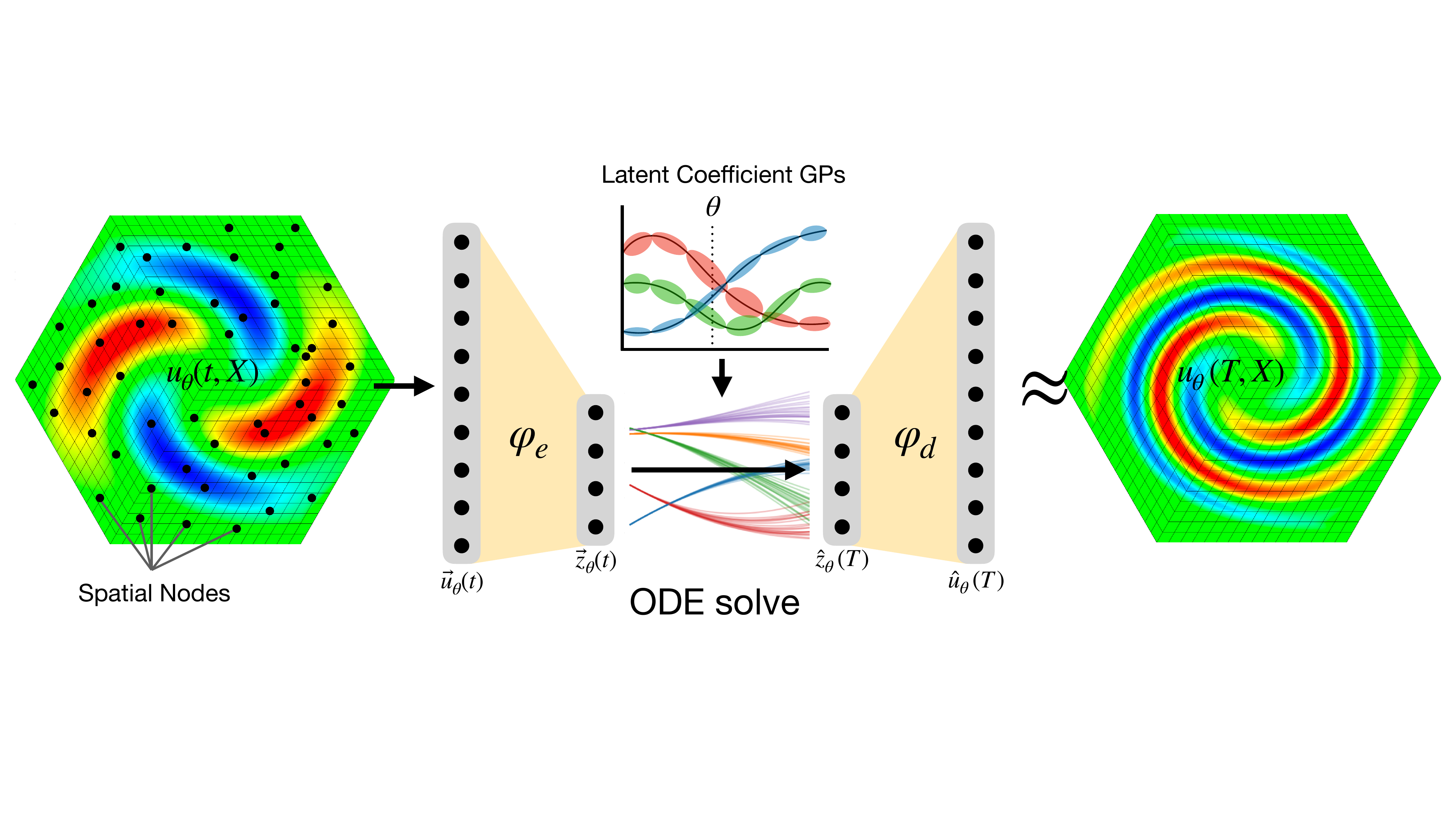}
    \caption{A schematic of Rollout. From left to right, we begin by encoding a discretization $\vec{u}_{\theta}(t)$ of the FOM solution, $u_{\theta}(t, X)$. This gives us the initial latent state, $\vec{z}_{\theta}(t)$. We sample the posterior distribution at $\theta$ of the gaussian processes for the latent dynamics coefficients to obtain a set of latent coefficients for parameter $\theta$. We use these samples to define the latent coefficients, yielding a dynamical system which we can integrate forward to obtain a prediction of the future latent state, $\hat{z}_{\theta}\left(t + \Delta t^{\theta} \right)$. We decode this to obtain $\hat{u}_{\theta}\left(t + \Delta t^{\theta} \right)$, a prediction of the future FOM state.}
    \label{fig:Inference}
\end{figure}

After greedy sampling, GPLaSDI generates the numerical FOM solution for the selected parameter and adds it to the training set. 
This cycle of training, greedy sampling, and solution generation constitutes an \emph{episode}.
GPLaSDI trains for several episodes until variance at all remaining testing parameters is sufficiently low.
After the final episode, GPLaSDI defines the latent dynamics using the trained Gaussian Processes: for any $\theta$, it uses the GP posterior mean at $\theta$ as the latent coefficients. 
Figure \ref{fig:Inference} gives a schematic of how LaSDI performs Inference.
%

GPLaSDI has proven remarkably effective at learning parameterized families of PDEs.
Because this procedure requires no FOM knowledge (only the ability to generate FOM solutions), it is \emph{non-intrusive}.
HLaSDI adopts GPLaSDI's episodic training procedure but uses different latent dynamics (see section \ref{Method}).

\subsection{Rollout LaSDI}
\label{Background:Rollout}

A primary limitation of GPLaSDI is poor performance on some FOMs over long time horizons, stemming from how it (and most reduced order modeling approaches) enforces latent dynamics.
The ROM learns latent dynamics that encoded states approximately satisfy at each time step, meaning it is trained only to \emph{instantaneously} satisfy these dynamics.
The training procedure does not directly incentivize accurate long-term predictions.
Even if the encoded time series approximately satisfies the latent ODE at each time step, small errors accumulate over long horizons, so the latent ODE solution may diverge significantly from the encoded FOM sequence. 
Thus, predicted FOM solutions may differ substantially from true solutions as $t$ increases, even when latent dynamics are locally correct.

\begin{figure}
    \centering
    \includegraphics[width=\linewidth, trim=0 200 0 200]{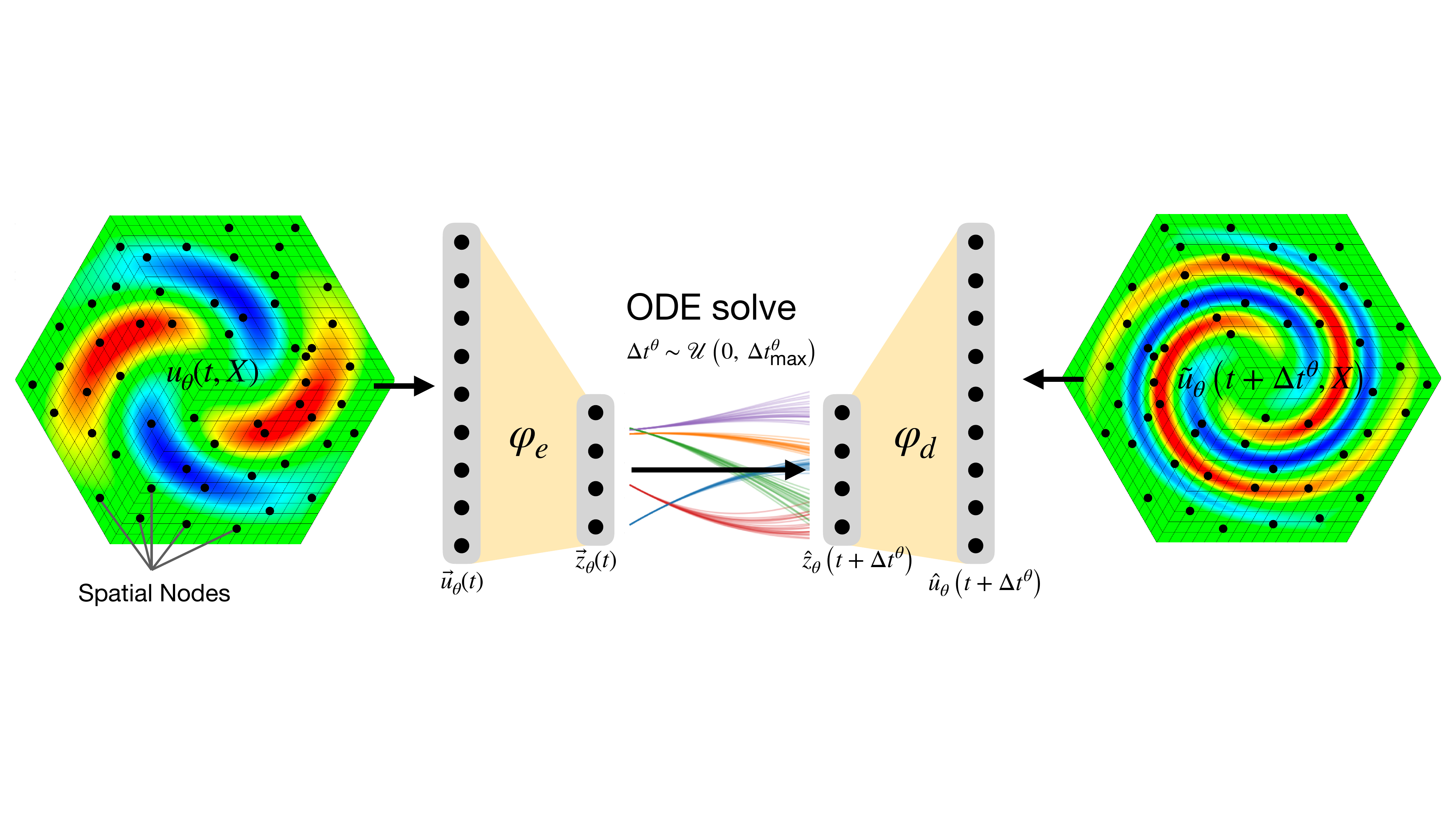}
    \caption{A schematic of Rollout. From left to right, we begin by encoding a discretization $\vec{u}_{\theta}(t)$ of the FOM solution, $u_{\theta}(t, X)$. This gives us the initial latent state, $\vec{z}_{\theta}(t)$. We use the latent dynamics for $\theta$ to integrate the initial latent state and obtain a prediction of the future latent state, $\hat{z}_{\theta}\left(t + \Delta t^{\theta} \right)$. We decode this to obtain $\hat{u}_{\theta}\left(t + \Delta t^{\theta} \right)$, a prediction of the future FOM state, which we compare to an interpolation $\tilde{u}_{\theta}\left( t + \Delta t^{\theta} \right)$ of the future FOM state.}
    \label{fig:Rollout}
\end{figure}

To achieve accurate long-horizon predictions, we should design loss functions that encode this behavior.
Lusch et al. \cite{lusch2018koopman} first attempted this for Koopman operators.
Rollout-LaSDI \cite{stephany2025rollout} extended this to the LaSDI framework by augmenting GPLaSDI with a \emph{Rollout-Loss}.
For each parameter $\theta$, Rollout-LaSDI chooses a maximum time horizon $\Delta t_{\max}^{\theta}$ that increases from near zero at training start to near $T$ at training end (see section \ref{Discussion} for details).
A frame $\vec{u}_{\theta}(t^{\theta})$ is \emph{rollable} if $t^{\theta} + \Delta t_{\max}^{\theta} \le T$.
Rollout-LaSDI encodes each rollable frame $\vec{u}_{\theta}(t)$ to obtain latent encoding $\vec{z}_{\theta}(t)$, which serves as the initial condition for the latent ODE \eqref{eq:Latent:IVP:Linear}.
It integrates the latent dynamics using a differentiable RK4 solver to a randomly sampled time $t + \Delta t^{\theta}$, where $\Delta t^{\theta} \sim \mathcal{U}\big(0,\Delta t_{\max}^{\theta}\big)$, with simulation times independently sampled per rollable frame and re-sampled each epoch.
Rollout-LaSDI decodes the final latent state $\hat{z}_{\theta}(t + \Delta t^{\theta})$ to obtain 
\begin{equation*}
    \hat{u}_{\theta}(t + \Delta t^{\theta}) = \varphi_d \big(\hat{z}_{\theta}(t + \Delta t^{\theta})\big),
\end{equation*}
a prediction of the future FOM state.
Let $\tilde{u}_{\theta}\left(t + \Delta t^{\theta} \right)$ denote an interpolation of the numerical FOM solution evaluated at $t+ \Delta_t^{\theta}$.
Rollout-LaSDI computes the $L^1$ norm of the difference between $\hat{u}_{\theta}(t + \Delta t^{\theta})$ and $\tilde{u}_{\theta}(t + \Delta t^{\theta})$.
Averaging across all rollable frames yields the \emph{Rollout-Loss}:
\begin{equation}
    \mathcal{L}_{\text{Rollout}}\left( \varphi_e, \varphi_d, \left\{ A_{\theta_i}, b_{\theta_i} \right\}_{i = 1}^{N_{\theta}} \right) = \frac{1}{N_{ro}} \sum_{i = 1}^{N_{\theta}} \sum_{j = 0}^{N_{ro}(\theta_i)} \left\|  \tilde{u}_{\theta_i} \left(t_{j}^{\theta_i} + \Delta t_{\text{ro}}^{\theta_i}(j) \right) - \hat{u}_{\theta_i}\left(t_j^{\theta_i} + \Delta t_{ro}^{\theta_i}(j) \right) \right\|_1,
\label{eq:loss:rollout}
\end{equation}
where $N_{ro}(\theta_i)$ is the number of rollable frames for $\theta_i$ and $N_{ro} = \sum_{i = 1}^{N_{\theta}} N_{ro}(\theta_i)$. 
Minimizing this loss yields a ROM that predicts future FOM solutions over arbitrary time horizons, significantly improving GPLaSDI's long-horizon accuracy.
Figure \ref{fig:Rollout} gives a schematic of Rollout.
See \cite{stephany2025rollout} for details.
%

However, Rollout-LaSDI only operates on FOMs with a single time derivative. 
Higher-Order LaSDI extends Rollout-LaSDI to multiple time derivatives while introducing enhancements that improve performance on complex physical systems (see section \ref{Method}).

Table \ref{Table:Background:Notation} lists the notation we introduced in this section.

\begin{table}[hbt]
    \centering 
    \rowcolors{2}{white}{cyan!10}
    
    \begin{tabulary}{1.0\linewidth}{p{1.5cm}L}
        \toprule[0.3ex]
        \textbf{Notation} & \textbf{Meaning} \\
        \midrule[0.1ex]
        $L$ & The dimension of the ROM's latent space. \\
        \addlinespace[0.4em]
        $\varphi_e$ & The ROM's encoder. This is a map from $\mathbb{R}^{N_u}$ to $\mathbb{R}^L$ which sends frames in the numerical FOM solution to low-dimensional latent encodings. In general, this is a parameterized, trainable map. \\
        \addlinespace[0.4em]
        $\varphi_d$ & The ROM's decoder. This is a map from $\mathbb{R}^{L}$ to $\mathbb{R}^{N_u}$ which acts somewhat like the inverse of $\varphi_e$, sending encoded frames back to (an approximation of) their original value. In general, this is a parameterized, trainable map. \\
        \addlinespace[0.4em]
        $\sigma_{\theta_i}$ & The standard deviation of the components of the numerical FOM solution for $\theta_i$. That is, $\sigma_{\theta_i} = \text{STD}\left( \vec{u}_{\theta_i}\left(t_j^{\theta_i}\right) \cdot e_r\ \Big|\ r \in \left\{ 1, 2, \ldots, N_u \right\},\ j \in \left\{ 1, 2, \ldots, N_t(\theta_i) \right\} \right)$  \\
        \addlinespace[0.4em]
        $N_s$ & The number of samples we draw from the posterior distribution of the trained Gaussian Processes when performing greedy sampling. See section \ref{Background:GPLaSDI}. \\
        \addlinespace[0.4em]
        $\vec{z}_{\theta}(t)$ & The latent state at time $t$ when the parameter value is $\theta$. $D_t^{(k)} \vec{z}_{\theta}(t)$ is the encoding of $D_t^{(k)} \vec{u}_{\theta}(t)$. \\
        \addlinespace[0.4em]
        $\hat{u}_{\theta}(t)$ & The predicted future latent state we get when we integrate the latent dynamics forward in time. \\
        \bottomrule[0.3ex]
    \end{tabulary}
    
    \caption{The notation and terminology of section (\ref{Problem}).}
    \label{Table:Background:Notation}.
\end{table}
\section{Methodology}
\label{Method}

Higher-Order LaSDI extends Rollout-LaSDI to FOMs with multiple time derivatives.
In this section, we let $K$ (the highest time derivative in the FOM) be arbitrary.
For each training parameter $\theta$, the numerical FOM solution consists of $K$ time series:
\begin{equation}
    \left\{ D_t^{(k)} \vec{u}\left(t_j^{\theta_i} \right) \right\}_{j = 0}^{N_t(\theta)}, \qquad k \in \{ 0, 1, \ldots, K - 1 \}
\end{equation}
%

HLaSDI's encoder-decoder consists of $K$ autoencoders.
For each $k \in \left\{ 0, 1, \ldots, K - 1 \right\}$, let $\varphi_e^{(k)} : \mathbb{R}^{N_u} \to \mathbb{R}^L$ and $\varphi_d^{(k)} : \mathbb{R}^L \to \mathbb{R}^{N_u}$ denote the encoder and decoder for the $k$-th time derivative, respectively.
We train $\varphi_e^{(k)}$, $\varphi_{d}^{(k)}$ to minimize $\mathcal{L}_{\text{Recon}}\left( \varphi_e^{(k)}, \varphi_d^{(k)} \right)$ (see equation \eqref{eq:loss:recon}), encouraging the encoder-decoder pair to approximate $\varphi_d^{(k)} \circ \varphi_e^{(k)} \approx \text{Id}$ on the set of $k$-th time derivatives,
\begin{equation*}
    \left\{ D_t^{(k)} \vec{u}\left( t_j^{\theta_i} \right) : i \in \left\{ 1, 2, \ldots, N_{\theta} \right\},\ j \in \left\{ 1, 2, \ldots, N_t\left(\theta_i\right) \right\} \right\}.
\end{equation*}
%

Each autoencoder has its own latent space.
HLaSDI unifies these latent spaces through a common set of latent dynamics.
Consider the following general order-$K$ dynamical system:
\begin{align}
    D_t^{(K)} \vec{z}_{\theta}(t) &= C^{(K - 1)}_{\theta} D_t^{(K - 1)}\vec{z}_{\theta}(t) + \cdots + C^0_{\theta} \vec{z}_{\theta}(t) + b_{\theta}, \qquad t \in [0, T] \\
    D_t^{(k)} \vec{z}_{\theta}(0) &= z_0^{(k)}\left( \theta \right),
\end{align}
where $C_{\theta}^{(K - 1)}, \ldots, C_{\theta}^{0} \in \mathbb{R}^{L \times L}$ are parameter-specific matrices and $b_{\theta} \in \mathbb{R}^L$ is a parameter-specific vector.
Since the $k$-th autoencoder operates on the $k$-th time derivative of the numerical FOM solution, its latent encoding should approximate the $k$-th time derivative of the latent state: $D_t^{(k)} z(t) \approx \varphi_e^{(k)}\left( D_t^{(k)} \vec{u}(t) \right)$.
Motivated by this, HLaSDI's latent dynamics are 
\begin{align}
    D_t^{(K)} \vec{z}_{\theta}(t) &= C^{(K - 1)}_{\theta} \varphi_e^{(K - 1)}\left( D_t^{(K - 1)} \vec{u}_{\theta}(t))\right) + \cdots + C^{(0)}_{\theta}  \varphi_e^{(0)}\left(\vec{u}_{\theta}(t)\right) + b_{\theta}, \quad && t \in [0, T] \label{eq:Latent:ODE:higher} \\
    D_t^{(k)} \vec{z}_{\theta}(0) &= \varphi_e^{(k)}\left( \vec{u}_0^{(k)}\left( \cdot,\theta \right) \right), \quad && k \in \left\{ 0, 1, \ldots, K - 1 \right\}. \label{eq:Latent:IC:higher}
\end{align}
Thus, $C^{(K - 1)}_{\theta}, \ldots, C^{(0)}_{\theta}$ and $b_{\theta}$ are latent coefficients.
The latent dynamics form a coupled system of order-$K$ ODEs, which HLaSDI integrates using a differentiable RK-4 solver.
Figure 
\begin{figure}
    \centering
    \includegraphics[width=\linewidth, trim=0 150 0 150]{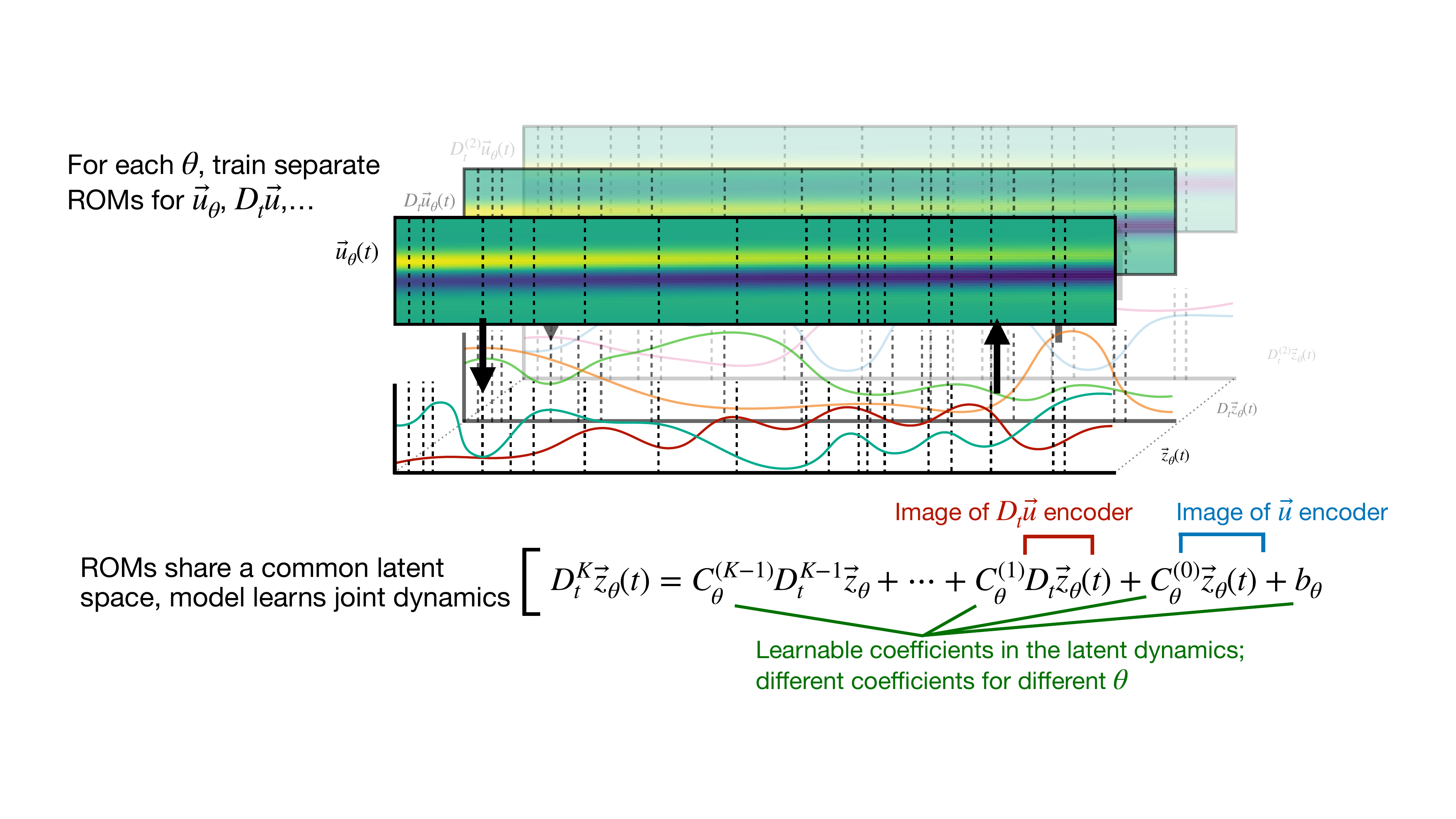}
    \caption{A schematic of HLaSDI's architecture, consisting of $K$ autoencoders with a unified set of latent dynamics. The $k$'th encoder, $\varphi_e^{(k)}$ encodes snapshots of the $k$'th time derivative of the FOM solution. Each Autoencoder maps to the same latent space, and we assume that their latent states are governed by a common set of latent dynamics, equation \eqref{eq:Latent:ODE:higher}.}
    \label{fig:Many_ROMs}
\end{figure}
%

During inference, HLaSDI uses GPLaSDI's Gaussian Process approach to interpolate the latent coefficients.
Since the latent dynamics have $L\left( K L + 1 \right)$ components, HLaSDI trains $L(K L + 1)$ Gaussian processes during each greedy sampling step.

Unlike GPLaSDI and Rollout-LaSDI, HLaSDI treats the coefficients as \emph{trainable parameters} learned directly with the encoder-decoder parameters.
Each training parameter has a distinct set of trainable latent coefficients, all zero-initialized.

HLaSDI learns the latent coefficients by minimizing the \emph{Rollout Loss}, \emph{Initial Condition Rollout Loss}, and \emph{Latent Dynamics Loss}.
We define the Rollout losses in section \ref{Method:Rollout}.
To compute the Latent Dynamics loss, for each testing parameter $\theta_i$ and time value $t_j$, we evaluate both sides of \eqref{eq:Latent:IC:higher} at $t_j$ with parameter $ \theta_i$ and compute the $\ell^1$ norm of their difference:
\begin{equation}
\begin{aligned}
    \mathcal{L}_{LD}&\Big(\varphi_e^{(0)}, \ldots, \varphi_e^{(K - 1)}, \left\{ b_{\theta_i}, C_{\theta_i}^{(0)}, \ldots,  C_{\theta_i}^{(K - 1)} \right\}_{i = 1}^{N_{\theta}} \Big) \\
    &= \sum_{i = 1}^{N_{\theta}} \sum_{j = 0}^{N_t(\theta_i)} \left\| D_t^{(K)} z_{\theta_i}(t_j) - \left( C^{(K - 1)}_{\theta_i} \varphi_e^{(K - 1)}\left( D_t^{(K - 1)} \vec{u}_{\theta_i}(t_j)\right) + \cdots + C^{(0)}_{\theta_i}  \varphi_e^{(0)}\left(\vec{u}_{\theta_i}(t_j)\right) + b_{\theta_i} \right) \right\|_1
\end{aligned}
\label{eq:loss:LD}
\end{equation}
%

Evaluating this loss requires approximating $D_t^{(K)} z_{\theta_i}(t_j)$. 
Unlike the $K = 1$ case, finite differences do not provide a unique approach.
Since $\varphi_e^{(k)}$'s encodings proxy the $k$-th time derivative of the latent state $\vec{z}_{\theta}$, $D_t^{(K)} \vec{z}_{\theta}$ should equal the $(K - k)$-th time derivative of $\varphi_e^{(k)} \left( D_t^{(k)}\vec{u}_{\theta} \right)$.
This provides multiple ways to compute $D_t^{(K)} \vec{z}_{\theta}$; any affine combination of these finite differences is a valid approximation of $D_t^{(K)} z_{\theta_i}(t_j)$.
We use a convex combination of the first time derivative of $\varphi_e^{(K - 1)} \left( D_t^{(K - 1)}\vec{u}_{\theta} \right)$ and the second time derivative of $\varphi_e^{(K - 2)} \left( D_t^{(K - 2)}\vec{u}_{\theta} \right)$.
See section \ref{Discussion:DtK_Z} for further discussion.
We use specialized finite-difference schemes for these derivatives, discussed in section \ref{Method:Finite_Diff}.

\subsection{Finite-Differences for nonuniform time series}
\label{Method:Finite_Diff}

Like Rollout-LaSDI, HLaSDI does not assume uniform time stepping in $\{ t_j^{\theta} \}_{j = 0}^{N_t(\theta)}$, enabling it to handle irregularly sampled data and adaptive time stepping.
However, variable time steps preclude conventional higher-order finite-difference schemes.
In Section \ref{Appendix:Finite_Difference}, we outline a general framework for deriving high-accuracy finite-difference schemes to approximate higher-order time derivatives of irregularly sampled time series.
Here, we introduce $\mathcal{O}(h^2)$ finite-difference schemes for first and second time derivatives, deriving them and proving their asymptotic convergence rates in Appendix \ref{Appendix:Finite_Difference}.
We use these schemes to compute the \emph{Rollout} and \emph{Consistency} losses.

\subsubsection{First-time derivatives}
\label{Method:Finite_Diff:First}

Let $S \subseteq \mathbb{R}$ be an open subset of $\mathbb{R}$ and let $f : S \to \mathbb{R}$ be of class $C^3$ on $S$.
Let $x \in S$, $a, b > 0$, and $h = \max\{a, b\}$.

\textbf{Forward-Difference:} Let us begin with a finite-difference scheme for first-time derivatives.
Suppose we know $f(x)$, $f(x + a)$, and $f(x + a + b)$ with $[x, x + a + b] \subseteq S$.
We seek $c_0, c_1, c_2 \in \mathbb{R}$ such that
\begin{equation}
    c_0 f(x) + c_1 f(x + a) + c_2 f(x + a + b) = f'(x) + \mathcal{O}(h^2).
    \label{eq:Finite_Diff:First:Forward}
\end{equation}
In section \ref{Appendix:Finite_Difference:First:Forward}, we show that if we show that if $\tfrac{a}{b}$ is bounded above, then
\begin{equation}
\begin{aligned}
    c_0 &= -\frac{2a + b}{a(a + b)}, \\
    c_1 &= \frac{a + b}{ab}, \\
    c_2 &= -\frac{a}{b(a + b)}
\end{aligned}
\label{eq:Finite_Diff:First:Forward:Coefficients}
\end{equation}
satisfy equation \eqref{eq:Finite_Diff:First:Forward}.
Further, when $a = b = h$, $c_0 = -\tfrac{3}{2h}$, $c_1 = \tfrac{2}{h}$, and $c = -\tfrac{1}{2h}$, which is the standard three-point forward-difference approximation to $f'(x)$ \cite{burden1997numerical}.

\textbf{Central-Difference:} Next, we consider a central difference scheme. 
Suppose we know $f(x - a)$, $f(x)$, and $f(x + b)$ and that $[x - a, x + b] \subseteq S$.
We seek $c_{-1}, c_0, c_1 \in \mathbb{R}$ such that
\begin{equation}
    c_{-1} f(x - a) + c_0 f(x) + c_1 f(x + b) = f'(x) + \mathcal{O}(h^2).
    \label{eq:Finite_Diff:Second:Central}
\end{equation}
In section \ref{Appendix:Finite_Difference:First:Central}, we show that this problem has a unique solution,
\begin{equation}
\begin{aligned}
    c_{-1}  &= -\frac{b}{a(a + b)}, \\
    c_0     &= \frac{b - a}{ab}, \\
    c_1     &= \frac{a}{b(a + b)}.
\end{aligned}
\label{eq:Finite_Diff:First:Central:Coefficients}
\end{equation}
When $a = b = h$, $c_{-1} = -\tfrac{1}{2h}$, $c_0 = 0$, and $c_1 = \frac{1}{2h}$, the coefficients for the standard central difference approximation of $f'(x)$ \cite{burden1997numerical}.

\textbf{Backwards-difference:} Finally, let us consider a backwards-difference scheme. 
To do this, suppose we know $f(x)$, $f(x - a)$ and $f(x - a - b)$ and that $[x - a - b, x] \subseteq S$.
We want to find $c_{-2}, c_{-1}, c_0 \in \mathbb{R}$ such that
\begin{equation}
    c_{-2} f(x - a - b) + c_{-1} f(x - a) + c_0 f(x) = f'(x) + \mathcal{O}(h^2)
    \label{eq:Finite_Diff:First:Backward}
\end{equation}
Let $g(x) = f(-x)$. 
Then $g' = -f'$, $g$ is of class $C^3$ on $S' = \{ -x : x \in S \}$, we know $g(-x)$, $g(-x + a)$, and $g(-x + a + b)$, and $[-x, -x + a + b] \subseteq S$.
Equation \ref{eq:Finite_Diff:First:Backward} then becomes
\begin{equation*}
    -c_{-2} g(-x + a + b) - c_{-1} g(-x + a) - c_0 g(-x) = g'(-x) + \mathcal{O}(h^2).
\end{equation*}
Using the forward-difference results (with $g$ in place of $f$, $S'$ in place of $S$, $-x$ in place of $x$, and $-c_{-2}, -c_{-1}, c_0$ in place of $c_2, c_1, c_0$), we can conclude that if $\tfrac{a}{b}$ is bounded above, then
\begin{equation}
\begin{aligned}
    c_{-2} &= \frac{a}{b(a + b)}, \\
    c_{-1} &= -\frac{a + b}{ab}, \\
    c_{0} &= \frac{2a + b}{a(a + b)}
    \label{eq:Finite_Diff:First:Backward:Coefficients}
\end{aligned}
\end{equation}
satisfy equation \eqref{eq:Finite_Diff:First:Backward}.

\subsubsection{Second-time derivatives}
\label{Method:Finite_Diff:Second}

Let $f : S \to \mathbb{R}$ be of class $C^4$ on $S$.
Let $a, b, c > 0$ and $h = \max\{a, b, c\}$.
We derive forward- and mixed-difference schemes for $f''(x)$.
Using the same approach we used in \ref{Method:Finite_Diff:First}, we can use these results to derive a backwards-difference scheme for $f''(x)$.

\textbf{Forward-Difference:} Suppose we know $f(x),\ f(x + a),\ f(x + a + b)$, and $f(x + a + b + c)$ with $[x, x + a + b + c] \subseteq S$.
We seek $c_0,\ c_1,\ c_2,\ c_3 \in \mathbb{R}$ such that 
\begin{equation}
    c_0 f(x) + c_1 f(x + a) + c_2 f(x + a + b) + c_3 f(x + a + b + c) = f''(x) + \mathcal{O}(h^2).
    \label{eq:Finite_Diff:Second:Forward}
\end{equation}
In section \ref{Appendix:Finite_Difference:Second:Forward}, we show that if $\tfrac{a}{b}, \tfrac{a}{c}, \tfrac{b}{c}$, and $\tfrac{2a + b}{b + c}$ are bounded above, then 
\begin{equation}
\begin{aligned}
    c_0 &= \frac{2(3a + 2b + c)}{a(a +b)(a + b + c)}, \\
    c_1 &= \frac{-2(2a + 2b + c)}{ab(b + c)}, \\
    c_2 &= \frac{ 2(2a + b + c)}{bc(a + b)}, \\
    c_3 &= \frac{-2(2a + b)}{(a + b + c)(b + c)c}
\end{aligned}
\label{eq:Finite_Diff:Second:Forward:Coefficients}
\end{equation}
satisfy equation \eqref{eq:Finite_Diff:Second:Forward}.
If $a = b = c = h$, $c_0 = \tfrac{2}{h^2}$, $c_1 = -\tfrac{5}{h^2}$, $c_2 = \tfrac{4}{h^2}$, and $c_3 = -\tfrac{1}{h^2}$, which are the standard forward-difference coefficients for $f''(x)$ \cite{burden1997numerical}.

\textbf{Mixed-Difference:} Suppose we know $f(x - a),\ f(x), f(x + b)$ and $f(x + b + c)$ with $[x - a, x + b + c] \subseteq S$.
We seek $c_{-1},\ c_0,\ c_1,\ c_2 \in \mathbb{R}$ such that 
\begin{equation}
    c_{-1} f(x - a) + c_0 f(x) + c_1 f(x + b) + c_2 f(x + b + c) = f''(x) + \mathcal{O}(h^2).
    \label{eq:Finite_Diff:Second:Mixed}
\end{equation}
In section \ref{Appendix:Finite_Difference:Second:Mixed}, we show that if $\tfrac{a}{b},\ \tfrac{c}{b},\ \tfrac{a}{c},\ \tfrac{b}{c}$ are bounded above, then
\begin{equation}
\begin{aligned}
    c_{-1}  &= \frac{2(2b + c)}{a(a + b)(a + b + c)}, \\
    c_0     &= \frac{-2\left(b(a + 2b + 3c) + c^2 - a^2\right)}{ba(b + c)(a + b + c)}, \\
    c_1     &= \frac{ 2(b + c - a)}{bc(a + b)}, \\
    c_2     &= \frac{-2(b - a)}{c(b + c)(a + b + c)}
\end{aligned}
\label{eq:Finite_Diff:Second:Mixed:Coefficients}
\end{equation}
satisfy equation \eqref{eq:Finite_Diff:Second:Mixed}.
Finally, if $a = b = c = h$, then $c_{-1} = \tfrac{1}{h^2}$, $c_0 = -\tfrac{2}{h^2}$, $c_1 = \tfrac{1}{h^2}$, and $c_2 = 0$, which is the standard $\mathcal{O}(h^2)$ central-difference scheme for $f''(x)$.

\subsection{Rollout Losses}
\label{Method:Rollout}

HLaSDI uses the Rollout and Initial Condition Rollout losses to help learn the latent coefficients by encoding the mathematical structure of the latent dynamics into our loss.
Because the latent dynamics should encode the evolution of the FOM, we should be able to use them to predict future FOM states and to predict how the FOM will evolve when starting from a specific initial condition.
The Rollout Loss trains our model to integrate a FOM frame forward by an arbitrary amount, while the Initial Condition Rollout Loss trains our model to integrate the FOM's initial condition along a specific solution trajectory.

\textbf{Rollout Loss:} HLaSDI uses a modification of Rollout-LaSDI's \cite{stephany2025rollout} Rollout loss.
For each training parameter $\theta_i$, we encode each rollable frame and use each encoding as the initial condition for the latent dynamics.
This yields a collection of IVPs, each integrated using a differentiable RK4 solver to a sample-specific final time $t + \Delta t^{\theta_i}$, with $\Delta t^{\theta_i} \sim \mathcal{U}\big(0,\Delta t_{\max}^{\theta_i}\big)$. 
Since the latent dynamics has $K - 1$ time derivatives, the RK4 solver outputs the latent state and its first $K - 1$ time derivatives—a prediction of the future ROM state.
Let $D_t^{(k)} \hat{z}(t + \Delta t^{\theta_i})$ denote the $k$th time derivative of the predicted latent state at the final time.
We decode the predicted ROM state to obtain the future FOM state and its first $K - 1$ time derivatives to obtain
\begin{equation*}
    D_t^{(k)} \hat{u}_{\theta}(t + \Delta t^{\theta}) = \varphi_d^{(k)} \big(\hat{z}_{\theta}(t + \Delta t^{\theta})\big).
\end{equation*}
We compare this to a piecewise-linear interpolation of the FOM solution at $t + \Delta t^{\theta_i}$, denoted $D_{t}^{(k)} \tilde{u}_{\theta_i}\left(t_{j}^{\theta_i} + \Delta t^{\theta_i}(j) \right)$ for the $k$th time derivative.
The Rollout loss for HLaSDI is 
\begin{equation}
\begin{aligned}
    \mathcal{L}_{\text{Rollout}}\bigg( \varphi_e^{(0)}, \varphi_d^{(0)}, \ldots, &\varphi_e^{(K - 1)}, \varphi_d^{(K - 1)}, \left\{ b_{\theta_i}, C_{\theta_i}^{(0)}, \ldots, C_{\theta_i}^{(K - 1)} \right\}_{i = 1}^{N_{\theta}} \bigg) \\
    &= \frac{1}{N_{ro}} \sum_{k = 0}^{K - 1} \sum_{i = 1}^{N_{\theta}} \frac{1}{\sigma_{\theta_i}^{(k)}}\sum_{j = 0}^{N_{ro}(\theta_i)} \left\|  D_{t}^{(k)} \tilde{u}_{\theta_i} \left(t_{j}^{\theta_i} + \Delta t_{\text{ro}}^{\theta_i}(j) \right) - D_{t}^{(k)} \hat{u}_{\theta_i}\left(t_j^{\theta_i} + \Delta t_{ro}^{\theta_i}(j) \right) \right\|_1.
\end{aligned}
\label{eq:loss:rollout:higher}
\end{equation}
where $N_{ro}$ is the number of rollable frames (see section \ref{Background:Rollout}) and 
\begin{equation*}
    \sigma^{(k)}_{\theta_i} = \text{STD}\left(  D_t^{(k)} \vec{u}_{\theta_i}\left(t_j^{\theta_i}\right) \cdot e_r\ \Big|\ r \in \left\{ 1, 2, \ldots, N_u \right\},\ j \in \left\{ 1, 2, \ldots, N_t(\theta_i) \right\} \right) 
\end{equation*}
is the standard deviation of the components of the $k$th derivative of the numerical FOM solution.

\textbf{Initial Condition Rollout Loss:} HLaSDI introduces a second Rollout loss (inspired by Lusch et al.'s \cite{lusch2018koopman} work on Koopman operators) to improve long-term predictions.
For each training parameter $\theta_i$, we encode the first $K - 1$ derivatives of the FOM's initial condition, 
\begin{equation*}
 \left\{ \vec{u}_{\theta_i}^{(k)}(0) \right\}_{k = 0}^{K - 1}.
\end{equation*}
obtaining
\begin{equation*}
    \left\{ \varphi_e^{(k)} \left( \vec{u}_{\theta_i}^{(k)}(0) \right) \right\}_{k = 0}^{K - 1}.
\end{equation*}   
We use this encoding as the initial condition for the latent dynamics for $\theta_i$.
A differentiable RK4 solver integrates the latent dynamics to $t_0^{\theta_i}, \ldots, t_{N_{\text{IC max}}(\theta_i)}^{\theta_i}$, where $N_{\text{IC max}}(\theta_i) \in \left\{ 0, \ldots, N_t(\theta_i) \right\}$.
We call $t_{N_{\text{IC max}}(\theta_i)}$ the \emph{Initial Condition Rollout Horizon} for $\theta_i$, denoted $t_{\text{IC max}}^{\theta_i}$, which starts at $0$ and gradually increases during training (see section \ref{Discussion} for details).
We retain the entire solution trajectory:
\begin{equation*}
    \left\{ D_t^{(k)} \vec{z}_{\theta_i}(0), \ldots, D_t^{(k)} \vec{z}_{\theta_i}\left(t_{\text{IC max}}^{\theta_i}\right) \right\}_{k = 0}^{K - 1}.
\end{equation*}
Decoding this predicted trajectory yields
\begin{equation*}
    \left\{ \varphi_d^{(k)} \left( D_t^{(k)} \vec{z}_{\theta_i}(0) \right), \ldots, \varphi_d^{(k)} \left( D_t^{(k)} \vec{z}_{\theta_i}\left(t_{\text{IC max}}^{\theta_i}\right) \right) \right\}_{k = 0}^{K - 1},
\end{equation*}
which we compare to $\left\{ D_t^{(k)} \vec{u}_{\theta_i}(0) , \ldots, D_t^{(k)} \vec{u}_{\theta_i}\left(t_{\text{IC max}}^{\theta_i}\right) \right\}_{k = 0}^{K - 1}$.
Applying this to each training parameter gives the \emph{Initial Condition Rollout Loss}:
\begin{equation}
\begin{aligned}
    \mathcal{L}_{\text{IC Rollout}} \bigg( \varphi_e^{(0)}, &\varphi_d^{(0)}, \ldots, \varphi_e^{(K - 1)}, \varphi_d^{(K - 1)}, \left\{ b_{\theta_i}, C_{\theta_i}^{(0)}, \ldots, C_{\theta_i}^{(K - 1)} \right\}_{i = 1}^{N_{\theta}} \bigg) \\
    & = \sum_{k = 0}^{K - 1} \sum_{i = 1}^{N_{\theta}} \frac{1}{N_{\text{IC Rollout}}  (\theta_i) \sigma_{\theta_i}^{(k)}} \sum_{j = 0}^{N_{\text{IC Rollout}}(\theta_i)} \left\| \varphi_d^{(k)} \left( D_t^{(k)} \vec{z}_{\theta_i}\left(t_j^{\theta_i}\right) \right) - D_t^{(k)} \vec{u}\left(t_j^{\theta_i} \right) \right\|_1
\end{aligned}
\label{eq:Loss:IC_Rollout}
\end{equation}

\subsection{The Consistency and Chain-Rule Losses}
\label{Method:Consistency_Chain_Rule}

So far, we have assumed that $\varphi_e^{(k)}$'s encodings act as a proxy for $k$'th time derivative of the latent state.
In this section, we define two loss functions --- the Consistency and Chain-Rule losses --- which encode this mathematical structure into our model.

\textbf{Consistency Loss:} HLaSDI uses $\varphi_e^{(k)} \left( D_t^{(k)} \vec{u}_{\theta} \left( t \right) \right)$ as a proxy for $D_t^{(k)} \vec{z}_{\theta} \left( t \right)$, the $k$th time derivative of the latent state at time $t$ for parameter $\theta$.
Thus, for $k \in \{ 0, 1, \ldots, K - 2 \}$, we expect
\begin{equation*}
    \varphi_e^{(k + 1)} \left( D_t^{(k + 1)} \vec{u}_{\theta} \left( t \right) \right) \approx D_t \varphi_e^{(k)} \left( D_t^{(k)} \vec{u}_{\theta} \left( t \right) \right).
\end{equation*}
To enforce this, for each training parameter $\theta_i$, we use $\varphi_e^{(k)}$ and $\varphi_e^{(k + 1)}$ to encode 
\begin{equation*}
    \left\{ D_t^{(k)} \vec{u}\left( t_j^{\theta_i} \right) \right\}_{j = 0}^{N_t(\theta_i)}
    \quad\text{and}\quad
    \left\{ D_t^{(k + 1)} \vec{u}\left( t_j^{\theta_i} \right) \right\}_{j = 0}^{N_t(\theta_i)},
\end{equation*} 
respectively.
Applying the finite-difference schemes from Section \ref{Method:Finite_Diff:First} to the former yields a finite-difference approximation to
\begin{equation*}
    \left\{ D_t \varphi_e^{(k)} \left( D_t^{(k)} \vec{u}\left( t_j^{\theta_i} \right) \right) \right\}_{j = 0}^{N_t(\theta_i)},
\end{equation*}
which should approximate the latter.
Applying this to each training parameter yields the \emph{Consistency loss}:
\begin{equation}
\begin{aligned}
    \mathcal{L}_{\text{Consistency}} \Big( &\varphi_e^{(0)}, \dots, \varphi_e^{(K - 1)} \Big) \\
    &= \frac{1}{N_{\theta}} \sum_{i = 1}^{N_{\theta}} \frac{1}{1 + N_t(\theta_i)} \sum_{j = 0}^{N_t(\theta_i)} \left\| D_t \varphi_e^{(k)}\left( D_t^{(k)} \vec{u}_{\theta_i}\left( t_j^{\theta} \right) \right) - \varphi_e^{(k + 1)} \left( D_t^{(k + 1)} \vec{u}_{\theta_i} \left( t_j^{\theta_i} \right) \right) \right\|_1.
\end{aligned}
\label{eq:Loss:Consistency}
\end{equation}
The Consistency loss ensures that $\varphi_e^{(k)}$'s encodings reasonably approximate the $k$th time derivative of the latent state.
In principle, we could apply finite-difference schemes for $q$th time derivatives to $\varphi_e^{(k)}$'s encodings and compare them to encodings from $\varphi_e^{(k + q)}$. 
However, we restrict our attention to first time derivatives.

\textbf{Chain-Rule Loss:} By the chain rule, 
\begin{equation*}
\begin{aligned}
    \varphi_e^{(k + 1)} \left( D_t^{(k)} \vec{u}_{\theta}(t) \right) &\approx D_t \varphi_e^{(k)} \left( D_t^{(k)} \vec{u}_{\theta}(t) \right) \\
    &= \left[ D \varphi_e^{(k)} \left(  D_t^{(k)} \vec{u}_{\theta}(t) \right) \right] \left( D_t^{(k + 1)} \vec{u}_{\theta}(t) \right),
\end{aligned}
\end{equation*}
where $D \varphi_e^{(k)} (u) \in \mathbb{R}^{L \times N_u}$ denotes the Fréchet derivative (Jacobian) of $\varphi_e^{(k)}$ at $u \in \mathbb{R}^{N_u}$. 
We use this relationship to derive a loss function.

For each training parameter $\theta_i$, we use $\varphi_e^{(k)}$ and $\varphi_e^{(k + 1)}$ to encode 
\begin{equation*}
    \left\{ D_t^{(k)} \vec{u}_{\theta} \left( t_j^{\theta_i} \right) \right\}_{j = 0}^{N_t(\theta_i)}
    \quad\text{and}\quad
    \left\{ D_t^{(k + 1)} \vec{u}_{\theta} \left( t_j^{\theta_i} \right) \right\}_{j = 0}^{N_t(\theta_i)},
\end{equation*}
respectively.
For each time step, we use automatic differentiation to compute the Jacobian-Vector product 
\begin{equation*}
    \left[ D \varphi_e^{(k)} \left(  D_t^{(k)} \vec{u}_{\theta} \left( t_j^{\theta_i} \right) \right) \right] \left( D_t^{(k + 1)} \vec{u}_{\theta}\left( t_j^{\theta_i} \right) \right),
\end{equation*}
which we compare to 
\begin{equation*}
    \varphi_e^{(k + 1)} D_t^{(k + 1)} \vec{u}_{\theta} \left( t_j^{\theta_i} \right).
\end{equation*}
Applying this to each training parameter yields the \emph{Chain-Rule Loss},
\begin{equation}
\begin{aligned}
    &\mathcal{L}_{\text{Chain-Rule}} \Big( \varphi_e^{(0)}, \ldots, \varphi_e^{(K - 1)} \Big) \\
    &\ \ \ = \frac{1}{N_{\theta}} \sum_{i = 1}^{N_{\theta}} \frac{1}{1 + N_t(\theta_i)} \sum_{j = 0}^{N_t(\theta_i)} \left\| \left[ D \varphi_e^{(k)} \left(  D_t^{(k)} \vec{u}_{\theta} \left( t_j^{\theta_i} \right) \right) \right] \left( D_t^{(k + 1)} \vec{u}_{\theta}\left( t_j^{\theta_i} \right) \right) -  \varphi_e^{(k + 1)} D_t^{(k + 1)} \vec{u}_{\theta} \left( t_j^{\theta_i} \right) \right\|_1,
\end{aligned}
\label{eq:Loss:Chain_Rule}
\end{equation}
which relates the $(k + 1)$th time derivative of the numerical FOM solution to the encoding of the $k$th time derivative.
This loss reinforces that $\varphi_e^{(k)}$ predicts the $k$th time derivative of the latent state.

\subsection{The final loss function}
\label{Method:Loss}

Combining the developments above, HLaSDI trains the $K$ encoder-decoder pairs and learns the latent dynamics by minimizing the following loss function:
\begin{equation}
\begin{aligned}
    \mathcal{L}\bigg( \varphi_e^{(0)}, \varphi_d^{(0)}, \ldots, &\varphi_e^{(K - 1)}, \varphi_d^{(K - 1)}, \left\{ b_{\theta_i}, C_{\theta_i}^{(0)}, \ldots, C_{\theta_i}^{(K - 1)} \right\}_{i = 1}^{N_{\theta}} \bigg) \\ 
    =\ & \eta_{\text{Recon}} \sum_{i = 1}^{N_{\theta}} \mathcal{L}_{\text{Recon}} \left( \varphi_e^{(i)}, \varphi_d^{(i)} \right) \\
    +\ &\eta_{\text{LD}}\ \mathcal{L}_{LD}\Big(\varphi_e^{(0)}, \ldots, \varphi_e^{(K - 1)}, \left\{ b_{\theta_i}, C_{\theta_i}^{(0)}, \ldots,  C_{\theta_i}^{(K - 1)} \right\}_{i = 1}^{N_{\theta}} \Big) \\
    +\ &\eta_{\text{Rollout}}\ \mathcal{L}_{\text{Rollout}}\bigg( \varphi_e^{(0)}, \varphi_d^{(0)}, \ldots, \varphi_e^{(K - 1)}, \varphi_d^{(K - 1)}, \left\{ b_{\theta_i}, C_{\theta_i}^{(0)}, \ldots, C_{\theta_i}^{(K - 1)} \right\}_{i = 1}^{N_{\theta}} \bigg) \\
    +\ &\eta_{\text{IC Rollout}}\ \mathcal{L}_{\text{IC Rollout}} \bigg( \varphi_e^{(0)}, \varphi_d^{(0)}, \ldots, \varphi_e^{(K - 1)}, \varphi_d^{(K - 1)}, \left\{ b_{\theta_i}, C_{\theta_i}^{(0)}, \ldots, C_{\theta_i}^{(K - 1)} \right\}_{i = 1}^{N_{\theta}} \bigg) \\
    +\ &\eta_{\text{Consistency}}\ \mathcal{L}_{\text{Consistency}} \Big( \varphi_e^{(0)}, \dots, \varphi_e^{(K - 1)} \Big) \\
    +\ &\eta_{\text{Chain-Rule}}\ \mathcal{L}_{\text{Chain-Rule}} \Big( \varphi_e^{(0)}, \ldots, \varphi_e^{(K - 1)} \Big) \\
    +\ &\eta_{\text{Coefficient}} \sum_{i = 1}^{N_{\theta}} \left\| b_{\theta_i} \right|_2^2 + \left\| C_{\theta_i}^{(0)} \right\|_{F}^2 + \cdots + \left\| C_{\theta_i}^{(K- 1)} \right\|_{F}^2 
\end{aligned}
\label{eq:Loss}
\end{equation}
where $\eta_{\text{Recon}},\ \eta_{\text{LD}},\ \eta_{\text{Rollout}},\ \eta_{\text{IC Rollout}},\ \eta_{\text{Consistency}},\ \eta_{\text{Chain-Rule}}$, and $\eta_{\text{Coefficient}}$ are positive hyperparameters weighting the various loss functions.
The final term is the sum of squared coefficients, included as a regularizer.
HLaSDI minimizes \eqref{eq:Loss} using the Adam optimizer \cite{kingma2014adam}.
It follows GPLaSDI's episodic approach: training for a fixed number of epochs, then training Gaussian processes to predict latent coefficients for arbitrary parameter values, and then greedily selecting a new training parameter before resuming training. 
See Section \ref{Background:GPLaSDI} for details.

Table \ref{Table:Method:Notation} lists the notation we introduced in this section.

\begin{table}[hbt]
    \centering 
    \rowcolors{2}{cyan!10}{white}
    
    \begin{tabulary}{1.0\linewidth}{p{1.6cm}L}
        \toprule[0.3ex]
        \textbf{Notation} & \textbf{Meaning} \\
        \midrule[0.1ex]
        $\mathcal{L}_{LD}$ & The latent dynamics loss. See equation \eqref{eq:loss:LD}. \\
        \addlinespace[0.4em]
        $\mathcal{L}_{\text{Rollout}}$ & The Rollout Loss for HLaSDI. See equation \eqref{eq:loss:rollout:higher}. \\
        \addlinespace[0.4em]
        $\sigma_{\theta_i}^{(k)}$ & The standard deviation of the components of the $k$'th derivative of the numerical FOM solution. That is, $\theta^{(k)}_{\theta_i} = \text{STD}\left(  D_t^{(k)} \vec{u}_{\theta_i}\left(t_j^{\theta_i}\right) \cdot e_r\ \Big|\ r \in \left\{ 1, 2, \ldots, N_u \right\},\ j \in \left\{ 1, 2, \ldots, N_t(\theta_i) \right\} \right)$. \\
        \addlinespace[0.4em]
        $t_{\text{IC max}}^{\theta}$ & The Initial Condition Rollout Horizon for $\theta$. See section \ref{Method:Rollout}. This is the final time to which we integrate the latent dynamics (starting at the encoded initial condition when the parameter is $\theta$). We select this to be an element of $\{ t_j^{\theta} \}_{j = 0}^{N_t(\theta)}$. This value gradually increases from $0$ during training. \\
        \addlinespace[0.4em]
        $N_{\text{IC max}}(\theta_i)$ & By definition, $t_{\text{IC max}}^{\theta}$ is in $\{ t_j^{\theta} \}_{j = 0}^{N_t(\theta)}$. Specifically, $t_{\text{IC max}}^{\theta} = t_{N_{\text{IC max}}(\theta_i)}^{\theta_i}$. \\
        \addlinespace[0.4em]
        $\mathcal{L}_{\text{IC Rollout}}$ & The Initial Condition Rollout Loss. See equation \eqref{eq:Loss:IC_Rollout}. \\
        \addlinespace[0.4em]
        $\mathcal{L}_{\text{Consistency}}$ & The Consistency loss used to relate the first time derivative of $\varphi_e^{(k)}$'s encodings to $\varphi_e^{(k + 1)}$'s encodings. See equation \eqref{eq:Loss:Consistency}. \\
        \addlinespace[0.4em]
        $D f(u)$ & The Frechet derivative (Jacobian) of a function $f : \mathbb{R}^n \to \mathbb{R}^m$ applied to the vector $u \in \mathbb{R}^n$. $Df$ is an $m \times n$ matrix whose $i, j$ element is the $j$'th partial derivative of the $i$'th component function of $f$ at $u$. We use this in defining the Chain-Rule loss, equation \eqref{eq:Loss:Chain_Rule}. \\
        \addlinespace[0.4em]
        $\mathcal{L}_{\text{Chain-Rule}}$ & The Chain-Rule loss, which relates the $k + 1$'th time derivative of the numerical FOM solution to the encoding of the $k$'th time derivative of the numerical FOM solution. See equation \eqref{eq:Loss:Chain_Rule}. \\
        \addlinespace[0.4em]
        $\eta_{\text{Recon}}$ & A positive number representing the weight of the Reconstruction loss in the loss function, equation \eqref{eq:Loss}. There are analogous coefficients for the other pieces of the loss function. See equation \eqref{eq:Loss}. \\
        \bottomrule[0.3ex]
    \end{tabulary}
    
    \caption{The notation and terminology of section (\ref{Problem}).}
    \label{Table:Method:Notation}.
\end{table}
\section{Experiments}
\label{Experiments}



In this section, we evaluate HLaSDI on four parameterized PDEs with varying time derivative orders. 
Through these experiments, we demonstrate that HLaSDI achieves accurate predictions across diverse problem types using adaptive greedy sampling with minimal training data.

\begin{table}[t]
  \caption{Model architectures and hyperparameters for each experiment.}
  \label{tab:hyperparams}
  \centering
  \footnotesize
  \begin{tabular}{l c c c c c}
    \toprule
    \textbf{Experiment} & \textbf{Loss weights}$^{\dagger}$ &
    \textbf{Learning rate} & \textbf{Architecture} &
    \textbf{Iterations} & \textbf{Sampling freq.}$^{\ddagger}$ \\
    \midrule
    1D Burger's  & $(1, 1, 0, 0.2, 1, 1, 10^{-4})$ & $10^{-3}$ & 1001-250-100-100-5 & 20,000 & 2,500 \\
    Wave & $(1, 1, 1, 0, 1, 0, 10^{-3})$ & $10^{-3}$ & 1000-250-100-100-10 & 40,000 & 5,000 \\
    Telegrapher's & $(1, 1, 1, 1, 1, 1, 10^{-3})$ & $10^{-3}$ & 1000-250-100-100-10 & 20,000 & 4,000 \\
    Klein–Gordon & $(1, 1, 1, 1, 1, 1, 10^{-3})$ & $10^{-3}$ & 1000-250-100-100-10 & 36,000 & 6,000 \\
    \bottomrule
  \end{tabular}

  \vspace{3pt}
  \raggedright
  $^{\dagger}$Order of loss weights:
  $(\eta_{\mathrm{Recon}},\, \eta_{\mathrm{LD}},\, \eta_{\mathrm{Rollout}},\,
  \eta_{\mathrm{IC\ Rollout}},\, \eta_{\mathrm{Consistency}},\,
  \eta_{\mathrm{Chain\text{-}Rule}},\, \eta_{\mathrm{Coefficient}})$. \\
  $^{\ddagger}$ Number of training iterations between greedy sampling of parameter space.
\end{table}

\begin{table}[t]
  \caption{Sumary of HLaSDI results for each experiment.}
  \label{tab:results}
  \centering
  \footnotesize
  \begin{tabular}{l c c c c}
    \toprule
    \textbf{Experiment} & \textbf{Max relative error of $u$} &
    \textbf{Max relative error of $\tfrac{\partial u}{\partial t}$} & \textbf{Prediction Time} &
    \textbf{Speedup} \\
    \midrule
    1D Burger's  & $4.99\%$ & $8.08\%$ & 0.026s & 10$\times$  \\
    Wave & $1.60\%$ & $4.23\%$ & 0.031s & 40$\times$  \\
    Telegrapher's & $3.03.\%$ & $3.08\%$ & 0.031s & 37$\times$  \\
    Klein–Gordon & $7.93\%$ & $4.70\%$ & 0.031s & 38$\times$  \\
    \bottomrule
  \end{tabular}
\end{table}

\subsection{Experimental Setup}

\textbf{Network Architecture:} We use the notation 1000-100-10-5 to represent a fully connected network with input dimension 1000, two hidden layers (sizes 100 and 10), and output dimension 5. 
The decoder has the reverse architecture of the encoder (e.g., 5-10-100-1000). 
We employ sine activation functions after each layer except the encoder/decoder output layers, which have no activation.
We find sine activations effective for handling high-frequency derivative terms. 
We approximate $D_t^{(2)} z_{\theta}$ using the first time derivative of the latent velocity. 
Complete architectures, hyperparameters, and results are provided in Tables \ref{tab:hyperparams} and \ref{tab:results}.

\textbf{Training Strategy:} For all experiments, we generate FOM solutions on a uniform $11\times11$ grid across the two-dimensional parameter space, yielding 121 total simulations. 
Each parameter dimension is discretized with 11 equidistant points over its specified range. 
HLaSDI begins training on the four corners of the parameter space, and iteratively samples additional training points using the variance-based greedy sampling algorithm from GPLaSDI \cite{bonneville2024gplasdi}.
This adaptive sampling prioritizes regions where variance of the GPs is highest, improving sample efficiency by focusing on the most uncertain parameter regions of our model.
For our GP interpolation we use the \texttt{GaussianProcessRegressor} from scikit-learn \cite{sklearn} with the Mat\'{e}rn kernel
\begin{equation}
    k(\vc x, \vc y) = \frac{1}{\Gamma(\nu)2^{\nu-1}} \left(\frac{\sqrt{2\nu}}{l} \|\vc x - \vc y \|^2\right)^{\nu} K_{\nu}\left(\frac{\sqrt{2\nu}}{l} \| \vc x - \vc y \|^2\right),
\end{equation}
where $\Gamma (\cdot)$ is the Gamma function, $K_\nu$ is a modified Bessel function, and we set $\nu = 1.5$.
Once training and sampling are complete, the remaining FOM solutions are used for testing to evaluate performance of our algorithm. 
Training was done on a 2021 Macbook Pro with Apple M1 Chip with Pytorch's implementation of the Adam optimizer\cite{kingma2014adam}. 

\textbf{Error Metric:} We measure accuracy using the relative error metric
\begin{equation}
\varepsilon( \theta) = \max_{k = 0,1,..., N_t} \left( \frac{\mathrm{mean}i | \vec{u}_{\theta}(t_k)i  - \hat{u}_{\theta}(t_k)i |}{ \sigma{i, j }{ | \vec{u}_{\theta}(t_i)_j | } }\right),
\label{eq:errormetric}
\end{equation}
where $\sigma_\theta$ is the standard deviation of all components across all time steps in the FOM solution for parameter $\theta$.
We normalize by the standard deviation of FOM frames in the denominator of~\eqref{eq:errormetric} rather than the $\ell^2$-norm to avoid large errors in cases where the FOM solution has small magnitude.
This choice is particularly important for our experiments, as several initial conditions have zero initial velocity, which would prevents us using $\ell^2$ relative error for velocity.
The maximum relative error reported in our figures is computed by evaluating~\eqref{eq:errormetric} for each testing parameter in the $11 \times 11$ grid.

\subsection{1D Burger's Equation}
We first apply Higher-Order LaSDI to the 1D Burger's equation, previously explored with GPLaSDI \cite{bonneville2024gplasdi}:
\begin{align}
    \frac{\partial u}{\partial t}  = - u \frac{\partial u}{\partial x}, \quad u(0, x) = \cos (\pi w x ) \exp (-a x^2 ),
    \label{eq:1dburg}
\end{align}
where our input parameters $ \theta = (a, w)^\top$. control the width and periodicity of the initial condition. 

Although~\eqref{eq:1dburg} is a first order PDE, we approximate the velocity of the 1D Burger's equation using a fourth order finite-difference scheme.
This allows us to treat Burger's equation as a second order system and feed velocities $v(t,x)$ into HLaSDI.
By solving the Burger's equation forward in time, we approximate the initial velocity
\begin{equation}
    v(0,x) = \frac{\partial }{\partial t}u(t,x) \bigg|_{t = 0}.
\end{equation}

For the FOM, we apply explicit forward-Euler time integration with 501 uniformly spaced time steps on the interval $[0,1]$, and we use centered differences in space with $1001$ uniformly spaced spatial steps in the interval $[-3,3]$.
We run full-order simulations for parameter values $a \in [0.45, 0.55]$ and $w \in [0.18, 0.22]$.

\begin{figure}
  \centering
    \includegraphics[width= 0.49\linewidth]{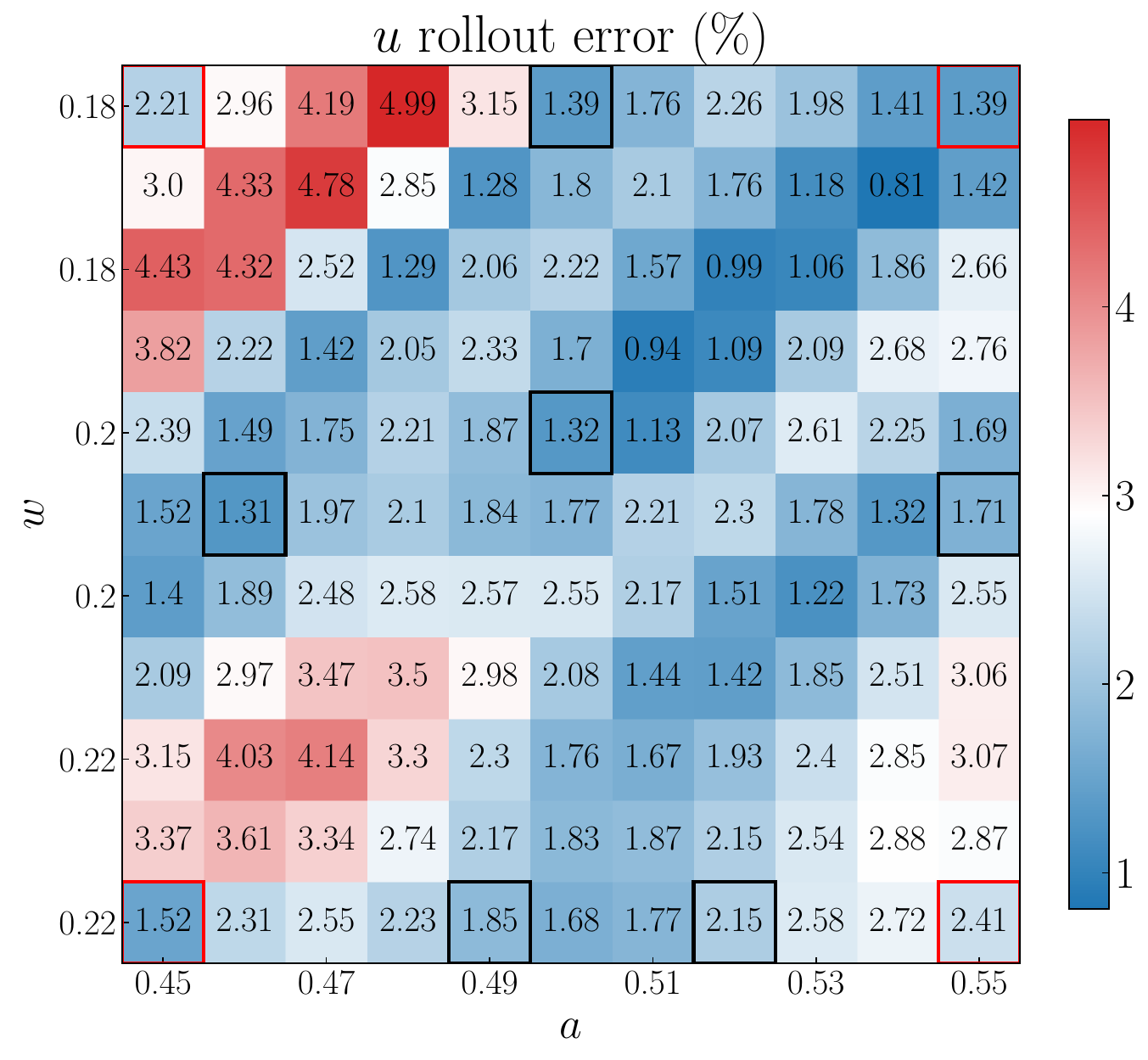}
    \includegraphics[width= 0.49\linewidth]{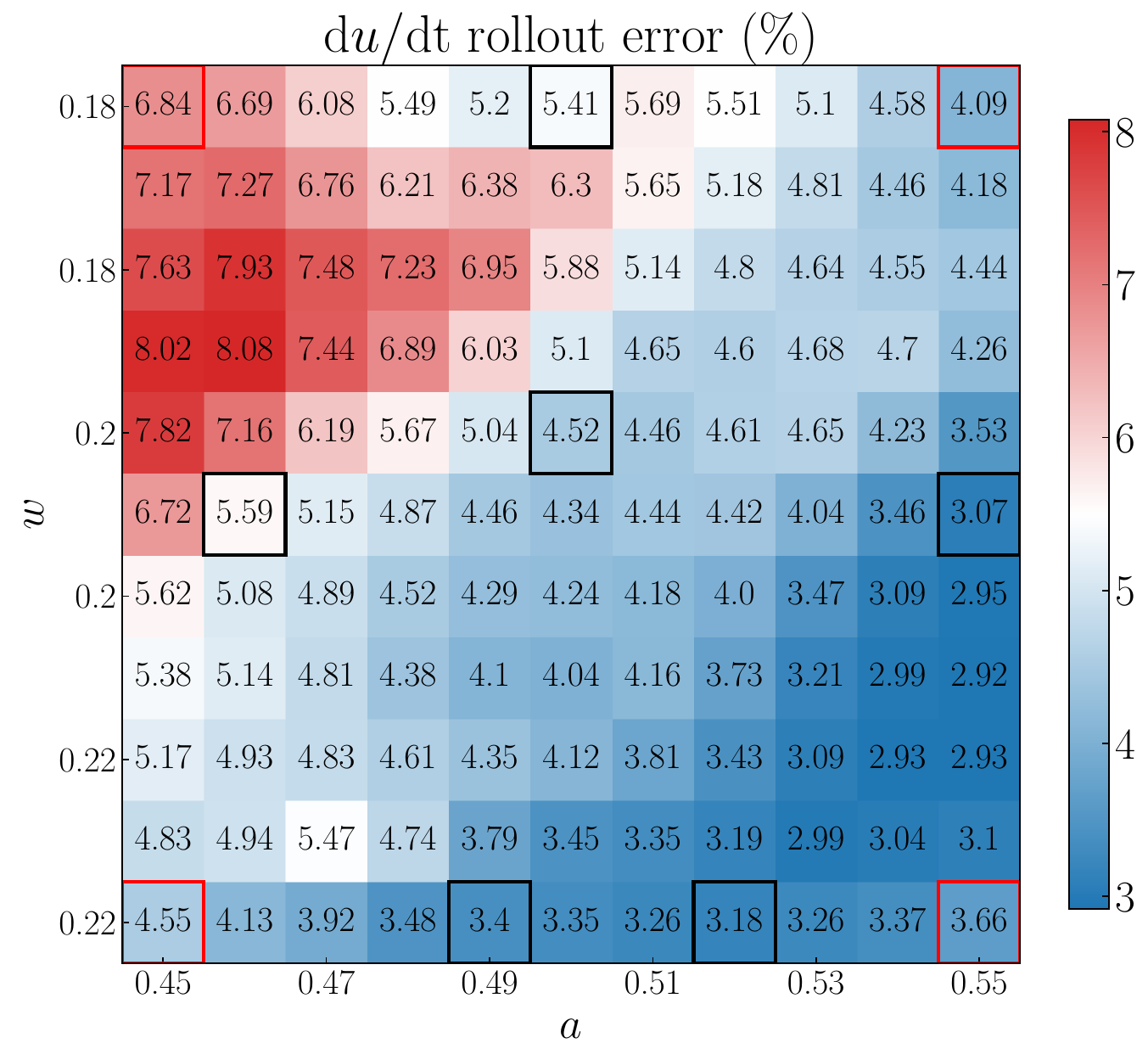}
  \caption{Relative error of (left) displacement and (right) velocity applying HLaSDI to the 1D Burger's equation~\eqref{eq:1dburg}. Red squares represent initial training points and black squares represent training points chosen through greedy sampling during training. All other grid points are testing parameters. HLaSDI achieves errors under 8.1\% using only 10 data points}
  \label{fig:1dbur_errs}
\end{figure}

Figure \ref{fig:1dbur_errs} shows that HLaSDI achieves a maximum relative error across the parameter space of 4.99\% for displacement and 8.08\% for velocity.
In Figure \ref{fig:wave_comp} we plot snapshots of the FOM and Higher-Order LaSDI solutions corresponding to the input parameter with worst relative error $(a,w) = (0.46, 0.19)$.
These results demonstrate that our model is able to achieve accurate predictions for both displacement and velocity, even for first order dynamical systems where the velocity is not explicitly provided.
Additionally, Higher-order LaSDI takes 0.026s to give a prediction, compared to 0.25s from the FOM.
This is a 10$\times$ speedup, demonstrating that Higher-Order LaSDI can provide accurate and rapid predictions.




\begin{figure}
  \centering
   \includegraphics[width= .9\linewidth]{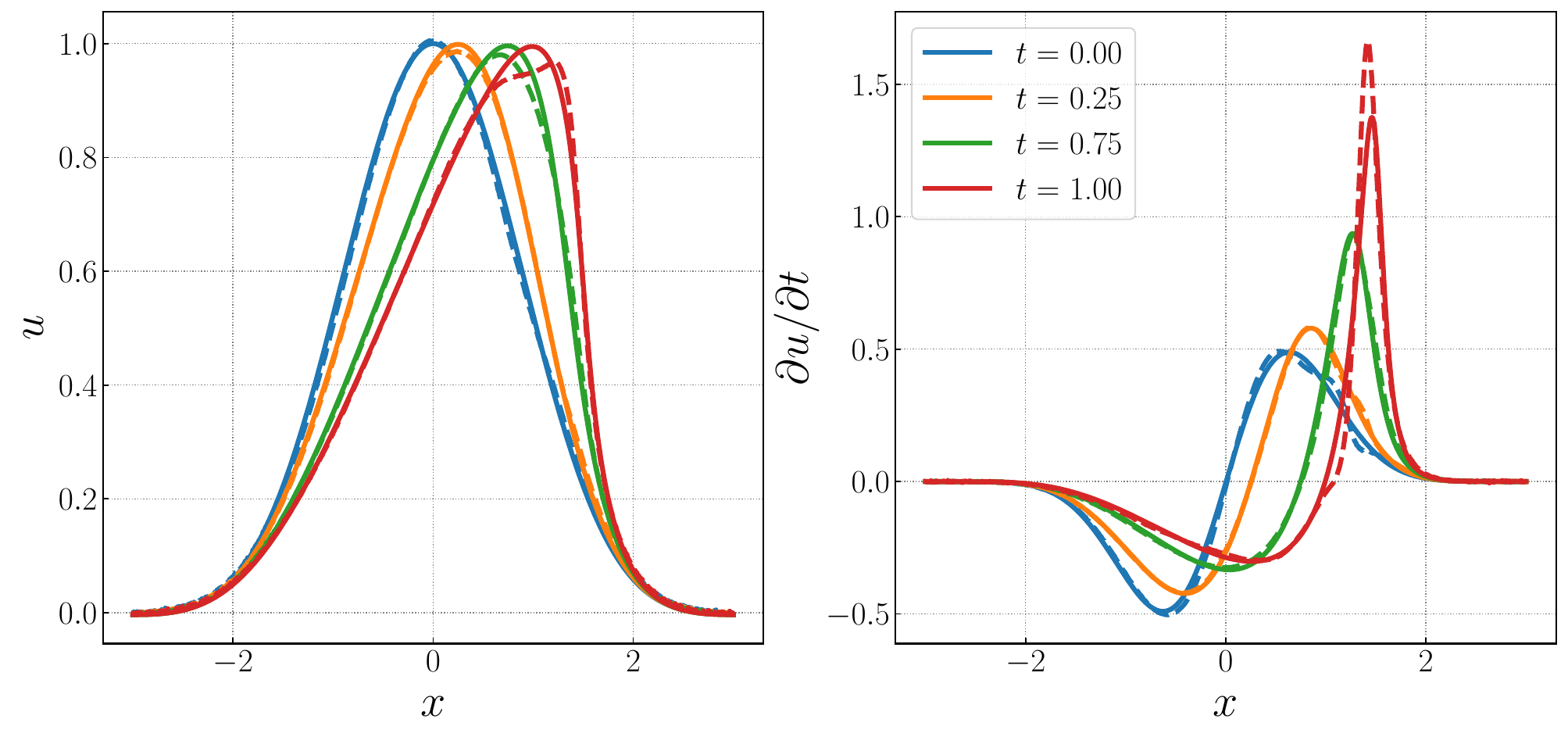}
  \caption{Solutions of FOM (solid lines) and Higher-order LaSDI (dashed lines) for the Burger's equation~\eqref{eq:1dburg} with parameter value $(a, w) = (0.46,0.19)$. We plot the (left) values of $u$ and (right) velocities.}
  \label{fig:burg_comp}
\end{figure}


\subsection{Wave Equation}
Next, we consider the 1D wave equation, a canonical second order system
\begin{equation}
    \frac{\partial^2 u}{\partial t^2} = c^2 \Delta u, \quad u(0, x, y) = \exp\big(-k (x^2 + y^2) \big), \quad
    \frac{\partial}{\partial t} u(0, x, y) = 0.
    \label{eq:wave}
\end{equation}
Here, our input parameters $ \theta = (c, k)^\top$ control the wave speed and initial condition of our simulations.
We note that, unlike the previous example, our input parameter $c$ affects the physics of the simulation rather than only parameterizing the initial condition.
This tests Higher-Order LaSDI's ability to learn and interpolate between input parameter values which alter the governing dynamics, not just initial conditions.

To generate our FOM solutions we use MFEM's \cite{anderson2021mfem} finite element implementation for wave equation on a star-shaped domain with natural boundary conditions.
We generate FOM simulations for parameter values $c \in [0.5, 0.6]$ and $w \in [2.0, 2.2]$.
After running the FOM simulations, we sample the solution at 1000 random points in the domain and use these points for training, testing, and evaluation of our model.

\begin{figure}
  \centering
    \includegraphics[width= 0.49\linewidth]{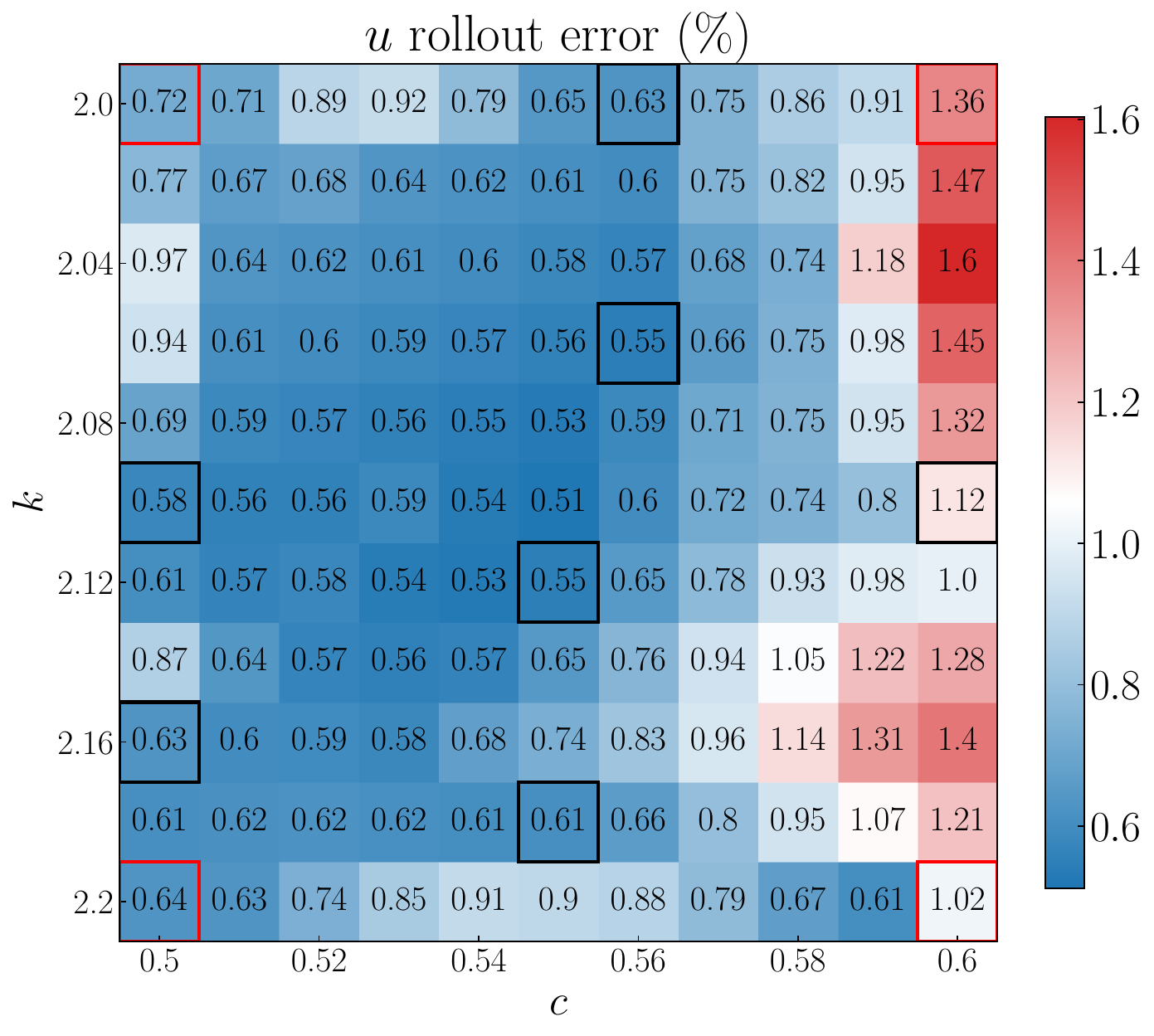}
    \includegraphics[width= 0.49\linewidth]{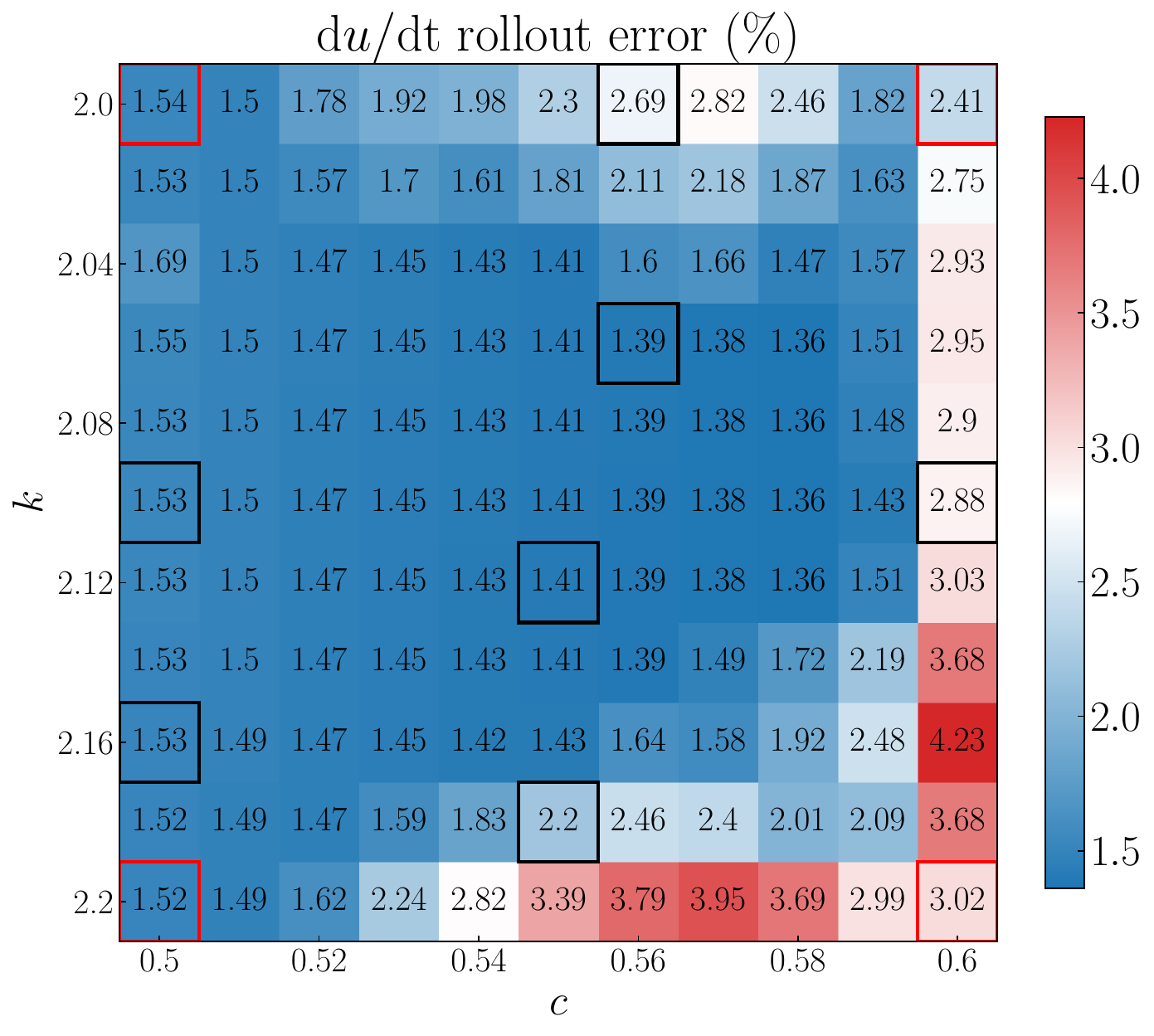}
  \caption{Relative error of (left) displacement and (right) velocity applying HLaSDI to the Wave equation~\eqref{eq:wave}. Red squares represent initial training points and black squares represent training points chosen through greedy sampling during training. All other grid points are testing parameters. HLaSDI achieves errors under 4.5\% using 11 data points}
  \label{fig:wave_errs}
\end{figure}

Figure \ref{fig:wave_errs} shows that HLaSDI achieves maximum errors of 1.60\% for displacement and 4.23\% for velocity across the entire parameter space only sampling 11 data points.
This is a substantial improvement over the Burger's equation, likely owing to the fact that we are dealing with a true second-order system here.
In addition to high prediction accuracy, Higher-order LaSDI provides a prediction in 0.031s compared to 1.24s for the FOM, giving 40$\times$ speedup.

\begin{figure}
  \centering
   \includegraphics[width= \linewidth]{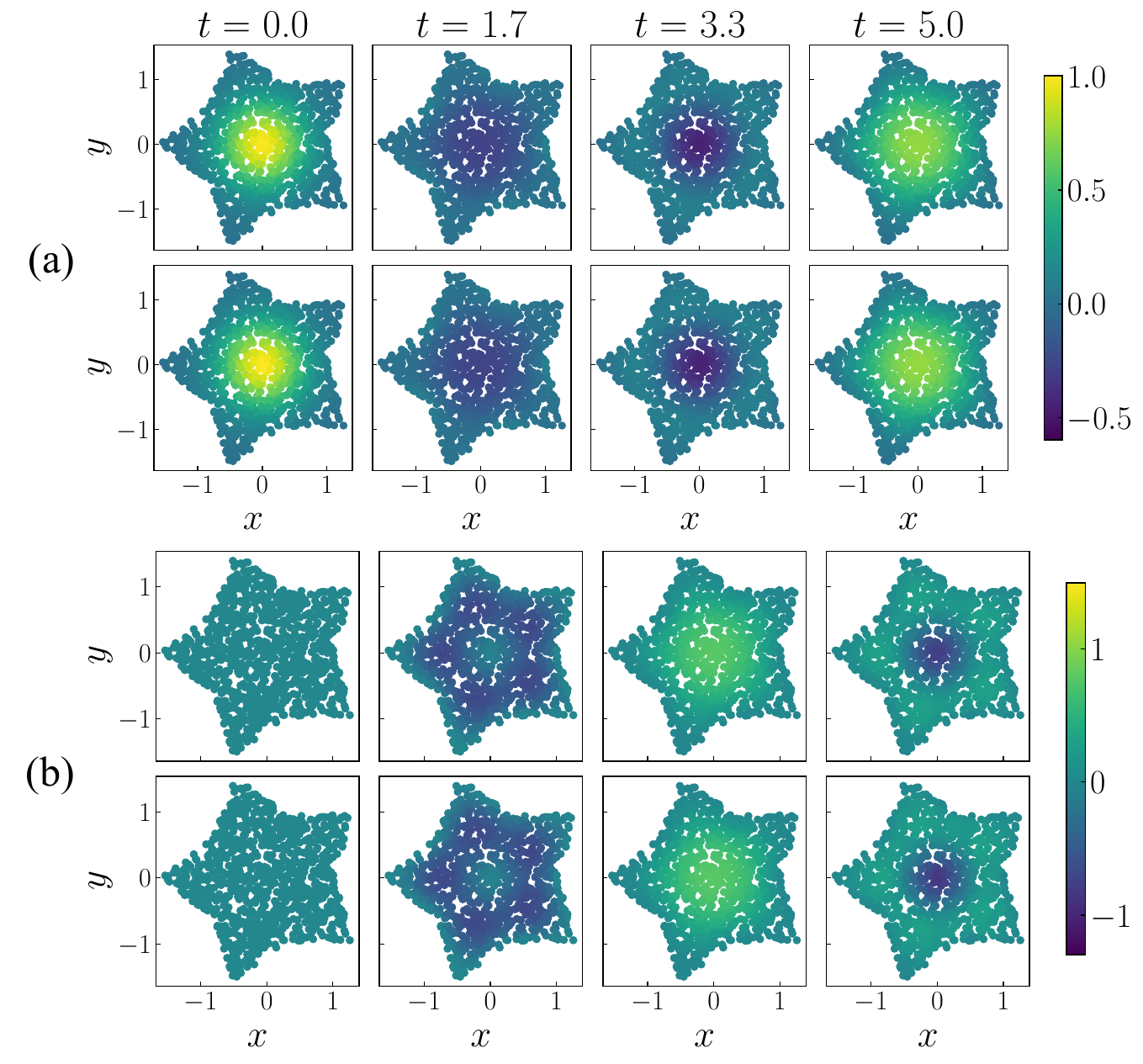}
  \caption{Solutions of FOM and Higher-order LaSDI for the wave equation~\eqref{eq:wave} with parameter value $(c, k) = (0.6,2.16)$. In (a) we plot the FOM values of $u$ on the top row and HLaSDI approximations of $u$ on the bottom row. Similarly, in (b) we plot FOM and HLaSDI velocities.}
  \label{fig:wave_comp}
\end{figure}

In Figure \ref{fig:wave_comp} we plot snapshots of the FOM and Higher-Order LaSDI solutions corresponding to the input parameter with worst relative error $(c,k) = (0.6, 2.16)$.
We see excellent agreement between the FOM and ROM predictions for both velocity and displacement, showing that we are able to capture propagation of the wave on a star-shaped domain which has non-trivial geometry.

\begin{figure}
  \centering
    \includegraphics[width= 0.49\linewidth]{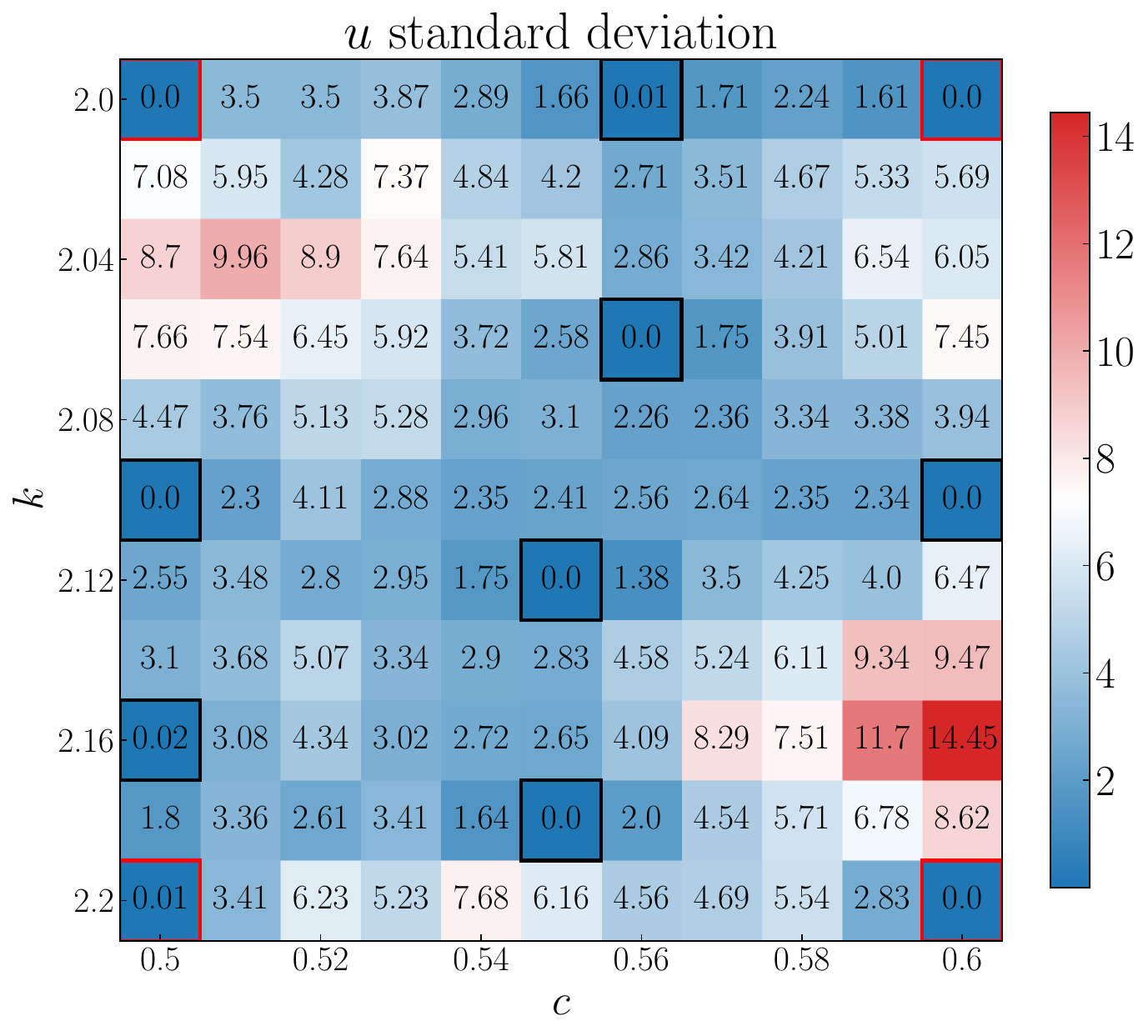}
    \includegraphics[width= 0.49\linewidth]{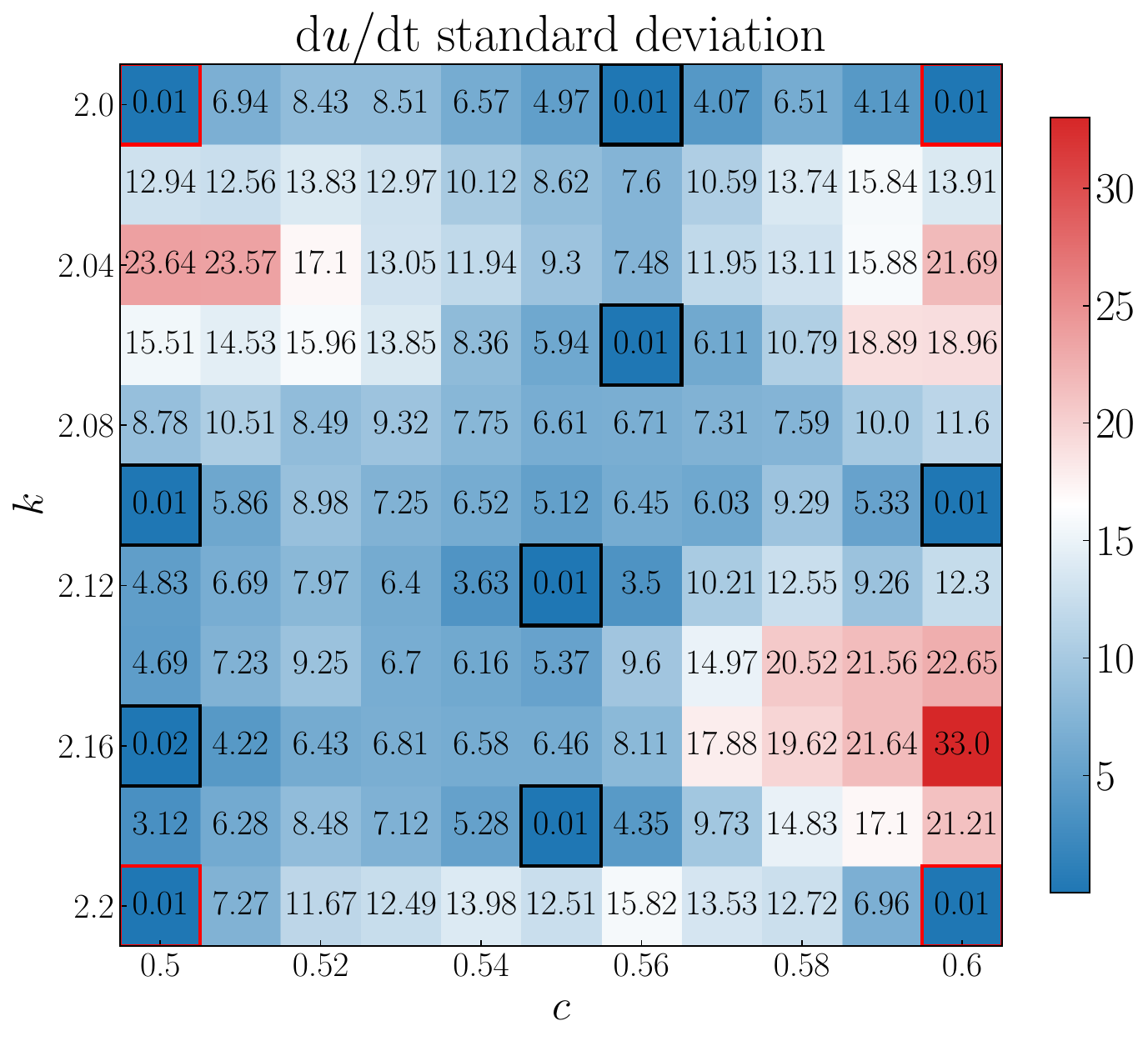}
  \caption{Standard deviation for our predictions of (left) $u$ and (right) $\tfrac{\partial u}{
  \partial t}$ applying HLaSDI to the Wave equation~\eqref{eq:wave}. Red squares represent initial training points and black squares represent training points chosen through greedy sampling during training. All other grid points are testing parameters.}
  \label{fig:wave_std}
\end{figure}

In Figure \ref{fig:wave_std} we plot standard deviation of our model predictions over the parameter space.
This allows us to obtain an a priori uncertainty estimate and enables us to continue our greedy sampling if we train the algorithm for longer.
As expected, training points will have the lowest standard deviation while regions further from the training points have larger standard deviation.
In this case, our GP sampling is validated: the highest uncertainty is located at the input parameter corresponding to the highest prediction error $(c,k) = (0.6, 2.16)$.
If we were to continue training, this point would be selected next by the greedy sampling algorithm based on its large uncertainty.

\subsection{Telegrapher's Equation}

Next, we consider the Telegrapher's equation which introduces a damping term to the wave equation
\begin{equation}
    \frac{\partial^2 u}{\partial t^2} = c^2 \Delta u  - 2 \alpha \frac{\partial u}{\partial t}, \quad u(0, x, y) = \exp\big(-k (x^2 + y^2) \big), \quad
    \frac{\partial}{\partial t} u(0, x, y) = 0.
    \label{eq:tele}
\end{equation}
Here, our input parameters $ \theta = (\alpha, k)^\top$ control strength of our damping term and width of the initial condition.
The Telegrapher's equation models propagation of voltage and current in transmission lines \cite{hayt2011engineering}.

\begin{figure}
  \centering
   \includegraphics[width= \linewidth]{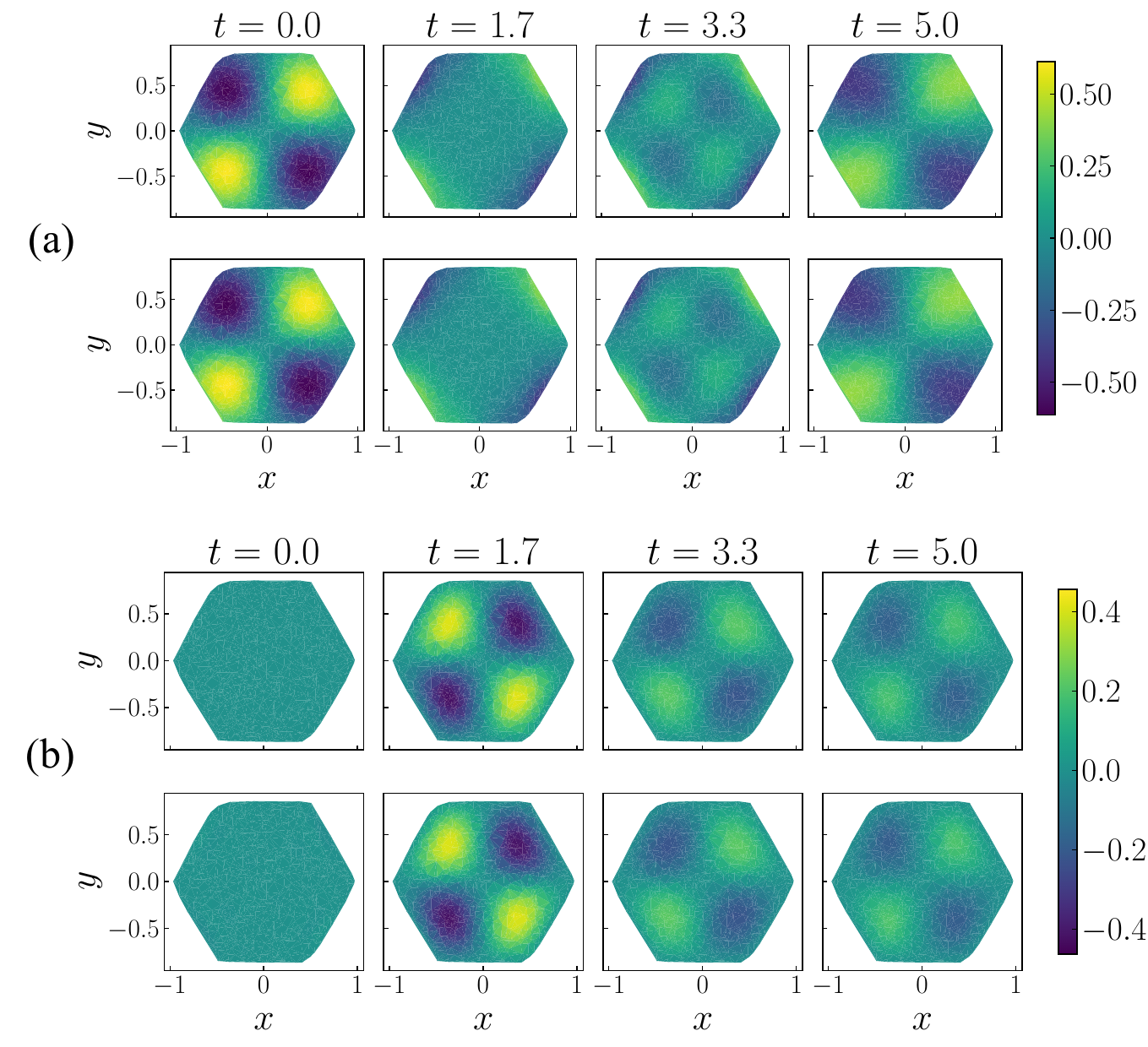}
  \caption{Solutions of FOM and Higher-order LaSDI for the Telegrapher's equation~\eqref{eq:tele} with parameter value $(\alpha, k) = (0.11,0.9)$. In (a) we plot the FOM values of $u$ on the top row and HLaSDI approximations of $u$ on the bottom row. Similarly, in (b) we plot FOM and HLaSDI velocities.}
  \label{fig:tele_comp}
\end{figure}

To generate FOM solutions we implement the Telegrapher's equation in MFEM on a hexagonal mesh with fixed wave speed $c = 0.2$.
We run full-order simulations for parameter values $\alpha \in [0.09, 1.1]$ and $w \in [0.09, 1.1]$.
Again, we sample the FOM solutions at 1000 random points in the domain for training, testing, and evaluation of our model.
We plot one example of the FOM and Higher-order LaSDI approximations in Figure \ref{fig:tele_comp}.

\begin{figure}
  \centering
    \includegraphics[width= 0.49\linewidth]{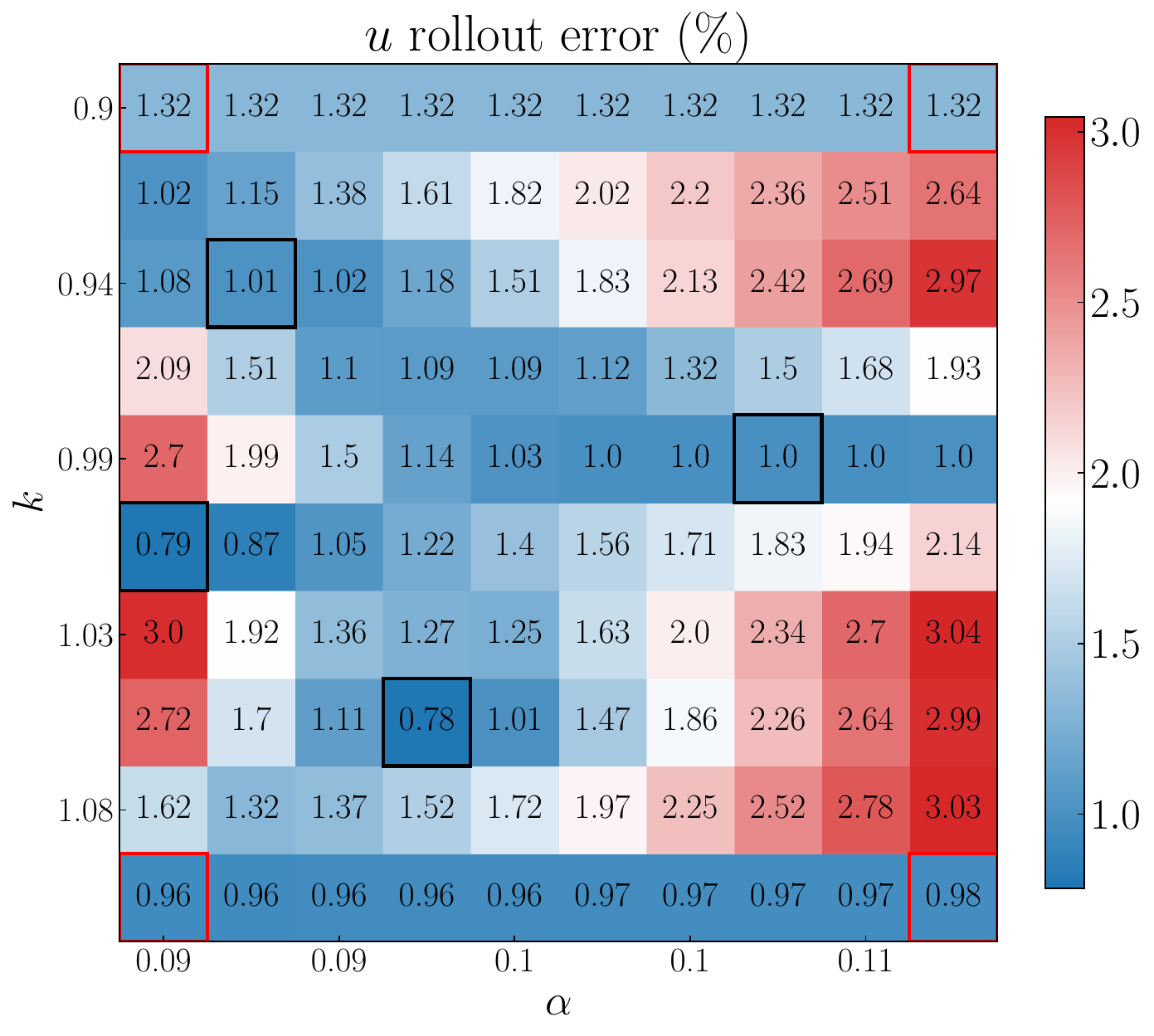}
    \includegraphics[width= 0.49\linewidth]{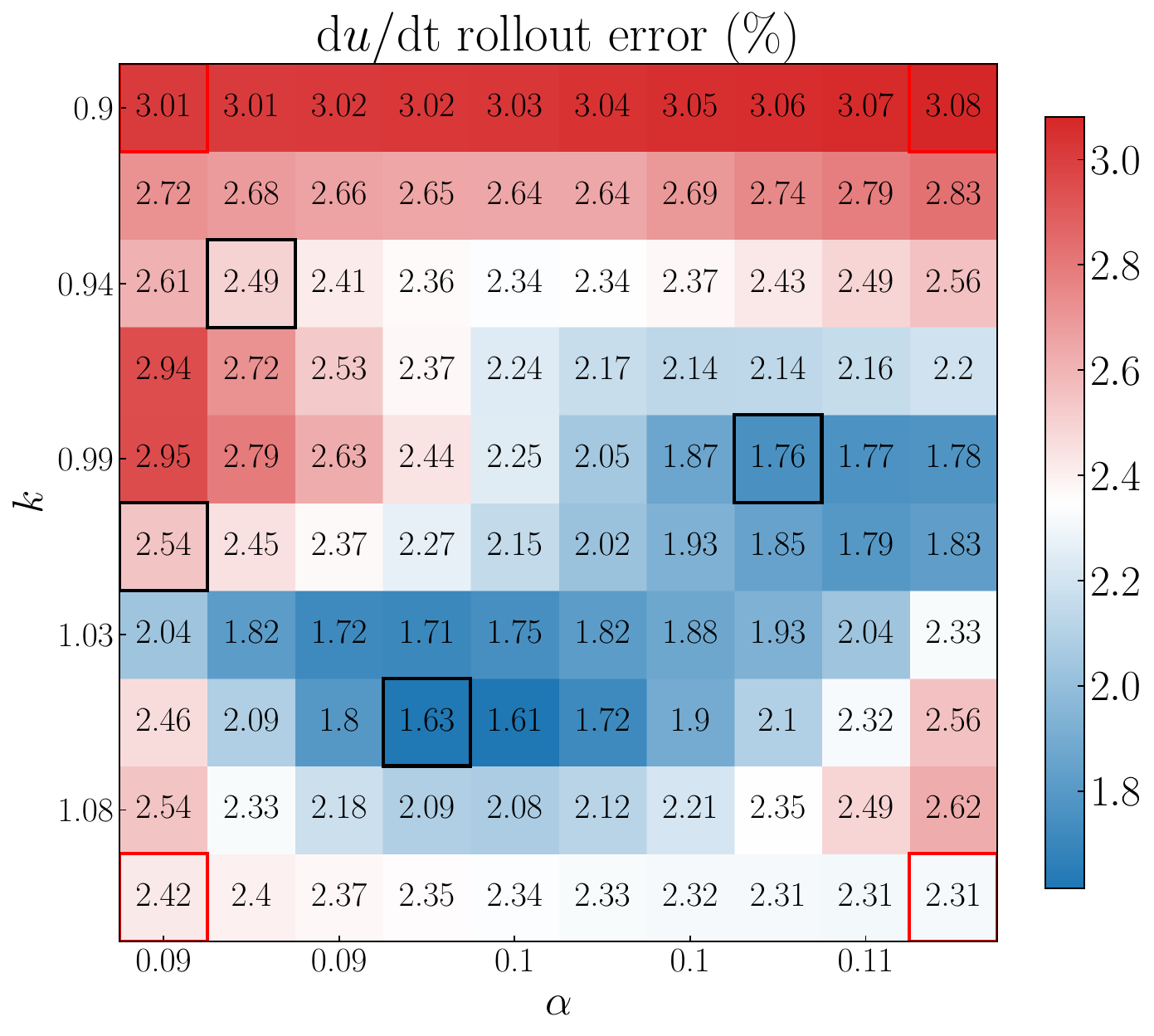}
  \caption{Relative error of (left) displacement and (right) velocity HLaSDI to the Telegrapher's equation~\eqref{eq:tele}. Red squares represent initial training points and black squares represent training points chosen through greedy sampling during training. All other grid points are testing parameters. HLaSDI achieves errors under 3.1\% using only 8 data points}
  \label{fig:tele_errs}
\end{figure}

In Figure \ref{fig:tele_errs} we plot the results after training with HLaSDI, achieving maximum displacement errors of 1.31\% and maximum velocity errors of 2.96\%. 
We note that the Telegrapher's equation required less training and greedy sampling than the wave equation, indicating that the damping term allowed us to achieve higher accuracy.
Higher-order LaSDI provides a prediction in 0.031s, which is the same as our model used for the wave equation.
This is not surprising given that our models have the same size for both problems (see Table \ref{tab:hyperparams}).
For the Telgrapher's equation, the FOM takes 1.16s, so our ROM provides 37$\times$ speedup.

\subsection{Klein-Gordon}

Finally, we solve the Klein-Gordon equation, a relativistic wave equation with a mass term
\begin{align}
    \frac{\partial^2 u}{\partial t^2} = c^2 \Delta u  - m^2 u, \quad u(0, x, y) = \exp\big(-k (x^2 + y^2) \big) \sin (\pi w x) \sin (\pi w y ), \quad \frac{\partial}{\partial t} u(0, x, y) = 0.
     \label{eq:klein}
\end{align}
Here, our input parameters $ \theta = (m, w)^\top$ control the mass term and spatial frequency of the initial condition.
The Klein-Gordon is a relativistic wave equation which often appears in quantum mechanics \cite{greiner2012relativistic}. 
The $-m^2u$ term couples the displacement directly to the second time derivative without involving spatial derivatives, providing a different coupling behavior than the Telegrapher's equation.

We implement the Klein-Gordon equation in MFEM on a hexagonal mesh with fixed wave speed $c = 0.2$.
As before, we sample the FOM solution at 1000 random points in the domain when applying HLaSDI .

In Figure \ref{fig:tele_errs} we plot the results after training with HLaSDI on the Klein-Gordon equation, achieving maximum displacement errors of 7.93\% and maximum velocity errors of 4.7\%. 
The initial condition for this experiment is the most complex of all our experiments, featuring exponential decay and oscillating sinusoidal terms. 
The relatively higher displacement error (7.93\%) likely reflects this increased complexity.

\begin{figure}
  \centering
    \includegraphics[width= 0.49\linewidth]{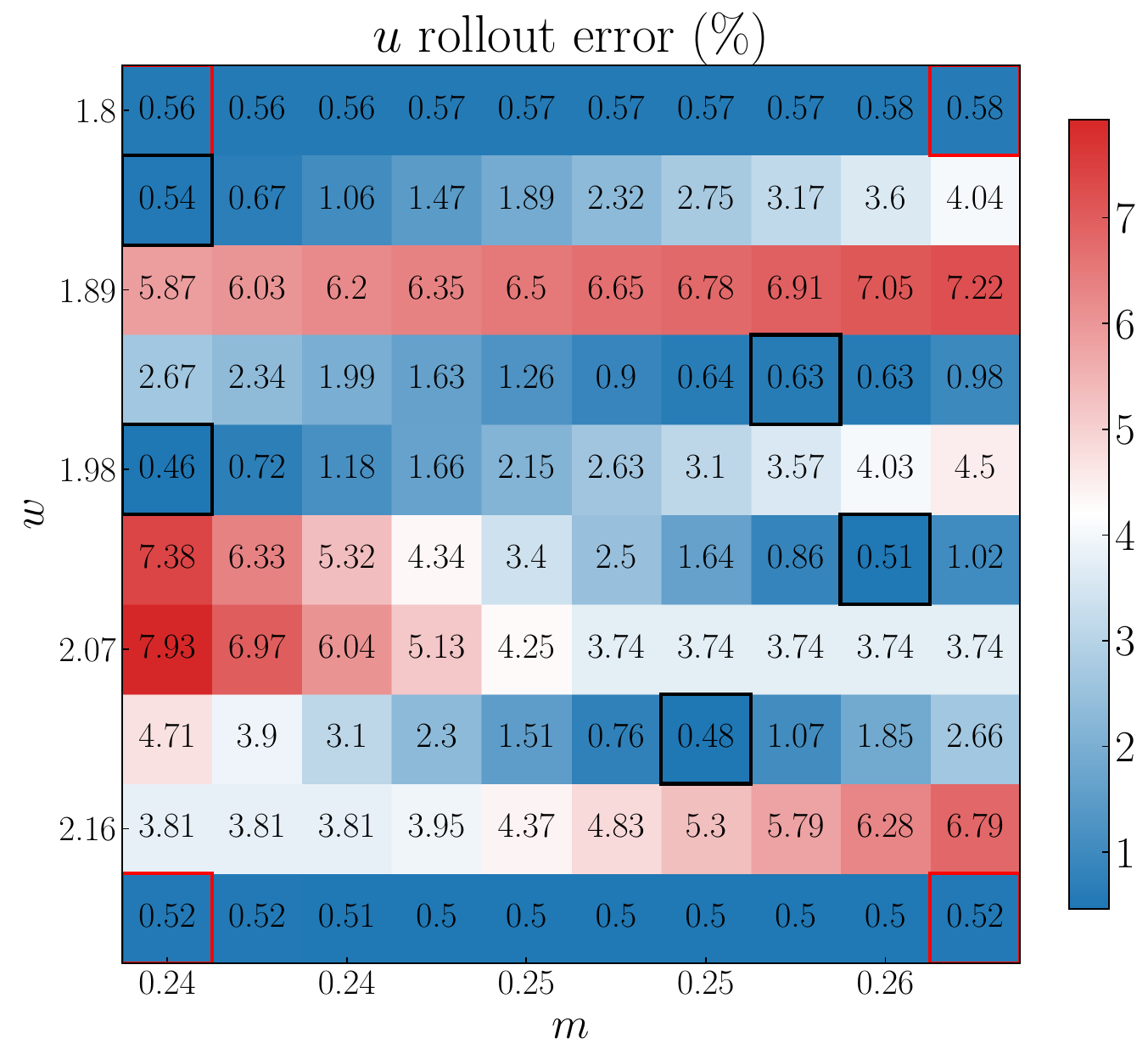}
    \includegraphics[width= 0.49\linewidth]{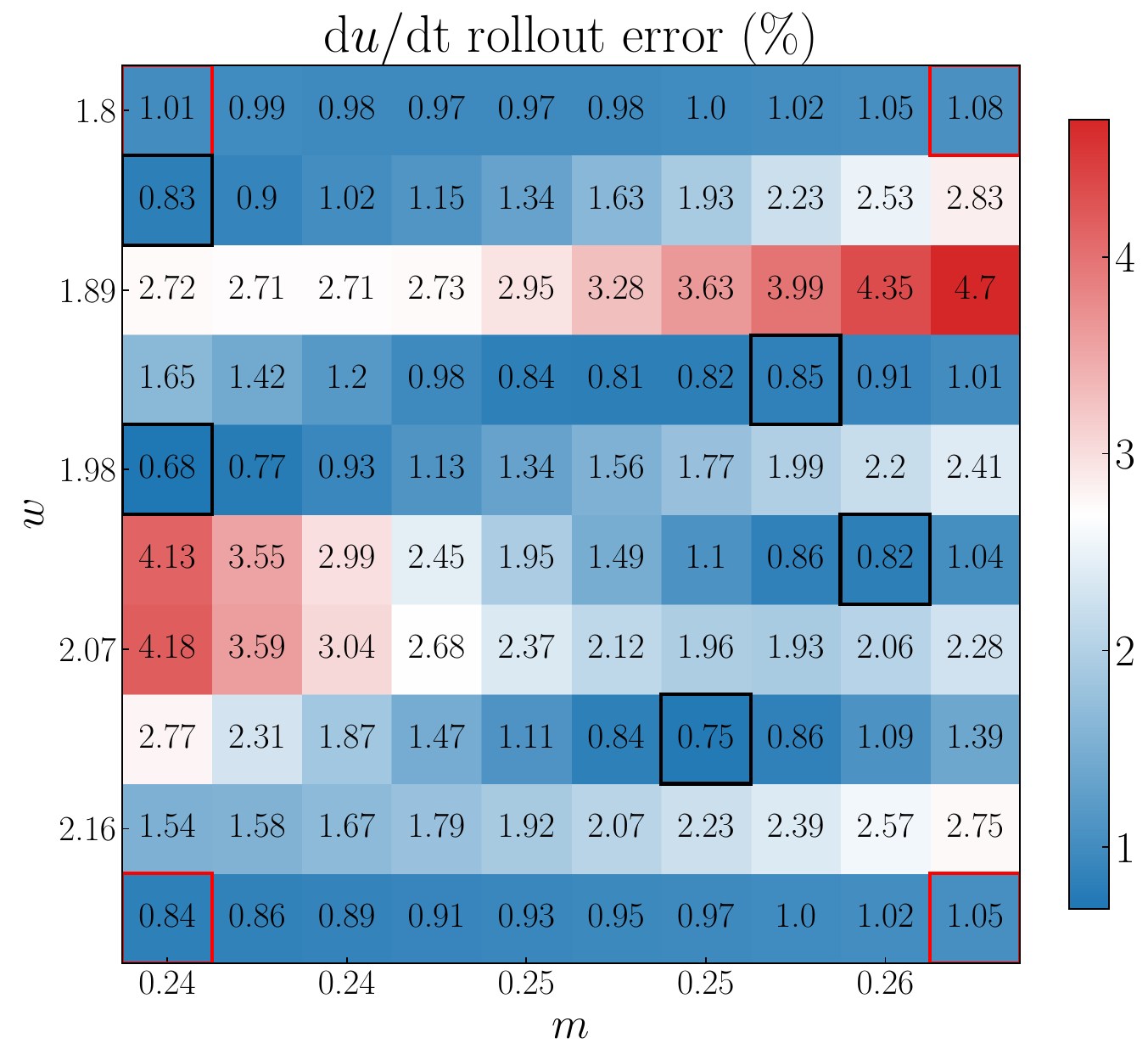}
  \caption{Relative error of (left) displacement and (right) velocity applying HLaSDI to the Klein-Gordon equation~\eqref{eq:klein}. Red squares represent initial training points and black squares represent training points chosen through greedy sampling during training. All other grid points are testing parameters.
  HLaSDI achieves errors under 8\% using only 9 data points}
  \label{fig:klein_errs}
\end{figure}

We plot one example of the FOM and Higher-order LaSDI approximations in Figure \ref{fig:klein_errs}.
Here we use a model which is the same size as wave and Telegrapher's equation examples, so again Higher-order LaSDI provides a prediction in 0.031s.
This provides 38$\times$ speedup over the FOM which takes 1.16s.

\begin{figure}
  \centering
   \includegraphics[width= \linewidth]{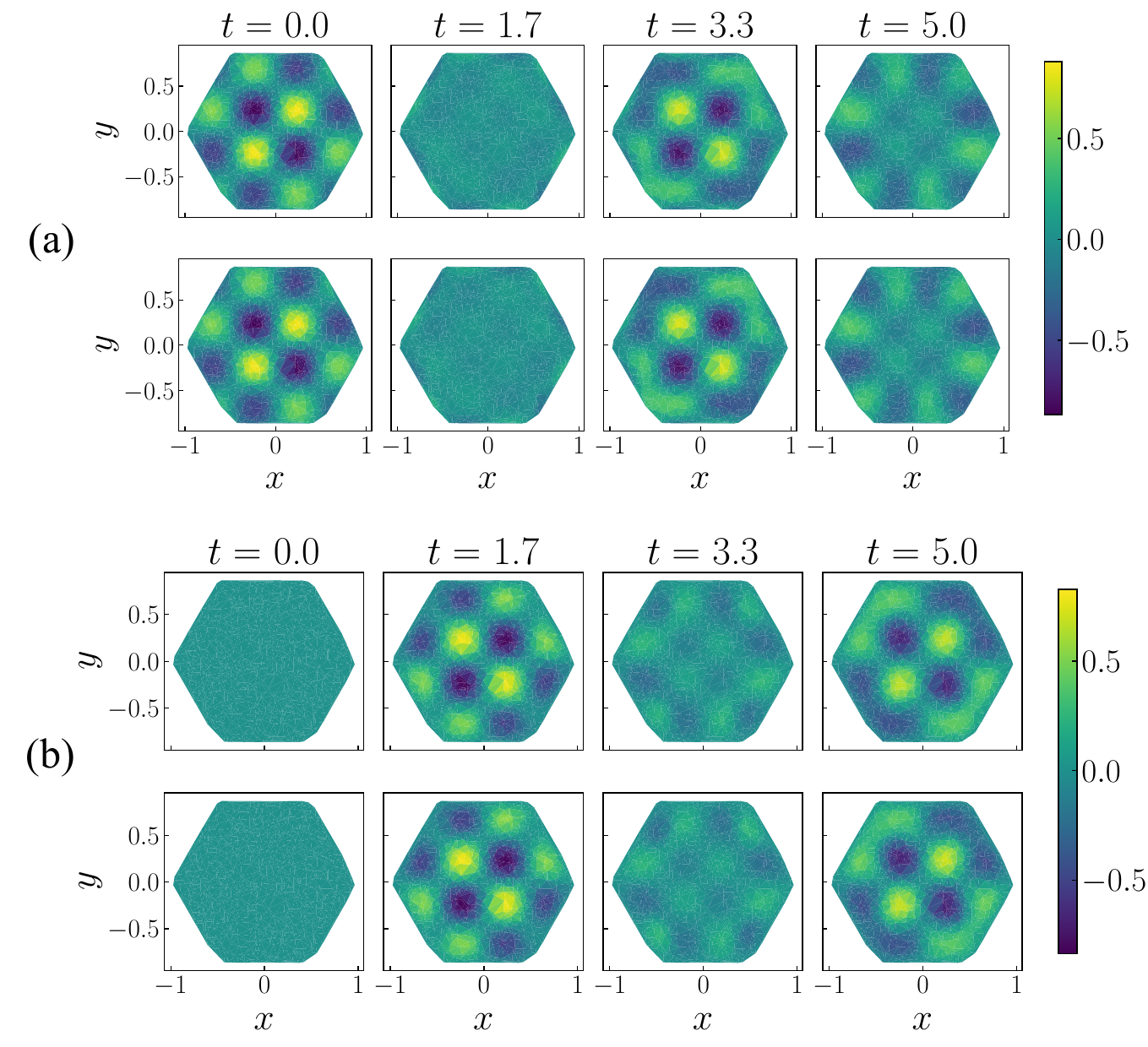}
  \caption{Solutions of FOM and Higher-order LaSDI for the Klein-Gordon equation~\eqref{eq:klein} with parameter value $(m, w) = (0.24,2.07)$. In (a) we plot the FOM values of $u$ on the top row and HLaSDI approximations of $u$ on the bottom row. Similarly, in (b) we plot FOM and HLaSDI velocities.}
  \label{fig:klein_comp}
\end{figure}


\subsection{Summary of results}
Our results show that HLaSDI achieves high prediction and reconstruction accuracy across a wide range of problems while sampling a small fraction of the parameter space.
As summarized in Table \ref{tab:results}, we achieve displacement and velocity errors below $8.1\%$ and speedups of up to 40$\times$ sampling a small portion of our parameter space.
We highlight several key observations about HLaSDI from out numerical results

\textbf{Application to first order systems:} We can apply HLaSDI even if the governing ODE is not truly second order. 
As demonstrated in the 1D Burger's equation example, HLaSDI accurately predicted the true solution and velocity of the solution through a finite difference approximation of the velocity.
This demonstrates the framework's flexibility to first order systems in addition to higher order systems.

\textbf{Damping aided learning:} The Telegrapher's equation achieved the best sample efficiency (8 training points, 20,000 iterations) among the second-order systems, compared to 11 points and 40,000 iterations for the undamped wave equation. 
This suggests that damping made our system easier to learn than conservative systems with persistent oscillations. 

\textbf{Uncertainty-based sampling:} We also emphasize the ability to perform GP-based sampling to minimize model uncertainty.
This approach ensures that we generate and sample FOM data efficiently so that our model achieves better performance.
This enables users to assess prediction reliability without access to ground truth, a critical feature for real-world deployment.

\textbf{Computational Efficiency.} Inference time for HLaSDI was under 0.032s for any model explored here, achieving 10-40$\times$ speedups over FOM solvers.
The relatively modest 10$\times$ speedup for Burger's equation (compared to 37-40$\times$ for second-order systems) reflects the simplicity of the FOM.
For more complex FOMs or higher-dimensional problems, we expect HLaSDI's speedup advantage to increase significantly.
Combined with the small training set requirements, Higher-Order LaSDI provides an efficient pathway to surrogate model construction for parameterized second order PDEs. 

\section{Discussion}
\label{Discussion}

This section discusses additional aspects of HLaSDI.
We begin by discussing several algorithmic choices: HLaSDI's use of mean absolute error in loss functions and the necessity of gradually increasing the Rollout horizon. 
Section \ref{Discussion:DtK_Z} addresses the challenges in computing the highest-order time derivative of the latent state in equation \eqref{eq:Latent:ODE:higher}. 
Section \ref{Discussion:Finite_Diff} outlines a general procedure for generating higher-order finite difference schemes for non-uniform time series. 
Finally, section \ref{Discussion:HLaSDI_GPLaSDI} demonstrates that HLaSDI generalizes Rollout-LaSDI.

\textbf{Using MAE in the loss functions:} HLaSDI uses Mean Absolute Error to compute the Reconstruction \eqref{eq:loss:recon}, Latent Dynamics \eqref{eq:loss:LD}, Consistency \eqref{eq:Loss:Consistency}, Chain-Rule \eqref{eq:Loss:Chain_Rule}, Rollout \eqref{eq:loss:rollout:higher}, and IC-Rollout \eqref{eq:Loss:IC_Rollout} losses.
Our implementation allows users to switch each loss function between MAE and MSE (with $\tfrac{1}{\sigma_{\theta_i}^2}$ replacing $\tfrac{1}{\sigma_{\theta_i}}$ for MSE).
Empirically, HLaSDI performs best with MAE loss for the Rollout losses.
Inaccurate latent dynamics lead to poor Rollout accuracy over many time horizons, producing many frames with large Rollout loss.
This distribution of losses (many outliers) favors MAE loss, which agrees with our empirical findings.
We observed no measurable penalty when switching all other loss functions to MAE and adopt this approach to simplify our algorithm's presentation.
However, further research is needed to determine whether an all-MAE approach is optimal or whether mixing MAE and MSE is superior.

\textbf{Annealing the Rollout Horizon:} HLaSDI gradually increases the maximum Rollout duration ($\Delta t_{\text{max}}^{\theta_i}$) and maximum IC-Rollout duration ($N_{\text{IC max}}(\theta_i)$).
This allows HLaSDI to focus on reconstructing FOM frames in early training, then refine the latent dynamics for long-term predictions in later stages.
Without gradual annealing, large initial Rollout durations prevent convergence; simultaneously learning to reconstruct and predict FOM frames is too challenging.
In our implementation, we increase the Rollout duration by $0.01T$ every $100$ epochs until it reaches $0.75T$, then hold it fixed.
Likewise, we increase $N_{\text{IC max}}(\theta_i)$ by $0.01 N_{t}(\theta_i)$ every $100$ epochs for the first $10{,}000$ epochs, then hold it fixed at $N_{t}(\theta_i)$.
These schedules are based on those proposed in Rollout-LaSDI \cite{stephany2025rollout}, but we do not claim they are optimal.

\subsection{\texorpdfstring{The many ways to compute $D_t^{(K)} \vec{z}_{\theta}$}{}}
\label{Discussion:DtK_Z}

To compute the Rollout, IC-Rollout, and Latent Dynamics losses, we need to compute $D_t^{(K)} \vec{z}_{\theta}$ using $\vec{z}_{\theta},\ D_t^{(1)} \vec{z}_{\theta},\ \ldots,\ D_t^{(K - 1)} \vec{z}_{\theta}$. 
As highlighted in Section \ref{Method}, when $K > 1$, there are infinitely many ways to do this.
Specifically, any affine combination (linear combination whose coefficients sum to 1) of the first time derivative of $D_t^{(K - 1)} \vec{z}_{\theta}$, the second time derivative of $D_t^{(K - 2)} \vec{z}_{\theta}$, etc., is a valid expression for $D_t^{(K)} \vec{z}_{\theta}$.

The specific combination of derivatives used to compute $D_t^{(K)} \vec{z}_{\theta}$ significantly impacts HLaSDI's performance.
For instance, when the Consistency and Chain-Rule losses are disabled ($\eta_{\text{Chain-Rule}} = \eta_{\text{Consistency}} = 0$), setting $D_t^{(K)} \vec{z}_{\theta}$ to the $K - i$'th time derivative of $D_t^{(i)} \vec{z}_{\theta}$ tends to force HLaSDI into a local minimum where the $i$'th encoder learns to encode all frames to a constant and the latent dynamics are identically zero, trivially satisfying the Latent Dynamics loss.
In this setup, the Latent Dynamics loss is zero, though the Rollout, IC-Rollout, and Reconstruction losses remain stubbornly high, causing HLaSDI to fail to converge.
Using a convex combination of the first time derivative of $D_t^{(K - 1)} \vec{z}_{\theta}$ and the second time derivative of $D_t^{(K - 2)} \vec{z}_{\theta}$ eliminates this problem, even when the Chain-Rule and Consistency losses are disabled.
With the Chain-Rule and Consistency losses enabled, HLaSDI converged in every experiment we ran, even when computing $D_t^{(K)} \vec{z}_{\theta}$ using the first time derivative of $D_t^{(K - 1)} \vec{z}_{\theta}$.

In some cases, using higher-order derivatives to compute $D_t^{(K)} \vec{z}_{\theta}$, can cause the latent dynamics to diverge.
These higher derivatives rely on more points and are prone to exacerbating errors when the encoders have not yet learned to encode the FOM frames.
This property does not appear to be fundamental, however, and we generally find that (when the Chain-Rule and Consistency losses are enabled) HLaSDI performs equally well when using either the second time derivative of $D_t^{(K - 2)} \vec{z}_{\theta}$ or the first time derivative of $D_t^{(K - 1)} \vec{z}_{\theta}$ to approximate $D_t^{(K)} \vec{z}_{\theta}$.
Further research is needed to determine whether better affine combinations of the time derivatives exist or whether approximating $D_t^{(K)} \vec{z}_{\theta}$ using the first time derivative of $D_t^{(K - 1)} \vec{z}_{\theta}$ is sufficient in all cases.

\subsection{Finite Differences}
\label{Discussion:Finite_Diff}
In this subsection, we outline a general procedure to find finite difference schemes for time series with non-uniform time step sizes, then comment on the computational properties of these schemes.

\textbf{Finite difference schemes for higher-order derivatives:} We can generalize the results of section \ref{Method:Finite_Diff} and appendix \ref{Appendix:Finite_Difference} to outline a procedure for finding finite difference schemes of order $\mathcal{O}(h^k)$ for $f^{(d)}(x)$:
\begin{enumerate}
    \item Assume $f$ is of class $C^{(k + d)}$ on $S$.
    \item Assume we know the value of $f$ at $k + d$ points in $S$, including $x$. Suppose that $j$ of these points occur before $x$ and $k + d - j - 1$ occur after $x$. Then we assume we know $f$ at $f(x - a_1 - \cdots - a_j),\ \ldots,\ f(x - a_1),\ f(x),\ f(x + b_1),\ \ldots,\ f(x + b_1 + \cdots + b_{k + d - j - 1})$, where $a_j, \ldots, a_1, b_1, \ldots, b_{k + d - j - 1} > 0$.
    \item Assume $[x - a_1 - \cdots - a_j,\  x + b_1 + \cdots + b_{k + d - j - 1}] \subseteq S$ and search for coefficients $c_{-j}, \ldots, c_{k + d - j - 1}$ such that $c_{-j} f(x - a_1 - \cdots - a_j) + \cdots + c_{k + d - j - 1} f(x + b_1 + \cdots + b_{k + d - j - 1}) = f^{(d)}(x) + \mathcal{O}(h^k)$, where $h = \max\{a_j, \ldots, a_1, b_1, \ldots, b_{k + d - j - 1}\}$.
    \item At each point (besides $x$), take a Taylor expansion of order $k + d$ of $f$ about $x$. 
    \item Substitute the expansions into the proposed linear combination. 
    \item Generate a system of equations by asserting that the coefficient of $f^{(d)}(x)$ is $1$ and that the coefficients of $f^{(j)}(x)$ for $j \in \{ 0, 1, \ldots, k + d - 1 \} \setminus \{ d \}$ are zero. This gives us a system of $k + d$ linear equations whose coefficient matrix contains polynomials of the displacements from $x$ to each of the $k + d - 1$ other points.
    \item Solving this system of equations gives us the linear combination of $f(x - a_1 - \cdots - a_j),\ \ldots,\ f(x - a_1),\ f(x),\ f(x + b_1),\ \ldots,\ f(x + b_1 + \cdots + b_{k + d - j - 1})$ that equals $f^{(d)}(x) + \mathcal{O}(h^k)$.
    \item By assuming that $a_1, \ldots, a_j, b_1, \ldots, b_{k + d - j - 1}$ are of the same order (there does not exist a pair in $\{ a_1, \ldots, a_j, b_1, \ldots, b_{k + d - j - 1} \}$ whose ratio is unbounded), we can prove that the resulting linear combination equals $f^{(d)}(x) + \mathcal{O}(h^k)$.
\end{enumerate}
In general, the coefficients $c_{-j}, \ldots, c_{k + d - j - 1}$ will be rational functions of $a_{j}, \ldots, a_j, b_1, \ldots, b_{k + d - j - 1}$ whose denominator polynomial has a greater degree than the numerator polynomial.

\textbf{Computational Considerations:} While we only considered scalar-valued functions, the schemes extend to vector-valued functions by applying the rules component-wise.
Since the coefficients are rational functions, they can be computed quickly.
Suppose we are given a time series $f(t_0), \ldots, f(t_N)$ and want to approximate $f^{(d)}$ with an $\mathcal{O}(h^k)$ finite difference scheme at each $t_i$.
We will assume we use the same scheme (e.g., central or mixed difference) at almost every time value, with forward or backward difference schemes at the edges.

For a particular coefficient in this scheme, we can use a finite number of point-wise multiplications, additions, and slices of $[f(t_0), \ldots, f(t_N)]$ to compute the numerator and denominator polynomials.
We can then compute the coefficient at each point in the time series using a single point-wise division of the vector holding the numerator polynomials by the corresponding vector for the denominator polynomials.
Thus, we can compute each coefficient for each point in the time series using a fixed number of linear passes through the vector $[f(t_0), \ldots, f(t_N)]$.
Accounting for the handful of forward or backward difference computations at the boundaries, we can compute $f^{(d)}(t_0), \ldots, f^{(d)}(t_N)$ in $\mathcal{O}(N)$ time.

Therefore, the non-uniform time step increases the runtime by only a constant (often small) factor relative to the uniform time step case.
Therefore, our approach is both general and computationally efficient.
Our implementation of HLaSDI contains optimized implementations of the three-point $\mathcal{O}(h^2)$ scheme for $f'(t)$ and the four-point $\mathcal{O}(h^2)$ scheme for $f''(t)$.

\subsection{HLaSDI for FOMs with a single time derivative}
\label{Discussion:HLaSDI_GPLaSDI}

While HLaSDI is designed to handle problems with multiple time derivatives, it is equally applicable to systems with a single time derivative.
Specifically, when $K = 1$, the Consistency and Chain-Rule losses are excluded as they no longer apply.
As a result, HLaSDI reduces to Rollout-LaSDI with a few enhancements, namely the IC-Rollout loss.
To illustrate this, we test HLaSDI on the 2D Burgers equation:
\begin{equation}
\begin{aligned}        
    \frac{\partial u}{\partial t} &= -u \frac{\partial u}{\partial x} + \nu \Delta u, \\
    u\left((x, y), 0\right) &= \exp\left(-k \left( x^2 + y^2 \right) \right) \sin\left(\pi \omega x \right) \sin(\pi \omega y) 
    \label{eq:Burgers:2D}
\end{aligned}
\end{equation}
where our input parameters are $\theta = (k, \nu)$.
For these experiments, we set $\omega = 0.5$.
The problem domain is $\Omega \times [0, T] = [-2, 2]^2 \times [0, 2]$.
The FOM consists of a nonlinear partial differential equation with two spatial variables.

To solve the FOM, we use an explicit forward Euler time integrator with $501$ uniformly spaced time steps on $[0, 2]$.
We use a centered difference spatial discretization with $31$ uniformly spaced steps along both spatial axes. 
We run FOM simulations for parameter values $k \in [0.45, 0.55]$ and $\nu \in [0.009, 0.011]$.

For training, we set $\eta_{\text{Chain-Rule}} = \eta_{\text{Consistency}} = 0$, $\eta_{\text{Recon}} = \eta_{\text{LD}} = \eta_{\text{Rollout}} = \eta_{\text{IC-Rollout}} = 1.0$, and $\eta_{\text{Coefficient}} = 0.0001$.
We use the architecture $961 - 250 - 100 - 100 - 100 - 5$. 
We use a learning rate of $10^{-3}$ and train for $20{,}000$ iterations with greedy sampling every $4{,}000$ iterations.

\begin{figure}[htbp]
    \centering
    \includegraphics[width = 0.6\linewidth]{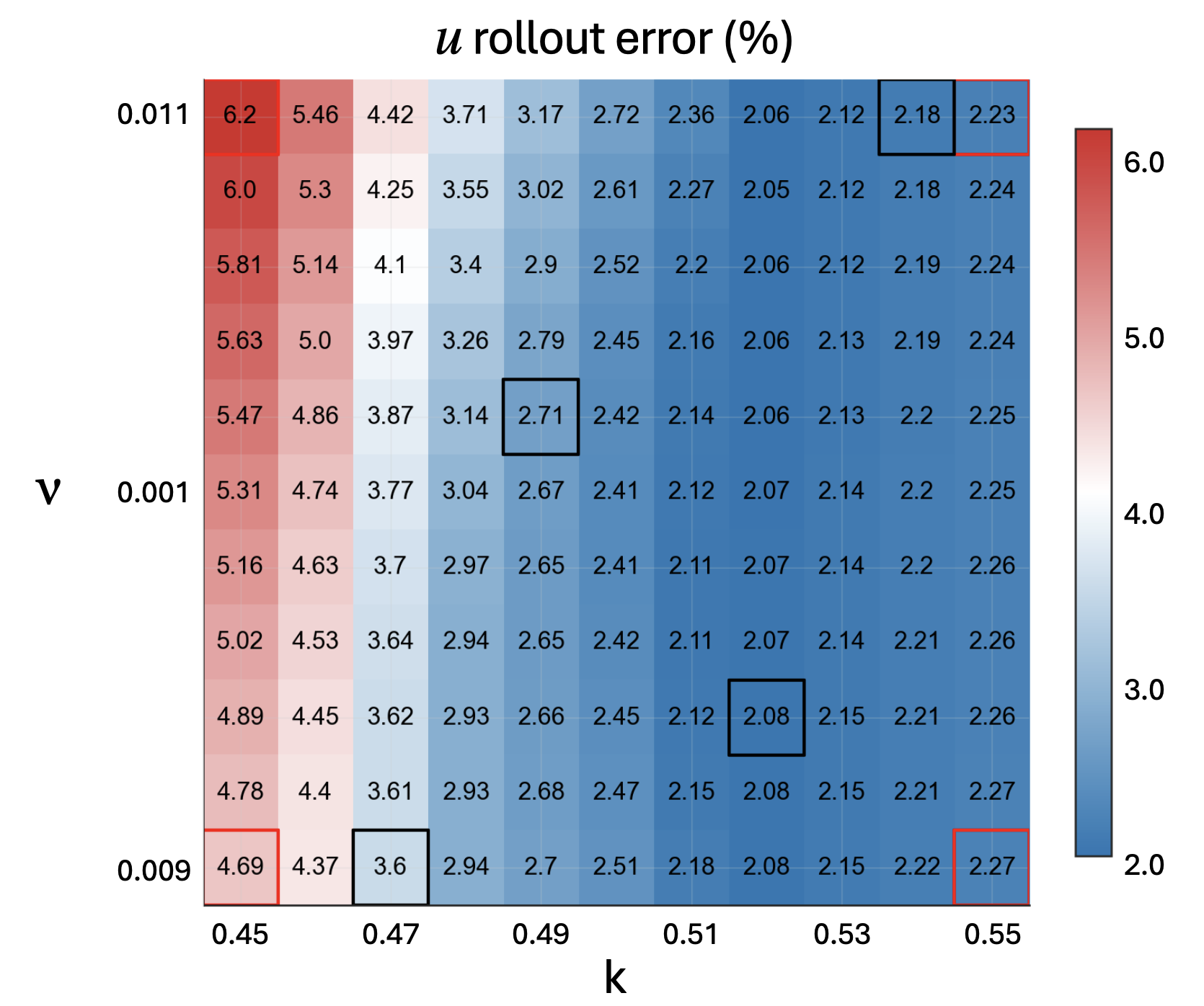}
    \caption{The relative error of the displacement reconstruction when applying HLaSDI to the 2D Burgers equation \eqref{eq:Burgers:2D}. 
    Red squares represent initial training points and black squares represent training points chosen through greedy sampling during training. 
    HLaSDI achieves errors under 6.2\% using 8 data points.}
    \label{fig:Burgers:2D}
\end{figure}

Figure \ref{fig:Burgers:2D} shows that HLaSDI achieves a maximum relative error of $6.2\%$ across the parameter space, demonstrating that our model achieves accurate predictions across all parameter values.
Thus, while HLaSDI is designed for systems with multiple time derivatives, its pair of Rollout losses makes it a powerful architecture for systems with a single time derivative.
\section{Conclusion}
\label{Conclusion}

We have introduced Higher-Order LaSDI (HLaSDI), a data-driven, non-intrusive reduced order model that extends Rollout-LaSDI to FOMs with an arbitrary number of time derivatives.
HLaSDI learns a ROM capable of solving complex, multi-derivative parameterized FOMs using a relatively limited set of training data (only a few parameter values).
For a FOM with $K$ time derivatives, HLaSDI consists of $K$ autoencoders, where the $k$'th encoder/decoder handles the $k - 1$'th time derivative of the FOM.
HLaSDI connects the latent states of the $K$ encoders through a joint set of latent dynamics, consisting of a $K$'th order dynamical system that uses each encoder's latent state to define the right-hand side.
We define a collection of novel loss functions, including the Consistency, Chain-Rule, and IC-Rollout losses, which encode the mathematical structure of the FOM into HLaSDI's architecture, providing a powerful inductive bias for learning ROM approximations to complex FOMs.

We demonstrated the efficacy of HLaSDI on a collection of linear and nonlinear Full Order Models from multiple problem domains.
For all examples, HLaSDI obtained accurate predictions across all testing parameter values.
Our algorithm extends the LaSDI framework to realistic, multi-derivative domains.
This extension and its novel loss functions offer an exciting opportunity for expanding the scope and reach of reduced order modeling.

\section{Acknowledgments}

Robert Stephany was supported by the Sydney Fernbach Postdoctoral Fellowship under LDRD number 25-ERD-
049.
William Anderson work was partially supported by the Lawrence Livermore National Laboratory (LLNL) under Project No. 50284.
Youngsoo Choi was supported by the U.S. Department of Energy (DOE), Office of Science, Office of Advanced Scientific Computing Research (ASCR), through the CHaRMNET Mathematical Multifaceted Integrated Capability Center (MMICC) under Award Number DE-SC0023164 and the LEADS SciDAC Institute under Project Number SCW1933.
This work was performed under the auspices of the U.S. Department of Energy by Lawrence Livermore National Laboratory under Contract DE-AC52-07NA27344.
LLNL document release number: LLNL-JRNL-2013669-DRAFT.

\bibliographystyle{plainnat}
\bibliography{Bibliography}

@article{anderson2021mfem,
  title         = {MFEM: A modular finite element methods library},
  author={Anderson, Robert and Andrej, Julian and Barker, Andrew and Bramwell, Jamie and Camier, Jean-Sylvain and Cerveny, Jakub and Dobrev, Veselin and Dudouit, Yohann and Fisher, Aaron and Kolev, Tzanio and others},
  journal       = {Computers \& Mathematics with Applications},
  volume        = {81},
  pages         = {42--74},
  year          = {2021},
  publisher     = {Elsevier}
}

@article{andrej2024mfem_exascale,
  title         = {High-performance finite elements with MFEM},
  author        = {Andrej, Julian and Atallah, Nabil and B{\"a}cker, Jan-Phillip and Camier, Jean-Sylvain and Copeland, Dylan and Dobrev, Veselin and Dudouit, Yohann and Duswald, Tobias and Keith, Brendan and Kim, Dohyun and others},
  journal       = {The International Journal of High Performance Computing Applications},
  volume        = {38},
  number        = {5},
  pages         = {447--467},
  year          = {2024},
  publisher     = {SAGE Publications Sage UK: London, England}
}

@inproceedings{jasak2007openfoam,
  title={OpenFOAM: A C++ library for complex physics simulations},
  author={Jasak, Hrvoje and Jemcov, Aleksandar and Tukovic, Zeljko and others},
  booktitle={International workshop on coupled methods in numerical dynamics},
  volume={1000},
  pages={1--20},
  year={2007},
  organization={Dubrovnik, Croatia)}
}

@book{thijssen2007comp_physics,
  title={Computational physics},
  author={Thijssen, Jos},
  year={2007},
  publisher={Cambridge university press}
}

@article{hinton2006networks,
  title         = {A fast learning algorithm for deep belief nets},
  author        = {Hinton, Geoffrey E and Osindero, Simon and Teh, Yee-Whye},
  journal       = {Neural computation},
  volume        = {18},
  number        = {7},
  pages         = {1527--1554},
  year          = {2006},
  publisher     = {MIT Press}
}

@inproceedings{glorot2010initialization,
  title         = {Understanding the difficulty of training deep feedforward neural networks},
  author        = {Glorot, Xavier and Bengio, Yoshua},
  booktitle     = {Proceedings of the thirteenth international conference on artificial intelligence and statistics},
  pages         = {249--256},
  year          = {2010},
  organization  = {JMLR Workshop and Conference Proceedings}
}

@article{kingma2014adam,
  title         = {Adam: A method for stochastic optimization},
  author        = {Kingma, Diederik P and Ba, Jimmy},
  journal       = {arXiv preprint arXiv:1412.6980},
  year          = {2014}
}

@book{williams2006gaussian,
  title={Gaussian processes for machine learning},
  author={Williams, Christopher KI and Rasmussen, Carl Edward},
  volume={2},
  year={2006},
  publisher={MIT press Cambridge, MA}
}

@article{paszke2019pytorch,
  title         = {Pytorch: An imperative style, high-performance deep learning library},
  author        = {Paszke, Adam and Gross, Sam and Massa, Francisco and Lerer, Adam and Bradbury, James and Chanan, Gregory and Killeen, Trevor and Lin, Zeming and Gimelshein, Natalia and Antiga, Luca and others},
  journal       = {Advances in neural information processing systems},
  volume        = {32},
  year          = {2019}
}

@inproceedings{abadi2016tensorflow,
  title         = {$\{$TensorFlow$\}$: a system for $\{$Large-Scale$\}$ machine learning},
  author        = {Abadi, Mart{\'\i}n and Barham, Paul and Chen, Jianmin and Chen, Zhifeng and Davis, Andy and Dean, Jeffrey and Devin, Matthieu and Ghemawat, Sanjay and Irving, Geoffrey and Isard, Michael and others},
  booktitle     = {12th USENIX symposium on operating systems design and implementation (OSDI 16)},
  pages         = {265--283},
  year          = {2016}
}

@inproceedings{he2016residual,
  title={Deep residual learning for image recognition},
  author={He, Kaiming and Zhang, Xiangyu and Ren, Shaoqing and Sun, Jian},
  booktitle={Proceedings of the IEEE conference on computer vision and pattern recognition},
  pages={770--778},
  year={2016}
}

@article{krizhevsky2017alexnet,
  title={ImageNet classification with deep convolutional neural networks},
  author={Krizhevsky, Alex and Sutskever, Ilya and Hinton, Geoffrey E},
  journal={Communications of the ACM},
  volume={60},
  number={6},
  pages={84--90},
  year={2017},
  publisher={AcM New York, NY, USA}
}

@article{vaswani2017attention,
  title={Attention is all you need},
  author={Vaswani, Ashish and Shazeer, Noam and Parmar, Niki and Uszkoreit, Jakob and Jones, Llion and Gomez, Aidan N and Kaiser, {\L}ukasz and Polosukhin, Illia},
  journal={Advances in neural information processing systems},
  volume={30},
  year={2017}
}

@article{brown2020gpt3,
  title={Language models are few-shot learners},
  author={Brown, Tom and Mann, Benjamin and Ryder, Nick and Subbiah, Melanie and Kaplan, Jared D and Dhariwal, Prafulla and Neelakantan, Arvind and Shyam, Pranav and Sastry, Girish and Askell, Amanda and others},
  journal={Advances in neural information processing systems},
  volume={33},
  pages={1877--1901},
  year={2020}
}

@article{rackauckas2020ude,
  title         = {Universal differential equations for scientific machine learning},
  author        = {Rackauckas, Christopher and Ma, Yingbo and Martensen, Julius and Warner, Collin and Zubov, Kirill and Supekar, Rohit and Skinner, Dominic and Ramadhan, Ali and Edelman, Alan},
  journal       = {arXiv preprint arXiv:2001.04385},
  year          = {2020}
}

@article{raissi2019pinn,
  title         = {Physics-informed neural networks: A deep learning framework for solving forward and inverse problems involving nonlinear partial differential equations},
  author        = {Raissi, Maziar and Perdikaris, Paris and Karniadakis, George E},
  journal       = {Journal of Computational physics},
  volume        = {378},
  pages         = {686--707},
  year          = {2019},
  publisher     = {Elsevier}
}

@article{brunton2016sindy,
  title         = {Discovering governing equations from data by sparse identification of nonlinear dynamical systems},
  author        = {Brunton, Steven L and Proctor, Joshua L and Kutz, J Nathan},
  journal       = {Proceedings of the national academy of sciences},
  volume        = {113},
  number        = {15},
  pages         = {3932--3937},
  year          = {2016},
  publisher     = {National Academy of Sciences}
}

@article{rudy2017pdefind,
  title         = {Data-driven discovery of partial differential equations},
  author        = {Rudy, Samuel H and Brunton, Steven L and Proctor, Joshua L and Kutz, J Nathan},
  journal       = {Science advances},
  volume        = {3},
  number        = {4},
  pages         = {e1602614},
  year          = {2017},
  publisher     = {American Association for the Advancement of Science}
}

@article{stephany2022pde_read,
  title={PDE-READ: Human-readable partial differential equation discovery using deep learning},
  author={Stephany, Robert and Earls, Christopher},
  journal={Neural Networks},
  volume={154},
  pages={360--382},
  year={2022},
  publisher={Elsevier}
}

@article{stephany2024pde_learn,
  title={PDE-LEARN: Using deep learning to discover partial differential equations from noisy, limited data},
  author={Stephany, Robert and Earls, Christopher},
  journal={Neural Networks},
  volume={174},
  pages={106242},
  year={2024},
  publisher={Elsevier}
}

@article{stephany2025dde_find,
  title={DDE-Find: learning delay differential equations from noisy, limited data},
  author={Stephany, Robert},
  journal={Proceedings of the Royal Society A},
  volume={481},
  number={2310},
  pages={20240403},
  year={2025},
  publisher={The Royal Society}
}

@article{lusch2018koopman,
  title={Deep learning for universal linear embeddings of nonlinear dynamics},
  author={Lusch, Bethany and Kutz, J Nathan and Brunton, Steven L},
  journal={Nature communications},
  volume={9},
  number={1},
  pages={4950},
  year={2018},
  publisher={Nature Publishing Group UK London}
}

@article{berkooz1993pod,
  title         = {The proper orthogonal decomposition in the analysis of turbulent flows},
  author        = {Berkooz, Gal and Holmes, Philip and Lumley, John L},
  journal       = {Annual review of fluid mechanics},
  volume        = {25},
  number        = {1},
  pages         = {539--575},
  year          = {1993},
  publisher     = {Annual Reviews 4139 El Camino Way, PO Box 10139, Palo Alto, CA 94303-0139, USA}
}

@article{schmid2010dmd,
  title         = {Dynamic mode decomposition of numerical and experimental data},
  author        = {Schmid, Peter J},
  journal       = {Journal of fluid mechanics},
  volume        = {656},
  pages         = {5--28},
  year          = {2010},
  publisher     = {Cambridge University Press}
}

@article{iliescu2014NavierStokesPOD,
  title={Variational multiscale proper orthogonal decomposition: Navier-stokes equations},
  author={Iliescu, Traian and Wang, Zhu},
  journal={Numerical Methods for Partial Differential Equations},
  volume={30},
  number={2},
  pages={641--663},
  year={2014},
  publisher={Wiley Online Library}
}

@article{stabile2018NavierStokesPOD,
  title={Finite volume POD-Galerkin stabilised reduced order methods for the parametrised incompressible Navier--Stokes equations},
  author={Stabile, Giovanni and Rozza, Gianluigi},
  journal={Computers \& Fluids},
  volume={173},
  pages={273--284},
  year={2018},
  publisher={Elsevier}
}

@article{choi2019DesignROM,
  title={Accelerating design optimization using reduced order models},
  author={Choi, Youngsoo and Oxberry, Geoffrey and White, Daniel and Kirchdoerfer, Trenton},
  journal={arXiv preprint arXiv:1909.11320},
  year={2019}
}

@article{mcbane2021DesignROM,
  title={Component-wise reduced order model lattice-type structure design},
  author={McBane, Sean and Choi, Youngsoo},
  journal={Computer methods in applied mechanics and engineering},
  volume={381},
  pages={113813},
  year={2021},
  publisher={Elsevier}
}

@article{mclaughlin2016AdvectionDiffusionROM,
  title={Stabilized reduced order models for the advection--diffusion--reaction equation using operator splitting},
  author={McLaughlin, Benjamin and Peterson, Janet and Ye, Ming},
  journal={Computers \& Mathematics with Applications},
  volume={71},
  number={11},
  pages={2407--2420},
  year={2016},
  publisher={Elsevier}
}

@article{kim2021AdvectionDiffusionROMs,
  title={Efficient space--time reduced order model for linear dynamical systems in python using less than 120 lines of code},
  author={Kim, Youngkyu and Wang, Karen and Choi, Youngsoo},
  journal={Mathematics},
  volume={9},
  number={14},
  pages={1690},
  year={2021},
  publisher={MDPI}
}

@article{champion2019sindy_autoencoder,
  title         = {Data-driven discovery of coordinates and governing equations},
  author        = {Champion, Kathleen and Lusch, Bethany and Kutz, J Nathan and Brunton, Steven L},
  journal       = {Proceedings of the National Academy of Sciences},
  volume        = {116},
  number        = {45},
  pages         = {22445--22451},
  year          = {2019},
  publisher     = {National Academy of Sciences}
}

@article{kaiser2018sindy_mpc,
  title         = {Sparse identification of nonlinear dynamics for model predictive control in the low-data limit},
  author        = {Kaiser, Eurika and Kutz, J Nathan and Brunton, Steven L},
  journal       = {Proceedings of the Royal Society A},
  volume        = {474},
  number        = {2219},
  pages         = {20180335},
  year          = {2018},
  publisher     = {The Royal Society Publishing}
}

@article{gorban2018manifold_hypothesis,
  title         = {Blessing of dimensionality: mathematical foundations of the statistical physics of data},
  author        = {Gorban, Alexander N and Tyukin, Ivan Yu},
  journal       = {Philosophical Transactions of the Royal Society A: Mathematical, Physical and Engineering Sciences},
  volume        = {376},
  number        = {2118},
  pages         = {20170237},
  year          = {2018},
  publisher     = {The Royal Society Publishing}
}

@article{chen2018NODE,
  title={Neural ordinary differential equations},
  author={Chen, Ricky TQ and Rubanova, Yulia and Bettencourt, Jesse and Duvenaud, David K},
  journal={Advances in neural information processing systems},
  volume={31},
  year={2018}
}

@article{rubanova2019latent_NODE,
  title={Latent ordinary differential equations for irregularly-sampled time series},
  author={Rubanova, Yulia and Chen, Ricky TQ and Duvenaud, David K},
  journal={Advances in neural information processing systems},
  volume={32},
  year={2019}
}

@article{lee2021parameterized_NODE,
  title={Parameterized neural ordinary differential equations: Applications to computational physics problems},
  author={Lee, Kookjin and Parish, Eric J},
  journal={Proceedings of the Royal Society A},
  volume={477},
  number={2253},
  pages={20210162},
  year={2021},
  publisher={The Royal Society}
}

@article{bonneville2023gplasdi_neuralips,
  title         = {Data-Driven Autoencoder Numerical Solver with Uncertainty Quantification for Fast Physical Simulations},
  author        = {Bonneville, Christophe and Choi, Youngsoo and Ghosh, Debojyoti and Belof, Jonathan L},
  journal       = {arXiv preprint arXiv:2312.01021},
  year          = {2023}
}

@article{fries2022lasdi,
  title         = {{LaSDI}: Parametric latent space dynamics identification},
  author        = {Fries, William D and He, Xiaolong and Choi, Youngsoo},
  journal       = {Computer Methods in Applied Mechanics and Engineering},
  volume        = {399},
  pages         = {115436},
  year          = {2022},
  publisher     = {Elsevier}
}

@article{tran2024wlasdi,
  title         = {Weak-form latent space dynamics identification},
  author        = {Tran, April and He, Xiaolong and Messenger, Daniel A and Choi, Youngsoo and Bortz, David M},
  journal       = {Computer Methods in Applied Mechanics and Engineering},
  volume        = {427},
  pages         = {116998},
  year          = {2024},
  publisher     = {Elsevier}
}

@article{he2023glasdi,
  title         = {{gLaSDI}: Parametric physics-informed greedy latent space dynamics identification},
  author        = {He, Xiaolong and Choi, Youngsoo and Fries, William D and Belof, Jonathan L and Chen, Jiun-Shyan},
  journal       = {Journal of Computational Physics},
  volume        = {489},
  pages         = {112267},
  year          = {2023},
  publisher     = {Elsevier}
}

@article{bonneville2024gplasdi,
  title         = {{GPLaSDI}: Gaussian process-based interpretable latent space dynamics identification through deep autoencoder},
  author        = {Bonneville, Christophe and Choi, Youngsoo and Ghosh, Debojyoti and Belof, Jonathan L},
  journal       = {Computer Methods in Applied Mechanics and Engineering},
  volume        = {418},
  pages         = {116535},
  year          = {2024},
  publisher     = {Elsevier}
}

@article{park2024tlasdi,
  title         = {tLaSDI: Thermodynamics-informed latent space dynamics identification},
  author        = {Park, Jun Sur Richard and Cheung, Siu Wun and Choi, Youngsoo and Shin, Yeonjong},
  journal       = {Computer Methods in Applied Mechanics and Engineering},
  volume        = {429},
  pages         = {117144},
  year          = {2024},
  publisher     = {Elsevier}
}

@article{anderson2025mlasdi,
  title={mLaSDI: Multi-stage latent space dynamics identification},
  author={Anderson, William and Chung, Seung Whan and Choi, Youngsoo},
  journal={arXiv preprint arXiv:2506.09207},
  year={2025}
}

@article{chung2025LaSDI_it,
  title={Latent Space Dynamics Identification for Interface Tracking with Application to Shock-Induced Pore Collapse},
  author={Chung, Seung Whan and Miller, Christopher and Choi, Youngsoo and Tranquilli, Paul and Springer, H Keo and Sullivan, Kyle},
  journal={arXiv preprint arXiv:2507.10647},
  year={2025}
}

@article{he2025wgLaSDI,
  title={Physics-Informed Active Learning With Simultaneous Weak-Form Latent Space Dynamics Identification},
  author={He, Xiaolong and Tran, April and Bortz, David M and Choi, Youngsoo},
  journal={International Journal for Numerical Methods in Engineering},
  volume={126},
  number={1},
  pages={e7634},
  year={2025},
  publisher={Wiley Online Library}
}

@article{bonneville2024lasdi_review,
  title         = {A comprehensive review of latent space dynamics identification algorithms for intrusive and non-intrusive reduced-order-modeling},
  author        = {Bonneville, Christophe and He, Xiaolong and Tran, April and Park, Jun Sur and Fries, William and Messenger, Daniel A and Cheung, Siu Wun and Shin, Yeonjong and Bortz, David M and Ghosh, Debojyoti and others},
  journal       = {arXiv preprint arXiv:2403.10748},
  year          = {2024}
}

@article{stephany2025rollout,
  title={Rollout-LaSDI: Enhancing the long-term accuracy of Latent Space Dynamics},
  author={Stephany, Robert and Choi, Youngsoo},
  journal={arXiv preprint arXiv:2509.08191},
  year={2025}
}

@misc{burden1997numerical,
  title         = {Numerical analysis, brooks},
  author        = {Burden, Richard L and Faires, J Douglas},
  year          = {1997},
  publisher     = {Cole publishing company Pacific Grove, CA:}
}

@article{sklearn,
author = {Pedregosa, F. and Varoquaux, G. and Gramfort, A. and Michel, V. and Thirion, B. and Grisel, O. and Blondel, M. and Prettenhofer, P. and Weiss, R. and Dubourg, V. and Vanderplas, J. and Passos, A. and Cournapeau, D. and Brucher, M. and Perrot, M. and Duchesnay, \'{E}.},
title = {Scikit-learn: Machine Learning in Python},
year = {2011},
issue_date = {2/1/2011},
publisher = {JMLR.org},
volume = {12},
number = {null},
issn = {1532-4435},
journal = {J. Mach. Learn. Res.},
month = nov,
pages = {2825–2830},
numpages = {6}
}

@book{hayt2011engineering,
  title={Engineering Electromagnetics},
  author={Hayt, W.H. and Buck, J.A.},
  isbn={9780071317078},
  lccn={2010048332},
  year={2011},
  publisher={McGraw-Hill}
}

@book{greiner2012relativistic,
  title={Relativistic Quantum Mechanics: Wave Equations},
  author={Greiner, W.},
  isbn={9783642880827},
  year={2012},
  publisher={Springer Berlin Heidelberg}
}

\appendix
\section{Derivation of Finite Difference Schemes}
\label{Appendix:Finite_Difference}

In this appendix, we derive the finite difference schemes we introduced in section \ref{Method:Finite_Diff}.
Throughout this section, we will let $S \subseteq \mathbb{R}$ be an open subset of $\mathbb{R}$.
Likewise, let $f : S \to \mathbb{R}$.

\subsection{Three-point finite-difference approximations of \texorpdfstring{$f'(x)$}{f'(x)}}
\label{Appendix:Finite_Difference:First}

Here, we will derive, forward- and central-difference approximations of $f'(x)$.
Throughout this sub-section, we assume $f$ is of class $C^3$ on $S$.
Let $a, b > 0$ and $h = \max\{a, b\}$.

\subsubsection{Three point forward-difference approximation of \texorpdfstring{$f'(x)$}{f'(x)}.}
\label{Appendix:Finite_Difference:First:Forward}

Suppose we know $f(x)$, $f(x + a)$, and $f(x + a + b)$.
We assume that $[x, x + a + b] \subseteq S$.
We seek coefficients $c_0$, $c_1$, $c_2$ such that
\begin{equation}
    c_0 f(x) + c_1 f(x + a) + c_2 f(x + a + b) = f'(x) + \mathcal{O}(h^2)
    \label{eq:Appendix:Finite_Diff:First:Forward}
\end{equation}
We will also assume that there is some $C > 0$ such that $\tfrac{a}{b} \leq C$.
Thus, we do not allow $b$ to be arbitrary small relative to $a$; this is a natural requirement, as allowing $\tfrac{a}{b}$ to be too small would mean that the points $x + a$ and $x + a + b$ are effectively identical.
In this case, $f(x + a)$ and $f(x + a + b)$ provide essentially identical information, thereby meaning that the collection $f(x)$, $f(x + a)$, and $f(x + a + b)$ contains too little information to properly define a finite difference scheme.
With this established, let us take a third-order Taylor expansion of $f(x + a)$ and $f(x + a + b)$ about $x$:
\begin{alignat*}{3}
    f(x + a)        &= f(x) + a       f'(x) &&   + \tfrac{1}{2} a^2        f''(x) &&  + \tfrac{1}{6} a^3       f'''(\xi_1),\\
    f(x + a + b)    &= f(x) + (a + b) f'(x) &&   + \tfrac{1}{2}(a + b)^2 f''(x) &&  + \tfrac{1}{6} (a + b)^3 f'''(\xi_2),
\end{alignat*}
for some $\xi_1 \in (x, x + a)$ and $\xi_2 \in (x + a, x + a + b)$. 
Substituting this into equation \eqref{eq:Appendix:Finite_Diff:First:Forward} gives
\begin{equation}
\begin{aligned}
    c_0 f(x) + c_1 f(x + a) + c_2 f(x + a + b) = 
                   &\left(c_0   +   c_1                 +   c_2                         \right) f(x) + \\
                   &\left(          c_1 a               +   c_2 (a + b)                 \right) f'(x) + \\
       \tfrac{1}{2}&\left(          c_1 a^2             +   c_2 (a + b)^2               \right) f''(x) + \\
       \tfrac{1}{6}&\left(          c_1 a^3 f'''(\xi_1) +   c_2 (a + b)^3 f'''(\xi_2)   \right).
\end{aligned}
\label{eq:Appendix:Finite_Diff:First:Forward:Taylor}
\end{equation}
For this to have any hope of equaling $f'(x) + \mathcal{O}(h^2)$, we need the expression in front of $f(x)$ and $f''(x)$ to disappear and the one in front of $f'(x)$ to be $1$. 
In other words, we must have
\begin{equation*}
    \begin{bmatrix}
        1, &&1,     &&1             \\
        0, &&a,     &&(a + b)   \\
        0, &&a^2,   &&(a + b)^2 
    \end{bmatrix} 
    \begin{bmatrix}
        c_0 \\
        c_1 \\
        c_2
    \end{bmatrix}
    = 
    \begin{bmatrix}
        0 \\
        1 \\
        0
    \end{bmatrix}.
\end{equation*}
Brute force computation (row-reduction) shows that this system has the following solution:
\begin{equation}
\begin{aligned}
    c_0 &= -\frac{\left(2 a + b\right)}{a(a + b)}, \\
    c_1 &= \frac{a + b}{a b}, \\
    c_2 &= -\frac{a}{b\left( a + b \right)}.
\end{aligned}
\label{eq:Appendix:Finite_Diff:First:Forward:Coefficients}
\end{equation}
Plugging these into equation \eqref{eq:Appendix:Finite_Diff:First:Forward:Taylor} gives
\begin{equation*}
    c_0 f(x) + c_1 f(x + a) + c_2 f(x + a + b) = f'(x) + \frac{1}{6} \left( \frac{(a + b)a^2}{b} f'''\left(\xi_1\right) - \frac{(a + b)^2a}{b} f'''\left(\xi_2\right) \right)
\end{equation*}
Because $f$ is of class $C^3$ on $S$ and because $[x, x + a + b] \subseteq S$ is compact, $f'''$ must be bounded on $[x, x + a + b]$.
Let $M > 0$ denote an upper bound. 
Then, we must have
\begin{equation*}
\begin{aligned}
    \frac{\left| c_0 f(x) + c_1 f(x + a) + c_2 f(x + a + b) - f'(x)\right|}{h^2} &= \frac{1}{6} \frac{\left|\left( \frac{(a + b)a^2}{b} f'''\left(\xi_1\right) - \frac{(a + b)^2a}{b} f'''\left(\xi_2\right) \right)\right|}{h^2} \\
    &\leq \frac{1}{6} M \left( \frac{(a + b)a^2}{b h^2}  + \frac{(a + b)^2 a}{b h^2} \right)
\end{aligned}
\end{equation*}
However, because $h = \max\{a, b\}$, we must have $h \geq a$, $2h \geq a + b$, which means that
\begin{equation*}
\begin{aligned}
    \frac{\left| c_0 f(x) + c_1 f(x + a) + c_2 f(x + a + b) - f'(x)\right|}{h^2} &\leq \frac{1}{6} M \left( \frac{(a + b)h^2}{b h^2} + \frac{2(a + b)h^2}{b h^2} \right) \\
    &= \frac{1}{2} M \frac{(a + b)}{b} \\
    &\leq \frac{1}{2} M (C + 1)
\end{aligned}
\end{equation*}
Taking the limit as $h$ goes to zero tells us that the derived scheme is indeed of order $\mathcal{O}(h^2)$.

\subsubsection{Three point central-difference approximation of \texorpdfstring{$f'(x)$}{f'(x)}.}
\label{Appendix:Finite_Difference:First:Central}

Next, suppose we know $f(x - a)$, $f(x)$, and $f(x + b)$.
We assume that $[x - a, x + b] \subseteq S$.
We seek coefficients $c_{-1}, c_0, c_1 \in \mathbb{R}$ such that
\begin{equation}
    c_{-1} f(x - a) + c_0 f(x) + c_1 f(x + b) = f'(x) + \mathcal{O}(h^2)
    \label{eq:Appendix:Finite_Diff:First:Central}
\end{equation}
Once again, we begin by taking third-order taylor expansions of $f(x - a)$ and $f(x + b)$ about $x$:
\begin{equation*}
\begin{aligned}
    f(x - a) &= f(x) - a f'(x) + \tfrac{1}{2} a^2 f''(x) - \tfrac{1}{6} a^3 f'''\left(\xi_{-1} \right), \\
    f(x + b) &= f(x) + b f'(x) + \tfrac{1}{2} b^2 f''(x) + \tfrac{1}{6} b^3 f'''\left( \xi_1 \right),
\end{aligned}
\end{equation*}
where $\xi_{-1} \in (x - a, x)$ and $\xi_1 \in (x, x + b)$.
Thus, we have
\begin{equation*}
\begin{aligned}    
    c_{-1} f(x - a) + c_0 f(x) + c_1 f(x + b) =& 
                    \left( c_{-1}                   +   c_0     + c_1                   \right) f(x) \\
    &+              \left(-c_{-1}a                              + c_1 b                 \right) f'(x) \\
    &+  \tfrac{1}{2}\left( c_{-1}a^2                            + c_1 b^2               \right) f''(x) \\
    &+  \tfrac{1}{6}\left(-c_{-1}a^3 f'''(\xi_{-1})             + c_1 b^3 f'''(\xi_1)   \right)
\end{aligned}
\end{equation*}
For this to equal $f'(x) + \mathcal{O}(h^2)$, we need the coefficients in front of $f(x)$ and $f''(x)$ to disappear and the one in front of $f'(x)$ to equal $1$. 
In other words, we must have
\begin{equation*}
    \begin{bmatrix}
        1,      && 1,    && 1       \\
        -a,     && 0,    && b       \\
        a^2,    && 0,    && b^2 
    \end{bmatrix}    
    \begin{bmatrix}
        c_{-1} \\
        c_0 \\
        c_1
    \end{bmatrix}
    = 
    \begin{bmatrix}
        0 \\
        1 \\
        0
    \end{bmatrix}.
\end{equation*}
Solving this via row-reduction tells us that
\begin{equation}
\begin{aligned}
    c_{-1}  &= -\frac{b}{a(a + b)} \\
    c_0     &= \frac{b - a}{a b} \\
    c_1     &= \frac{a}{b\left( a + b \right)} 
\end{aligned}
\label{eq:Appendix:Finite_Diff:First:Central:Coefficients}
\end{equation}
Plugging these into equation \eqref{eq:Appendix:Finite_Diff:First:Central} gives 
\begin{equation*}
\begin{aligned}
    c_{-1} f(x - a) + c_0 f(x) + c_1 f(x + b) = f'(x) + \frac{1}{6} \left( -\frac{ba^2}{a + b} f'''\left(\xi_{-1}\right) + \frac{ab^2}{a + b} f'''\left(\xi_1\right) \right).
\end{aligned}
\end{equation*}
Since $[x - a, x + b]$ is compact and $f'''$ is continuous, it must be bounded on $[x - a, x + b]$.
Let $M > 0$ be a bound on $f'''$.
Then, because $a, b > 0$,
\begin{equation*}
\begin{aligned}
    \frac{\left|c_{-1} f(x - a) + c_0 f(x) + c_1 f(x + b)\right|}{h^2} &= \frac{1}{6} \frac{\left| \frac{ba^2}{a + b} f'''\left(\xi_{-1}\right) + \frac{ab^2}{a + b} f'''\left( \xi_1 \right) \right|}{h^2} \\
    &\leq \frac{1}{6} M \left( \frac{ba^2}{(a + b)h^2} + \frac{ab^2}{(a + b)h^2} \right) \\
    &= \frac{1}{6} M \left( \frac{b}{a + b} + \frac{a}{a + b} \right) \\
    &= \frac{1}{6}M
\end{aligned}
\end{equation*}
Notably, this result holds regardless of the value of $a$ or $b$; the values need only be positive.
Note that if the ratio $\tfrac{a}{b}$ is very large or very small, then $c_0$ will become very large.
However, because $c_0$ does not appear in the error ($f'''$) term of equation \eqref{eq:Appendix:Finite_Diff:First:Central}, a large $c_0$ value does not impact the asymptotic error.
Thus, unlike for forward-differences, we do not need to make any assumptions on the ratio $\tfrac{a}{b}$.

\subsection{Five-point finite difference approximations of \texorpdfstring{$f''(x)$}{f''(x)}}
\label{Appendix:Finite_Difference:Second}

Here, we will derive, forward- and mixed-difference approximations of $f''(x)$.
Throughout this sub-section, we assume $f$ is of class $C^4$ on $S$.
Let $a, b, c > 0$ and $h = \max\{a, b, c\}$.

\subsubsection{Five point forward-difference approximation of \texorpdfstring{$f''(x)$}{f''(x)}.}
\label{Appendix:Finite_Difference:Second:Forward}

Suppose we know $f(x)$, $f(x + a)$, $f(x + a + b)$, and $f(x + a + b + c)$.
We assume that $[x, x + a + b + c] \subseteq S$.
We seek coefficients $c_0$, $c_1$, $c_2$, and $c_3$ such that
\begin{equation}
    c_0 f(x) + c_1 f(x + a) + c_2 f(x + a + b) + c_3 f(x + a + b + c) = f''(x) + \mathcal{O}(h^2)
    \label{eq:Appendix:Finite_Diff:Second:Forward}
\end{equation}
We will also assume that there is some $C > 0$ such that $\tfrac{a}{b}, \tfrac{b}{c}, \tfrac{c}{b}, \tfrac{b}{c}, \tfrac{a}{c} \leq C$.
These constraints require that $a, b$, and $c$ have to be of the same order.
To begin, let us take forth-order Taylor expansions of $f(x + a)$, $f(x + a + b)$, and $f(x + a + b + c)$ about $x$:
\begin{alignat*}{4}
    &f(x + a)         &= f(x) &+ a           f'(x) &+ \tfrac{1}{2}a^2           f''(x) &+ \tfrac{1}{6}a^3           f'''(x) &+ \tfrac{1}{24}a^4           f^{(4)}(\xi_1), \\
    &f(x + a + b)     &= f(x) &+ (a + b)     f'(x) &+ \tfrac{1}{2}(a + b)^2     f''(x) &+ \tfrac{1}{6}(a + b)^3     f'''(x) &+ \tfrac{1}{24}(a + b)^4     f^{(4)}(\xi_2), \\
    &f(x + a + b + c) &= f(x) &+ (a + b + c) f'(x) &+ \tfrac{1}{2}(a + b + c)^2 f''(x) &+ \tfrac{1}{6}(a + b + c)^3 f'''(x) &+ \tfrac{1}{24}(a + b + c)^4 f^{(4)}(\xi_3)
\end{alignat*}
for some $\xi_1 \in (x, x + a)$, $\xi_2 \in (x + a, x + a + b)$, and $\xi_3 \in (x, x + a + b + c)$. 
Substituting this into equation \eqref{eq:Appendix:Finite_Diff:Second:Forward} gives
\begin{equation}
\begin{aligned}
     c_0 f(x) + c_1 f(x + a) + c_2 f(x + a + b) + &c_3 f(x + a + b + c) \\
     = &\left( c_0 + c_1                       + c_2                           + c_3                               \right) f(x)    \\
    + &\left(       c_1 a                     + c_2 (a + b)                   + c_3 (a + b + c)                   \right) f'(x)   \\
    + &\tfrac{1}{2} \left(       c_1 a^2                   + c_2 (a + b)^2                 + c_3 (a + b + c)^2                 \right) f''(x)  \\
    + &\tfrac{1}{6} \left(       c_1 a^3                   + c_2 (a + b)^3                 + c_3 (a + b + c)^3                 \right) f'''(x) \\
    + &\tfrac{1}{24}\left(       c_1 a^4 f^{(4)}(\xi_1)    + c_2 (a + b)^4 f^{(4)}(\xi_2)  + c_3 (a + b + c)^4 f^{(4)}(\xi_3)  \right)
\end{aligned}
\label{eq:Appendix:Finite_Diff:Second:Forward:Taylor}
\end{equation}
For this to have any hope of equaling $f'(x) + \mathcal{O}(h^2)$, we need the expression in front of $f(x)$ and $f''(x)$ to disappear and the one in front of $f'(x)$ to be $1$. 
In other words, we must have
\begin{equation*}
    \begin{bmatrix}
        1, && 1,    && 1,           && 1                \\
        0, && a,    &&( a + b),     && a + b + c        \\
        0, && a^2,  && (a + b)^2,   && (a + b + c)^2    \\
        0, && a^3,  && (a + b)^3,   && (a + b + c)^3 
    \end{bmatrix}    
    \begin{bmatrix}
        c_0 \\
        c_1 \\
        c_2 \\
        c_3
    \end{bmatrix}
    = 
    \begin{bmatrix}
        0 \\
        0 \\
        2 \\
        0
    \end{bmatrix}.
\end{equation*}
Brute force computation (row-reduction) shows that this system has the following solution:
\begin{equation}
\begin{aligned}
    c_0 &= \frac{ 2(3a + 2b + c)}{a(a +b)(a + b + c)}, \\
    c_1 &= \frac{-2(2a + 2b + c)}{ab(b + c)}, \\
    c_2 &= \frac{ 2(2a + b + c)}{bc(a + b)}, \\
    c_3 &= \frac{-2(2a + b)}{(a + b + c)(b + c)c}
\end{aligned}
\label{eq:Appendix:Finite_Diff:Second:Forward:Coefficients}
\end{equation}
Plugging these into equation \eqref{eq:Appendix:Finite_Diff:Second:Forward:Taylor} gives
\begin{equation*}
\begin{aligned}
    c_0 f(x) + &c_1 f(x + a) + c_2 f(x + a + b) + c_3 f(x + a + b + c) \\
    &= f''(x) \\
    &+ \frac{1}{24} \left( \frac{-2 a^3 (2a + 2b + c)}{b(b + c)} f'''\left(\xi_1\right) - \frac{ 2(a + b)^3 (2a + b + c)}{bc} f'''(\xi_2) + \frac{-2(2a + b)(a + b + c)^3}{(b + c)c} f'''\left(\xi_3\right) \right)
\end{aligned}
\end{equation*}
Because $f$ is of class $C^4$ on $S$ and because $[x, x + a + b + c] \subseteq S$ is compact, $f'''$ must be bounded on $[x, x + a + b + c]$.
Let $M > 0$ denote an upper bound. 
Then, we must have
\begin{equation*}
\begin{aligned}
    \frac{1}{h^2}\big| c_0 f(x) + c_1 f(x + a) + &c_2 f(x + a + b)  + c_3 f(x + a + b + c) - f''(x)\big| \\
    &= \frac{1}{24} \frac{\left|\frac{-2 a^3 (2a + 2b + c)}{b(b + c)} f'''\left(\xi_1\right) - \frac{ 2(a + b)^3 (2a + b + c)}{bc} f'''(\xi_2) + \frac{-2(2a + b)(a + b + c)^3}{(b + c)c} f'''\left(\xi_3\right)\right|}{h^2} \\
    &\leq \frac{1}{24} M \left( \frac{2 a^3 (2a + 2b + c)}{b(b + c)h^2}  + \frac{2(a + b)^3 (2a + b + c)}{bch^2} + \frac{2(2a + b)(a + b + c)^3}{(b + c)c h^2} \right)
\end{aligned}
\end{equation*}
However, because $h = \max\{a, b, c\}$, we must have $h \geq a$, $2h \geq a + b$, and $3h \geq a + b + c$ which means that
\begin{equation*}
\begin{aligned}
    \frac{1}{h^2}\big| c_0 f(x) + &c_1 f(x + a) + c_2 f(x + a + b) + c_3 f(x + a + b + c) - f'(x)\big| \\
    &\leq \frac{1}{24} M \left( \frac{2 a (2a + 2b + c)}{b(b + c)}  + \frac{8(a + b) (2a + b + c)}{bc} + \frac{18(2a + b)(a + b + c)}{(b + c)c} \right) \\
\end{aligned}
\end{equation*}
We also know that $a, b, c \geq 0$, which means that $\tfrac{1}{b + c} \leq \tfrac{1}{b}$, for instance.
We also know that $\tfrac{a}{b}, \tfrac{c}{b}, \tfrac{a}{c}, \tfrac{b}{c} \leq C$.
Therefore, 
\begin{equation*}
\begin{aligned}
    \frac{1}{h^2}\big| c_0 f(x) + &c_1 f(x + a) + c_2 f(x + a + b) + c_3 f(x + a + b + c) - f'(x)\big| \\
    &\leq \frac{1}{24} M \left( 2C \frac{(2a + 2b + c)}{(b + c)} + 8(1 + C)(3C + 1) + 18(2C + 1) \frac{2a + b}{b + c} \right) \\
    &\leq \frac{1}{24} M \left( 2C \frac{(2a + 2b + c)}{b} + 8(1 + C)(3C + 1) + 18(2C + 1) \frac{2a + b}{b} \right) \\
    &\leq \frac{1}{24} M \left( 2C (2 + 3C) + 8(1 + C)(3C + 1) + 18(2C + 1)^2 \right)
\end{aligned}
\end{equation*}
Taking the limit as $h$ goes to zero tells us that the derived scheme is indeed of order $\mathcal{O}(h^2)$.

\subsubsection{Five-point mixed-difference approximation of \texorpdfstring{$f''(x)$}{f''(x)}.}
\label{Appendix:Finite_Difference:Second:Mixed}

Suppose we know $f(x - a)$, $f(x)$, $f(x + b)$, and $f(x + b + c)$.
We assume that $[x - a, x + b + c] \subseteq S$.
We seek coefficients $c_{-1}$, $c_0$, $c_1$, and $c_2$ such that
\begin{equation}
    c_{-1} f(x - a) + c_0 f(x) + c_1 f(x + b) + c_2 f(x + b + c) = f''(x) + \mathcal{O}(h^2)
    \label{eq:Appendix:Finite_Diff:Second:Mixed}
\end{equation}
We also assume there is some $C > 0$ such that stuff happens.
To begin, let us take forth-order Taylor expansions of $f(x - a)$, $f(x + b)$, and $f(x + b + c)$ about $x$:
\begin{alignat*}{4}
    &f(x - a)       &= f(x) &- a       f'(x) &+ \tfrac{1}{2}a^2       f''(x) &- \tfrac{1}{6}a^3       f'''(x) &+ \tfrac{1}{24}a^4       f^{(4)}(\xi_1), \\
    &f(x + b)       &= f(x) &+       b f'(x) &+ \tfrac{1}{2}b^2       f''(x) &+ \tfrac{1}{6}b^3       f'''(x) &+ \tfrac{1}{24} b^4      f^{(4)}(\xi_2), \\
    &f(x + b + c)   &= f(x) &+ (b + c) f'(x) &+ \tfrac{1}{2}(b + c)^2 f''(x) &+ \tfrac{1}{6}(b + c)^3 f'''(x) &+ \tfrac{1}{24}(b + c)^4 f^{(4)}(\xi_3)
\end{alignat*}
for some $\xi_{-1} \in (x - a, x)$, $\xi_1 \in (x, x + b)$, and $\xi_2 \in (x, x + b + c)$. 
Substituting this into equation \eqref{eq:Appendix:Finite_Diff:Second:Mixed} gives
\begin{equation}
\begin{aligned}
     c_{-1} f(x - a) + c_0 f(x) + c_1 f(x + b) + & c_2 f(x + b + c) \\
     = &\left(                     c_{-1}     + c_0             + c_1                    + c_2                               \right) f(x)    \\
     + &\left(                    -c_{-1} a                     + c_1 b                  + c_2 (b + c)                   \right) f'(x)   \\
     + &\tfrac{1}{2} \left(       c_{-1} a^2                   + c_1 b^2                 + c_2 (b + c)^2                 \right) f''(x)  \\
     + &\tfrac{1}{6} \left(      -c_{-1} a^3                   + c_1 b^3                 + c_2 (b + c)^3                 \right) f'''(x) \\
     + &\tfrac{1}{24}\left(       c_{-1} a^4 f^{(4)}(\xi_{-1}) + c_1 b^4 f^{(4)}(\xi_1)  + c_2 (b + c)^4 f^{(4)}(\xi_2)  \right)
\end{aligned}
\label{eq:Appendix:Finite_Diff:Second:Mixed:Taylor}
\end{equation}
For this to have any hope of equaling $f'(x) + \mathcal{O}(h^2)$, we need the expression in front of $f(x)$ and $f''(x)$ to disappear and the one in front of $f'(x)$ to be $1$. 
In other words, we must have
\begin{equation*}
    \begin{bmatrix}
        1,    && 1,  && 1,     && 1            \\
        -a,   && 0,  && b,     && b + c        \\
        a^2,  && 0,  && b^2,   && (b + c)^2    \\
        -a^3, && 0,  && b^3,   && (b + c)^3 
    \end{bmatrix}    
    \begin{bmatrix}
        c_{-1} \\
        c_0 \\
        c_1 \\
        c_2
    \end{bmatrix}
    = 
    \begin{bmatrix}
        0 \\
        0 \\
        2 \\
        0
    \end{bmatrix}.
\end{equation*}
Brute force computation (row-reduction) shows that this system has the following solution:
\begin{equation}
\begin{aligned}
    c_{-1}  &= \frac{2(2b + c)}{a(a + b)(a + b + c)}                        \\
    c_0     &= \frac{-2\left(b(a + 2b + 3c) + c^2 - a^2\right)}{ba(b + c)(a + b + c)}  \\
    c_1     &= \frac{ 2(b + c - a)}{bc(a + b)}                              \\
    c_2     &= \frac{-2(b - a)}{c(b + c)(a + b + c)}
\end{aligned}
\label{eq:Appendix:Finite_Diff:Second:Mixed:Coefficients}
\end{equation}
Plugging these into equation \eqref{eq:Appendix:Finite_Diff:Second:Mixed:Taylor} gives
\begin{equation*}
\begin{aligned}
    c_{-1} f(x - a) + &c_0 f(x) + c_1 f(x + b) + c_2 f(x + b + c) \\
    &= f''(x) \\
    &+ \frac{1}{24} \left( \frac{2a^3(2b + c)}{(a + b)(a + b + c)} f'''\left(\xi_{-1}\right) - \frac{ 2b^3(b + c - a)}{c(a + b)} f'''(\xi_1) + \frac{-2(b + c)^3(b - a)}{c(a + b + c)} f'''\left(\xi_2\right) \right)
\end{aligned}
\end{equation*}
Because $f$ is of class $C^4$ on $S$ and because $[x, x + a + b + c] \subseteq S$ is compact, $f'''$ must be bounded on $[x, x + a + b + c]$.
Let $M > 0$ denote an upper bound. 
Then, we must have
\begin{equation*}
\begin{aligned}
    \frac{1}{h^2}\big| c_0 f(x) + c_1 f(x + a) + &c_2 f(x + a + b)  + c_3 f(x + a + b + c) - f''(x)\big| \\
    &= \frac{1}{24} \frac{\left|\frac{-2 a^3 (2a + 2b + c)}{b(b + c)} f'''\left(\xi_1\right) - \frac{ 2(a + b)^3 (2a + b + c)}{bc} f'''(\xi_2) + \frac{-2(2a + b)(a + b + c)^3}{(b + c)c} f'''\left(\xi_3\right)\right|}{h^2} \\
    &\leq \frac{1}{24} M \left( \frac{2 a^3 (2a + 2b + c)}{b(b + c)h^2}  + \frac{2(a + b)^3 (2a + b + c)}{bch^2} + \frac{2(2a + b)(a + b + c)^3}{(b + c)c h^2} \right)
\end{aligned}
\end{equation*}
However, because $h = \max\{a, b, c\}$, we must have $h \geq a$, $2h \geq a + b$, and $3h \geq a + b + c$ which means that
\begin{equation*}
\begin{aligned}
    \frac{1}{h^2}\big| c_0 f(x) + &c_1 f(x + a) + c_2 f(x + a + b) + c_3 f(x + a + b + c) - f'(x)\big| \\
    &\leq \frac{1}{24} M \left( \frac{2 a (2a + 2b + c)}{b(b + c)}  + \frac{8(a + b) (2a + b + c)}{bc} + \frac{18(2a + b)(a + b + c)}{(b + c)c} \right) \\
\end{aligned}
\end{equation*}
We also know that $\tfrac{a}{b}, \tfrac{c}{b}, \tfrac{a}{c}, \tfrac{b}{c} \leq C$.
Therefore, 
\begin{equation*}
\begin{aligned}
    \frac{1}{h^2}\big| c_0 f(x) + &c_1 f(x + a) + c_2 f(x + a + b) + c_3 f(x + a + b + c) - f'(x)\big| \\
    &\leq \frac{1}{24} M \left( 2C \frac{2a + 2b + c}{b + c} + 8(1 + C) \frac{2a + b + c}{c} + 18(2C + 1) \frac{a + b + c}{b + c} \right) \\
    &\leq \frac{1}{24} M \left( 2C \frac{2a + 2b + c}{b} + 8(1 + C) \frac{2a + b + c}{c} + 18(2C + 1) \frac{a + b + c}{b} \right) \\
    &\leq \frac{1}{24} M \left( 2C(3C + 1) + 8(1 + C)(3C + 1) + 18(2C + 1)(2C + 1) \right)
\end{aligned}
\end{equation*}
Taking the limit as $h$ goes to zero tells us that the derived scheme is indeed of order $\mathcal{O}(h^2)$.

\end{document}